\documentclass{article}

\usepackage[numbers]{natbib}

\usepackage[final]{neurips_2022}

\usepackage{tcolorbox}
\usepackage{pifont}
\definecolor{mydarkgreen}{RGB}{39,130,67}
\definecolor{mydarkred}{RGB}{192,47,25}

\usepackage{csquotes}
\usepackage{verbatim}
\usepackage[utf8]{inputenc} 
\usepackage[T1]{fontenc}    
\usepackage{hyperref}       
\usepackage{url}            
\usepackage{booktabs}       
\usepackage{amsfonts}       
\usepackage{nicefrac}       
\usepackage{amsmath}
\usepackage{amssymb}
\usepackage[colorinlistoftodos,bordercolor=orange,backgroundcolor=orange!20,linecolor=orange,textsize=scriptsize]{todonotes}
\usepackage{enumitem}
\usepackage{pifont}
\usepackage{natbib}
\usepackage{algorithm,algorithmicx,algpseudocode}
\usepackage{wrapfig}
\usepackage[flushleft]{threeparttable}
\usepackage{makecell}
\usepackage{multirow}
\usepackage{caption}
\usepackage[capitalize,noabbrev]{cleveref}
\usepackage{xcolor}
\definecolor{ForestGreen}{RGB}{34,139,34}

\newcommand{\myred}[1]{{\color{red}#1}}
\newcommand{\myblue}[1]{{\color{blue}#1}}
\allowdisplaybreaks
\usepackage{mathtools}
\newcommand{\eqdef}{\vcentcolon=}

\newcommand{\algnamenormal}[1]{{\color{ForestGreen} \sf #1}}
\newcommand{\algname}[1]{{\color{ForestGreen}\small \sf #1}}
\newcommand{\algnametiny}[1]{{\color{ForestGreen}\tiny \sf #1}}

\def\R{\mathbb{R}}

\def\R{\mathbb R}
\def\E{\mathbb E}
\def\EE{\mathbb E}

\def\Z{\mathcal Z}

\def\la{\langle}
\def\ra{\rangle}

\newcommand{\EndProof}{\begin{flushright}$\square$\end{flushright}}

\def\<#1,#2>{\langle #1,#2\rangle}

\newcommand{\add}[1]{{\color{black}#1}}

\newtheorem{theorem}{Theorem}[section]

\newtheorem{corollary}[theorem]{Corollary}

\newtheorem{assumption}[theorem]{Assumption}
\newtheorem{example}[theorem]{Example}

\title{Distributed Methods with Compressed Communication for Solving Variational Inequalities, with Theoretical Guarantees}

\author{%
  Aleksandr Beznosikov \\
  Innopolis University\thanks{Research Center for Artificial Intelligence, Innopolis University}, MIPT\thanks{Moscow Institute of Physics and Technology}, HSE University and Yandex, Russia \\
  \texttt{anbeznosikov@gmail.com}
  \And
  Peter Richt\'{a}rik \\
  KAUST\thanks{King Abdullah University of Science and Technology}, Saudi Arabia \\
  \texttt{peter.richtarik@kaust.edu.sa}
  \And
  Michael Diskin \\
  HSE University and Yandex, Russia \\
  \texttt{michael.s.diskin@gmail.com}
  \And
  Max Ryabinin \\
  Yandex and HSE University, Russia \\
  \texttt{mryabinin0@gmail.com}
  \And
  Alexander Gasnikov \\
  MIPT, HSE University and IITP RAS\thanks{Institute for Information Transmission Problems RAS}, Russia \\
  \texttt{gasnikov@yandex.ru}
}

\begin{document}

\maketitle

\begin{abstract}
  Variational inequalities in general and saddle point problems in particular are increasingly relevant in machine learning applications, including adversarial learning, GANs, transport and robust optimization. With increasing data and problem sizes necessary to train high performing models across various applications, we need to rely on parallel and distributed computing. However, in distributed training, communication among the compute nodes is a key bottleneck during training, and this problem is exacerbated for high dimensional and over-parameterized models. Due to these considerations, it is important to equip existing methods with strategies that would allow to reduce the volume of transmitted information during training while obtaining a model of comparable quality. In this paper, we present the first theoretically grounded distributed methods for solving variational inequalities and saddle point problems using compressed communication: \algname{MASHA1} and \algname{MASHA2}. Our theory and methods allow for the use of both unbiased (such as Rand$k$; \algname{MASHA1}) and contractive (such as Top$k$; \algname{MASHA2}) compressors. New algorithms support bidirectional compressions, and also can be modified for stochastic setting with batches and for federated learning with partial participation of clients. We empirically validated our conclusions using two experimental setups: a standard bilinear min-max problem, and large-scale distributed adversarial training of transformers.
\end{abstract}

\section{Introduction} \label{sec:intro}

\subsection{The expressive power of variational inequalities}  Due to their  abstract mathematical nature and the associated flexibility they offer in modeling various practical problems of interests, {\em variational inequalities (VI)} have been an active area of research in applied mathematics for more than half a century~\citep{NeumannGameTheory1944, HarkerVIsurvey1990, VIbook2003}. It is well known that VIs can be used to formulate and study optimization problems, {\em saddle point problems (SPPs)}, games and fixed point problems, for example, in an elegant unifying mathematical framework~\citep{Convex-Analysis-and-Monotone-Operator-Theory}. 

Recently, a series of works by various authors \cite{daskalakis2018training,gidel2018a,mertikopoulos2018optimistic,chavdarova2019reducing,pmlr-v89-liang19b} built a bridge between VIs/SPPs and GANs \citep{goodfellow2014generative}. This allows to successfully transfer established insights and well-known techniques from the vast literature on VIs/SPPs, such as averaging and extrapolation,  to the study of GANs. Besides their usefulness in  studying GANs and alternative adversarial learning models \citep{madry2018towards}, VIs/SPPs have recently attracted considerable attention of the machine learning community due to their ability to model other situations where  the minimization of a single  loss function does not suffice, such as auction theory \citep{Shapire-auction2015}, supervised learning with non-separable loss \cite{Thorsten} or non-separable regularizer \cite{bach2011optimization} and reinforcement learning \citep{Pinto2017,Omidshafiei2017:rl,Jin2020:mdp}.

In summary, VIs have recently become a potent tool enabling new advances in practical machine learning situations reaching beyond supervised learning where optimization problems and techniques, which can be seen as special instances of VIs and methods for solving them, reign supreme.

\subsection{Training of supervised models via distributed optimization} 

\add{On the other hand, for classical and much better understood {\em supervised machine learning}/minimization problems, researchers and practitioners face other challenges, which, until recently, have been outside of VI's research.}
Indeed, the training of modern supervised machine learning models in general, and deep neural networks in particular, is still extremely challenging. Due to their desire to improve the generalization of deployed models, machine learning engineers need to rely on  training datasets of ever increasing sizes and on elaborate over-parametrized models   \citep{ACH-overparameterized-2018}. Supporting workloads of such unprecedented magnitudes would be impossible without combining the latest advances in  hardware acceleration, distributed systems  and {\em distributed algorithm design} \citep{DMLsurvey}.

When training such modern supervised models in a distributed fashion, {\em communication cost} is often the bottleneck of the training system, and for this reason, a lot of effort was recently targeted at the design of communication efficient distributed optimization methods~\citep{FEDLEARN, cocoa-2018-JMLR, Ghosh2020, MARINA}. A particularly successful technique for improving the communication efficiency of distributed first order optimization methods is {\em communication compression.} The idea behind this technique is rooted in the observation that in practical implementations  it \add{is} often advantageous to communicate  messages compressed via (often randomized) {\em lossy compression techniques} instead of communicating the full messages~\citep{1bit, alistarh2017qsgd}. If the number of parallel workers is large enough, the noise introduced by compression is reduced, and training with compressed communication will often lead to comparable test error while reducing the amount of communicated bits, which results in faster training, both in theory and practice~\citep{DIANA, MARINA}.

\subsection{Two classes of compression operators}
The paper focuses on compression methods for distributed VIs and SPPs. Let us give the main definitions. We say that a (possibly) stochastic mapping $Q:\R^d\to \R^d$ is an {\em unbiased compression operator} if there exists a constant $q \geq 1$ such that
\begin{align}
\label{quant}
    \EE{Q(z)} = z,\quad \EE{\| Q(z) \|^2} \leq q \| z\|^2, \quad  \forall z \in \R^d. 
\end{align}
Further, we say that a stochastic mapping $C:\R^d\to \R^d$ is a {\em contractive compression operator} if there exists a constant $\delta \geq 1$ such that 
\begin{align}
\label{compr}
    \EE{\| C (z) - z\|^2} \leq (1 - 1/\delta) \| z\|^2, \quad  \forall z \in \R^d. 
\end{align}
If $b$ is the number of bits needed to represent a single float (e.g., $b=32$ or $b=64$), then the number of bits needed to represent a generic vector $z\in \R^d$ is $\| z \|_{\text{bits}} \eqdef bd$. To describe how much a compression operator reduces its input vector on average, we define the notion of expected density, denoted via $\beta^{-1} \eqdef \frac{1}{bd}\EE{\| Q(z) \|_{\text{bits}}}$, where $\| Q(z) \|_{\text{bits}}$ is the number of bits needed to represent the quantized vector $Q(z)$. Note that $\beta \geq 1$.  For the Rand$k$ operator~\citep{Alistarh-EF2018, beznosikov2020biased} we have $q = \beta = d/k$.

\subsection{Towards communication-efficient distributed methods for VIs and SPPs}
Classical VI/SPP algorithms such as the {\em Extra Gradient method} originally proposed by \cite{Korpelevich1976TheEM} and later studied by many authors \cite{Nemirovski2004, juditsky2008solving}, including in a distributed environment \cite{srivastava_distributed, liu2019decentralized, Mukherjee2020:decentralizedminmax, rogozin2021decentralized}.
Among them, a number of works stand out trying to solve the communication bottleneck challenge using various approaches such as local steps, data-similarity etc.\cite{doi:10.1177/0142331213487545, hou2021efficient, deng2021local, beznosikov2021distributed1, beznosikov2021distributed}. But despite the fact that the use of compression is one of the most popular communication-efficient approaches for distributed minimization problems, {\em no work has yet paid attention to the compression technique neither for distributed SPPs nor for VIs}, with the exception of the work \cite{doi:10.1177/0142331213487545}, which relies on  rounding to the nearest integer multiple of a certain quantity. This compression mechanism does not offer theoretical benefits and does not even lead to convergence to the solution \add{since} the errors introduced through rounding persist and prevent the method from solving the problem.

\section{Summary of Contributions} \label{sec:contr}

In this paper, we investigate whether it is possible to design communication-efficient algorithms for solving distributed VI/SPP by borrowing generic communication compression techniques \eqref{quant} and \eqref{compr} from the optimization literature~\citep{1bit, alistarh2017qsgd,DIANA, MARINA, EF21} and embedding them into established, efficient methods for solving VIs/SPPs \citep{Korpelevich1976TheEM, Nemirovski2004, juditsky2008solving,alacaoglu2021stochastic}. Whether or not this is possible is an open problem. In summary,
\vspace{-0.15cm}
\begingroup
\addtolength\leftmargini{-0.7cm}
\begin{quote}
{\em 
we design the first algorithms with compression for solving general distributed VI/SPP (see Section~\ref{sec:problem}, Equation~\ref{VI}) in the deterministic (see \eqref{MK}), stochastic (see \eqref{MK_fs}) and federated (see \eqref{pp_setup}) regimes, supporting both unbiased (\algname{MASHA1} = Algorithms~\ref{main_alg_qeg}, \ref{alg_qeg_vr}, \ref{alg_qeg_pp}) and contractive (\algname{MASHA2} = Algorithms~\ref{main_alg_qfb}, \ref{alg_qfbvr}, \ref{alg_qfb_pp}) compressors. Convergence of all our methods are analyzed in strongly-monotone (strongly convex - strongly concave), monotone (convex - concave) non-monotone/minty (non-convex-non-concave) cases.
}
\end{quote}
\endgroup
\vspace{-0.2cm}

\subsection{Two types of compressors}

We develop two approaches for distributed VIs/SPPs depending on whether we use unbiased \eqref{quant} or contractive \eqref{compr} compressors, since each type of compressor demands a different algorithmic design and a different analysis. In particular, contractive compressors are notoriously hard to analyze even for optimization problems~\citep{error_feedback, EF21}. Our method based on unbiased compressors is called \algname{MASHA1} (Algorithm~\ref{main_alg_qeg}), and our method based on contraction compressors is called \algname{MASHA2} (Algorithm~\ref{main_alg_qfb}). 

\subsection{Theoretical complexity results}

We establish a number of theoretical complexity results for our methods, which we summarize in Table~\ref{tab:comparison1} (Appendix \ref{sec:table}). We consider the strongly monotone (strongly convex - strongly concave), monotone (convex - concave) regimes as well as  the more general non-monotone/minty (non-convex-non-concave) regime. In the strongly monotone case we obtain linear convergence results ($O(\log 1/\epsilon)$) in terms of the distance to solution, in the monotone we obtain fast sublinear convergence results ($O(1/\epsilon)$) in terms of the {\em gap} function, and in the non-monotone case we have sublinear convergence results ($O(1/\epsilon^2)$) in terms of the \add{Euclidean} norm of the operator. To get an estimate for the number of information transmitted, one need to multiply the estimates from Table~\ref{tab:comparison1} by $1/\beta$. Then we get that from the point of view of the transmitted information (and also time for communications), \algname{MASHA1} is better by a factor $\sqrt{1/\beta + 1/M}$ ($M$ -- number of workers) in comparison with the classical Extra Gradient. It means that we get an acceleration of $\min\{\sqrt{\beta};\sqrt{M}\}$ times. For example, ADIANA  from \cite{pmlr-v119-li20g}(the theoretical SOTA method with unbiased compressions for strongly convex minimization) has the same accelaration. The same situation is with \algname{MASHA2}. The method has the same compression dependent multiplier as ECLK from \cite{qian2020error} (the theoretical SOTA with contractive compression for minimization).  Based on these facts, we hypothesize that \algname{MASHA1} and \algname{MASHA2} have unimprovable estimates (see Appendix \ref{sec:opt}).

\subsection{Stochastic case and variance reduction}

\algname{MASHA1} and  \algname{MASHA2} are designed to handle the {\em deterministic} setting. But often\add{,} in practice, the computation of the full \add{operators/gradients} is expensive, then we need to deal with {\em stochastic} realizations. In particular, a popular case is when each \add{operator/gradient} has a finite-sum structure on its own, e.g. , finite-sum of batches. For this issue, we consider two modifications: \algname{VR-MASHA1} (Algorithm~\ref{alg_qeg_vr}) and  \algname{VR-MASHA2} (Algorithm~\ref{alg_qfbvr}). Both are enhanced with bespoke {\em variance-reduction} techniques for better theoretical and practical performance. These results can be interesting in the non-distributed case. As far as we know, we are the first who consider variance reduction for non-monotone VIs. We found only one paper on non-convex-concave saddle point problems \cite{yang2020global} under the PL condition. See Appendix \ref{sec:stoch} \add{for} details.

\subsection{Federated learning and partial participation}

 \textit{Federated learning} \cite{FEDLEARN, kairouz2019advances} is an important and popular branch of distributed methods. Therefore, a good bonus for the algorithm is that it can be easily adapted for it. In a federated setup where the computing devices are mobile phones, tablets, personal computers etc, the importance of the communication bottleneck is even higher. \add{In such circumstances, devices can have weak and slow connections, or they can even disconnect for a while. At such moments, it is not necessary to interrupt the learning process, and only available devices can be used.} Therefore, we introduce two modifications: \algname{PP-MASHA1} (Algorithm~\ref{alg_qeg_pp}) and  \algname{PP-MASHA2} (Algorithm~\ref{alg_qfb_pp}), that support the mode of \textit{partial participation} of devices in the learning process. For minimization problems, a combination of quantization and partial participation occurs in \cite{horvath2020better, philippenko2020bidirectional, MARINA}. The results are contained in Appendix \ref{sec:fed}.

\subsection{Bidirectional compression}

Most methods, especially with contractive compressors, only use compression when transferring information from devices to the server. Meanwhile, quite often in practical situations, the transfer of information from the server to the device is also expensive \cite{horvath2019natural,tang2019doublesqueeze,philippenko2020bidirectional}. 
\add{In such situations it also makes sense to compress the information when sending it from the server to the agents. We can highlight some works on bidirectional unbiased \cite{philippenko2020bidirectional} and contractive compressors \cite{zheng2019communication,tang2019doublesqueeze,liu2020double,fatkhullin2021ef21} for distributed minimization problems. But most of these methods have their small shortcomings in theoretical analysis such as deterministic setting only, homogeneity of local functions, etc.} All our methods \algname{MASHA1}, \algname{MASHA2} and their modifications support bidirectional compression. See Appendix \ref{sec:masha1} and \ref{sec:MASHA2} for details.

\subsection{Experiments}

Toy experiments on bilinear problems show that methods with compression for minimization problems may not work (diverge) for SPPs. Also we verify that \algname{MASHA1} and \algname{MASHA2} are much better than the classical Extra Gradient with added unbiased compression. Experiments on adversarial training of large-scale  transformer (ALBERT) show the practical importance of compression in distributed methods for large SPPs.


\section{Problem Formulation and Assumptions}\label{sec:problem}


\subsection{Problem formulation}
We study distributed variational inequality (VI) problem
\begin{equation}
    \label{VI}
    \text{Find} ~~ z^* \in {\R}^d ~~ \text{such that} ~~ \langle F(z^*), z - z^* \rangle \geq 0,~~ \forall z \in {\R}^d ,
\end{equation}
where  $F: \R^d \to \R^d $ is an operator with certain favorable properties (e.g., Lipschitzness and monotonicity).  We assume that the training data describing $F$ is {\em distributed} across $M$ workers/nodes/clients 
\begin{equation}
    \label{MK}
  F(z) \eqdef \frac{1}{M}\sum\limits_{m=1}^M F_m(z),
\end{equation}
where $F_m:\R^d \to \R^d$ for all $m \in \{1,2,\dots,M\}$. Next, we give main examples of VIs to show the breadth of this formalism.
\begin{example}[Minimization] \label{ex:min} Consider the minimization problem:
\begin{align}
\label{eq:min}
\min_{z \in \R^d} f(z).
\end{align}
Suppose that $F(z) \eqdef \nabla f(z)$. Then, if $f$ is convex, it can be proved that $z^* \in \R^d$ is a solution for \eqref{VI} if and only if $z^* \in \R^d$ is a solution for \eqref{eq:min}. And if the function $f$ is non-convex, then $z^* \in \R^d$ is a solution for \eqref{VI} if and only if $\nabla f(z^*) = 0$, i.e. $z^*$ is a stationary point.
\end{example}
\begin{example}[Saddle point problem]  Consider the saddle point problem:
\begin{align}
\label{eq:minmax}
\min_{x \in \R^{d_x}} \max_{y \in \R^{d_y}} g(x,y).
\end{align}
Suppose that $F(z) \eqdef F(x,y) = [\nabla_x g(x,y), -\nabla_y g(x,y)]$ and $\Z = \R^{d_x} \times \R^{d_y}$. Then, if $g$ is convex-concave, it can be proved that $z^* \in \Z$ is a solution for \eqref{VI} if and only if $z^* \in \Z$ is a solution for \eqref{eq:minmax}. And if the function $g$ is non-convex-non-concave, then $z^* \in \Z$ is a solution for \eqref{VI} if and only if $\nabla_x g(x^*, y^*) = 0$ and $\nabla_y g(x^*, y^*) = 0$, i.e. $z^*$ is a stationary point.
\end{example}

If minimization problems are widely researched separately from variational inequalities. The study of saddle point problems often is associated with variational inequalities, therefore saddle point problems are strongly related to variational inequalities. 
\begin{example}[Fixed point problem]
Consider the fixed point problem:
\begin{align}
\label{eq:fixedpoint}
 \text{Find} ~~ z^* &\in \R^d ~~ \text{such that} ~~
    T(z^*) = z^*,
\end{align}
where $T: \R^d \to \R^d $ is an operator. With $F(z) = z - T(z)$, it can be proved that $z^* \in \R^d$ is a solution for \eqref{VI} if and only if $F(z^*) = 0$, i.e. $z^* \in \R^d$ is a solution for \eqref{eq:fixedpoint}.
\end{example}


\subsection{Assumptions}

Next, we list two key assumptions - both are standard in the literature on VIs. 
\begin{assumption}[Lipschitzness] \label{as3}
The operator $F$ is $L$-Lipschitz continuous, i.e. for all $z_1, z_2 \in \R^d$ we have
$
  \|F(z_1) - F(z_2)\| \leq L \|z_1-z_2\|.
$

Each operator $F_m$ is $L_m$-Lipschitz continuous, i.e. for all $z_1, z_2 \in \R^d$ it holds
$
  \|F_m(z_1) - F_m(z_2)\| \leq L_m \|z_1-z_2\|.
$
Let us define new constant $\tilde L$ as follows $\tilde L^2 = \frac{1}{M} \sum\limits_{m=1}^M L_m^2$.
\end{assumption}
For saddle point problems, these properties are equivalent to smoothness.
\begin{assumption}[Monotonicity] \label{as1}
We need three cases of monotonicity

\textbf{(SM) Strong monotonicity.} The operator $F$ is $\mu$-strongly monotone, i.e. for all $z_1, z_2 \in \R^d$ we have
$
    \langle F(z_1) - F(z_2), z_1 - z_2 \rangle \geq \mu\|z_1-z_2\|^2.
$

\textbf{(M) Monotonicity.} The operator $F$ is monotone, i.e. for all $z_1, z_2 \in \R^d$ we have
$
    \langle F(z_1) - F(z_2), z_1 - z_2 \rangle \geq 0.
$

\textbf{(NM) Non-monotonicity.} The operator $F$ is non-monotone (minty), if and only if there exists $z^* \in \R^d$ such that for all $z \in \R^d$ we have
$
    \langle F(z), z - z^* \rangle\geq 0.
$
\end{assumption}
The last assumption is called the minty or variational stability condition. It is not a general non-monotonicity, but is already associated in the community with non-monotonicity \cite{dang2015convergence,iusem2017extragradient, mertikopoulos2018optimistic, liu2019decentralizedprox, kannan2019optimal, hsieh2020explore, diakonikolas2021efficient}, particularly with the setup, which is somewhat appropriate for GANS \cite{liu2019towards, liu2019decentralized, dou2021one, BARAZANDEH2021108245}.

\section{\algname{MASHA}}

In this Section we present new algorithms and their convergence. Section \ref{sec:masha1_main} is devoted to the algorithm (\algname{MASHA1}) with unbiased compression. Section \ref{sec:masha2} -- to algorithm (\algname{MASHA2}) with contractive compression. Appendix gives modifications for the stochastic case -- Section \ref{sec:stoch}, and for the federated learning -- Section \ref{sec:fed}. Appendix \ref{sec:opt} is devoted to the hypothesis about optimality of \algname{MASHA1} and \algname{MASHA2}.

\subsection{\algname{MASHA1}: Handling Unbiased Compressors} \label{sec:masha1_main}

Before presenting our algorithm, let us discuss which approaches can be used to construct it. As discussed in Sections \ref{sec:intro} and \ref{sec:contr}, compression methods play an important role in distributed minimization problems. All these methods are modifications of the classical GD. For instance, the authors of \cite{alistarh2017qsgd} compress stochastic gradients. Therefore, it is a natural idea to use GD-type methods for VIs as well. But it is a well-known fact that GD-type methods can give bad convergence estimates (see Section B.1 from \cite{palaniappan2016stochastic}) or do not converge at all (see Section 7.2 and 8.2 from \cite{goodfellow2016nips}) even on the simplest SPPs and VIs. From a practical point of view, this approach can also fail (see QSGD and EF in Section \ref{sect:exp_bilinear}). In the non-distributed case, this problem has long been solved and the Extra Gradient method \cite{Korpelevich1976TheEM,Nemirovski2004,juditsky2008solving} is used instead of GD:
\begin{equation}
\label{eq:eg}
    z^{k+1/2} = z^k - \gamma F(z^k), \quad z^{k+1} = z^k - \gamma F(z^{k+1/2}).
\end{equation}
This method is optimal for both VIs and SPPs and has an estimate of convergence $\mathcal{\tilde O}(\nicefrac{L}{\mu})$ in the strongly monotone case. Therefore, the second idea for the compressed method is to add compression operators to the method \eqref{eq:eg}, e.g. use $Q_k(F(z^k))$ and $Q_{k+1/2}(F(z^{k+1/2}))$ instead of $F(z^k)$ and $F(z^{k+1/2})$. In Section \ref{sec:ceg} we analyse this method, but it gives an estimate $\mathcal{\tilde O}\left(1 + q/M) \cdot \nicefrac{L^2}{\mu^2}  \right)$, which is considerably worse in terms of $\nicefrac{L}{\mu}$ than the original Extra Gradient method. The key problem is that in the analysis one has to deal with $\|Q_k(F(z^{k+1/2})) - Q_{k+1/2}(F(z^k))\|^2$. Without compression operators, such difference is easily evaluated using Assumption \ref{as3}. But when the compression operators are different (in fact the same, but have different randomness) we cannot make a good estimate for this term. The idea arises to use the same randomness in both steps of the method \eqref{eq:eg}, namely to substitute $Q_k(F(z^k))$ and $Q_{k}(F(z^{k+1/2}))$. But then $z^{k+1/2}$ depends on the randomness $Q_k$, and hence $Q_{k}(F(z^{k+1/2}))$ is biased, which further complicates the analysis. For exactly the same reasons, the various optimistic/single call modifications \cite{popov1980modification, gidel2018a, hsieh2019convergence, mokhtari2020convergence} of the Extra Gradient method did not work for us either. We have also test the method \eqref{eq:eg} with compressions in practice (see \algname{CEG} in Section \ref{sect:exp_bilinear}), and it turns out to be worse than the method we will present below. 
In the end, the use of variance reduction and negative momentum techniques \cite{alacaoglu2021stochastic} is key in creating our algorithm. These tricks are not in themselves relevant to distributed problems, but, in our case, they help in creating \algname{MASHA1} and \algname{MASHA2}.

\setlength{\textfloatsep}{3pt}
\begin{algorithm}[h]
	\caption{\algname{MASHA1}}
	\label{main_alg_qeg}
	\begin{algorithmic}
\State
\noindent {\bf Parameters:}  Stepsize $\gamma>0$, parameter $\tau \in (0;1)$, number of iterations $K$.\\
\noindent {\bf Initialization:} Choose  $z^0 = w^0 \in \mathcal{Z}$. \\
Devices send $F_m(w^0)$ to server and get $F(w^0)$
\For {$k=0,1, 2, \ldots, K-1$ }
\For {each device $m$ in parallel}
\State $z^{k+1/2} = \tau z^k + (1 - \tau) w^k - \gamma F(w^k)$ 
\State Sends  \text{\small{$g^k_m = Q^{\text{dev}}_m(F_m(z^{k+1/2}) - F_m(w^k))$}} to server
\EndFor
\For {server}
\State Sends to devices $g^k = Q^{\text{serv}} \left [\frac{1}{M} \sum_{m=1}^M g^k_m\right]$
\State Sends to devices one bit $b_k$ : 1 with probability $1 - \tau$, 0 with with probability $\tau$
\EndFor
\For {each device $m$ in parallel}
\State  $z^{k+1} = z^{k+1/2} - \gamma  g^k$ 
\State If {$b_k = 1$} then $w^{k+1} = z^{k}$, 
sends $F_m(w^{k+1})$  to server and gets $F(w^{k+1})$
\State else $w^{k+1} = w^k$  
\EndFor
\EndFor
	\end{algorithmic}
\end{algorithm}

\add{At the beginning of each \algname{MASHA1} iteration, all devices know the value of $F(w^k)$, hence they can calculate the value of $z^{k+1/2}$ locally without communications. Further, each device sends the compressed version of the difference $F_m(z^{k+1/2}) - F_m(w^k)$ to the server. The compression on these transfers is done by their local $\{Q^{\text{dev}}_m\}$ operators. The server aggregates the information from devices, averages it, compresses by $Q^{\text{serv}}$ operator and makes a broadcast to all devices. As a result, an unbiased estimate of $F(z^{k+1/2}) - F(w^k)$ appears at each node. Also, the nodes receive one bit of information $b_k$. This bit is generated randomly on the server and is equal to $1$ with probability
$1-\tau$ (where $1-\tau$ is small). Note that $b_k$ can be generated locally, it is enough to use the same random generator and set the same seed on all devices. Next, the devices locally make a final update on $z^{k+1}$. The final step is an update of $w^{k+1}$: if $b_k = 1$, then $w^{k+1} = z^{k}$ or otherwise $w^{k+1} = w^k$. 
In the case when $w^{k+1} = z^{k}$, we need to exchange the uncompressed values of $F_m (w^{k+1})$ in order to ensure that at the beginning of the next iteration the value of $F(w^{k+1})$ is known to all agents.}
We use a possibly difference compressor on each device and also on the server. To distinguish between them, we denote the following notation: $Q^{\text{dev}}_m$, $q^{\text{dev}}_m$, $\beta^{\text{dev}}_m$  and $Q^{\text{serv}}$, $q^{\text{serv}}$, $\add{\beta}^{\text{serv}}$.

\begin{theorem} \label{main_th_q}
Let Assumption \ref{as3} and one case of Assumption \ref{as1} are satisfied. Then for some step $\gamma$ the following estimates on \algname{MASHA1} number of iterations to achieve $\varepsilon$-solution holds 

$\bullet$ in strongly monotone case (in terms of 
$
\EE[\| z^{K} - z^* \|^2 ] \sim \varepsilon
$):
$
 \mathcal{O}([\frac{1}{1 - \tau} + \tfrac{C_q}{\mu \sqrt{1-\tau}}]\log \tfrac{1}{\varepsilon});
$

$\bullet$ in monotone case (in terms of 
$
\EE\max_{z \in \mathcal{C}} [ \langle F(u),  (\tfrac{1}{K}\sum_{k=0}^{K-1} z^{k+1/2}) - u  \rangle] \sim \varepsilon
$):
$
\mathcal{O}(\tfrac{C_q\|z^{0} - z^* \|^2}{\varepsilon \sqrt{1-\tau}});
$

$\bullet$ in non-monotone case (in terms of 
$
    \EE[\tfrac{1}{K}\sum_{k=0}^{K-1}\| F(w^k)\|^2] \sim \varepsilon^2
$):
$ 
\mathcal{O}(\tfrac{C_q^2 \|z^{0} - z^* \|^2}{\varepsilon^2 (1 - \tau)});
$

where $C^2_q = \tfrac{q^{\text{serv}}}{M^2} \sum_{m=1}^M (q_m^{\text{dev}} L_m^2 + (M-1) \tilde L^2)$.
\end{theorem}
A full \add{description} of the algorithm, as well as a full statement of the theorem with proof, can be found in  Appendix \ref{sec:masha1}.

The bounds in Theorem \ref{main_th_q} are related to $\tau$. Let us find \add{an} optimal way to choose it. Note that (in average) once per $\nicefrac{1}{(1-\tau)}$ iterations (when $b_k = 1$), we send uncompressed information. Based on this observation, we can find the best option for $\tau$. Let us analyze the case of compressions only on the devices' side ($q^{\text{serv}} = 1$). For simplicity, we put $Q^{\text{dev}}_m = Q$ with $q^{\text{dev}}_m = q$ and $\beta^{\text{dev}}_m = \beta$, also $L_m = \tilde L = L$. Since compression is done only on devices, we assume that the server's broadcast is cheap and we only care about devices. Then at each iteration the device sends $\mathcal{O}\left( \nicefrac{1}{\beta} + 1-\tau\right)$ bits  -- each time information compressed by $\beta$ times and with probability $1-\tau$ we send the full package. From where we immediately get the optimal choice for $\tau$:
\begin{corollary} \label{main_cor1_q}
Let Assumption \ref{as3} and one case of Assumption \ref{as1} are satisfied. Then for some step $\gamma$ and $1-\tau = \nicefrac{1}{\beta}$ the following estimates on \algname{MASHA1} number of iterations to achieve $\varepsilon$-solution holds \\
$\bullet$ in \add{strongly} monotone case:
$
 \mathcal{O}([\beta + \sqrt{\tfrac{q\beta}{M} + \beta} \cdot \tfrac{L}{\mu}]\log \tfrac{1}{\varepsilon});
$\\
$\bullet$ in monotone case:
$
\mathcal{O}(\sqrt{\tfrac{q\beta}{M} + \beta} \cdot \tfrac{L \|z^{0} - z^* \|^2}{\varepsilon });
$\\
$\bullet$ in non-monotone case:
$
    \mathcal{O}([\tfrac{q\beta}{M} + \beta] \tfrac{L^2 \|z^{0} - z^* \|^2}{\varepsilon^2}).
$
\end{corollary}
We can see that \algname{MASHA1} can outperform the uncompressed Extra Gradient method. Let us compare them in the strongly monotone case. The communication complexity of the Extra Gradient method is $\mathcal{\tilde O} ( \nicefrac{L}{\mu} )$. \algname{MASHA1} has communication complexity $\mathcal{\tilde O} ( \sqrt{\nicefrac{q}{\beta M} + \nicefrac{1}{\beta}} \cdot \nicefrac{L}{\mu} )$. For practical compressors \cite{beznosikov2020biased}, $\beta \geq q$. \add{Then, one can note that the communication complexity of \algname{MASHA1} differs from the complexity of the uncompressed method by an additional factor $( \sqrt{\nicefrac{1}{ M} + \nicefrac{1}{\beta}})$. It is easy to see that even for a small number of devices $M$ and expected density $\beta$, this factor is less than $1$, hence \algname{MASHA1} outperforms the uncompressed method. We think that this factor $( \sqrt{\nicefrac{1}{ M} + \nicefrac{1}{\beta}} )$ is theoretically unimprovable and optimal -- see Section \ref{sec:opt} for details.}

 One can also consider the case of bidirectional compression ($q^{\text{serv}} \neq 1$). Table \ref{tab:comparison1} (line 3) shows the result for $q^{\text{serv}} = q^{\text{dev}}_m = q$, $\beta^{\text{serv}} = \beta^{\text{dev}}_m = \beta$ and $1 - \tau = \nicefrac{1}{\beta}$.

\subsection{\algname{MASHA2}: Handling Contractive Compressors} \label{sec:masha2}

The use of contractive compressions is a more complex issue. In particular, it is known that if one simply put a contractive cospressor instead of an unbiased one, the method may diverge even for quadratic problems \cite{beznosikov2020biased}. To fix this, an error compensation technique \cite{stich2018sparsified,error_feedback,stich2019error} is used. The point of this approach is to keep untransmitted information and add it to a new package at the next iteration.
This is the main difference between \algname{MASHA2} and \algname{MASHA1}. \algname{MASHA2} introduces additional sequences $e^k$, $e^k_m$ for the server's and devices' error. To define contractive operators on devices and on the server, we introduce the following notation: $C^{\text{dev}}_m, \delta^{\text{dev}}, \beta^{\text{dev}}$ and $C^{\text{serv}}_m, \delta^{\text{serv}}, \beta^{\text{serv}}$. 

\begin{algorithm}[h]
	\caption{\algname{MASHA2}}
	\label{main_alg_qfb}
	\begin{algorithmic}
\State
\noindent {\bf Parameters:}  Stepsize $\gamma>0$, parameter $\tau$, number of iterations $K$.\\
\noindent {\bf Initialization:} Choose  $z^0 = w^0 \in \mathcal{Z}$, $e^0_m = 0$, $e^0 = 0$. \\
Devices send $F_m(w^0)$ to server and get $F(w^0)$
\For {$k=0,1, 2, \ldots, K-1$ }
\For {each device $m$ in parallel}
\State $z^{k+1/2} = \tau z^k + (1 - \tau) w^k - \gamma F(w^k)$ 
\State Sends $g^k_m = C^{\text{dev}}_m(\gamma F_m(z^{k+1/2}) - \gamma F_m(w^k) + e^k_m)$ to server
\State $e^{k+1}_m = e^k_m + \gamma F_m(z^{k+1/2}) - \gamma F_m(w^k) - g^k_m$
\EndFor
\For {server}
\State Sends to devices $g^k = C^{\text{serv}} \left [\frac{1}{M} \sum_{m=1}^M g^k_m + e^k\right] $
\State $e^{k+1} = e^k + \frac{1}{M} \sum_{m=1}^M g^k_m - g^k$
\State Sends to devices one bit $b_k$ : 1 with probability $1 - \tau$, 0 with with probability $\tau$
\EndFor
\For {each device $m$ in parallel}
\State  $z^{k+1} = z^{k+1/2} - \gamma  g^k$ 
\State If {$b_k = 1$} then $w^{k+1} = z^{k}$, 
sends $F_m(w^{k+1})$ to server and gets $F(w^{k+1})$
\State else $w^{k+1} = w^k$  
\EndFor
\EndFor
	\end{algorithmic}
\end{algorithm}

In the case of \algname{MASHA1}, the key theoretical issue was the choice of a basic method (we discussed this at the beginning of Section \ref{sec:masha1_main}). \algname{MASHA2} raises another problem for theoretical analysis, how to combine \algname{MASHA1} and the error feedback technique. The analysis of methods with error compensation for the minimization problem $\min_x f(x)$ is entirely tied to the existence of the function $f$ \cite{stich2019error, qian2020error, EF21}. In particular, the differences $(f(\cdot) - f(x^*))$ appear in the whole analysis and is key in the technical lemmas. As a result $(f(\cdot) - f(x^*))$ is used as a convergence criterion even in the strongly convex case.
But for VIs there is no function $f$, only the operator $F$ (the existence of $g(x,y)$ in SPP setup does not save the situation). This problem is solved in the proof of Theorem \ref{main_th_comp} by using an additional sequence $\|z^{k+1/2} - w^k \|$.
\begin{theorem} \label{main_th_comp}
Let Assumption \ref{as3} and one case of Assumption \ref{as1} are satisfied. Then for some step $\gamma$ the following estimates on \algname{MASHA2} number of iterations to achieve $\varepsilon$-solution holds 

$\bullet$ in strongly monotone case (in terms of 
$
\EE[\| \hat z^{K} - z^* \|^2 ] \sim \varepsilon
$):
$ 
\mathcal{O}([\frac{1}{1 - \tau} + \frac{\delta^{\text{dev}} \delta^{\text{serv}} \tilde L}{\mu \sqrt{1-\tau}}]\log \frac{1}{\varepsilon});
$

$\bullet$ in monotone case ($
\EE\max_{z \in \mathcal{C}} [ \langle F(u),  (\tfrac{1}{K}\sum_{k=0}^{K-1} z^{k+1/2}) - u  \rangle] \sim \varepsilon
$):
$
\mathcal{O}(\frac{\delta^{\text{dev}} \delta^{\text{serv}} \tilde L\|z^{0} - z^* \|^2}{\varepsilon \sqrt{1-\tau}});
$

$\bullet$ in non-monotone case (in terms of 
$
    \EE[\tfrac{1}{K}\sum_{k=0}^{K-1}\| F(w^k)\|^2] \sim \varepsilon^2
$):
$
    \mathcal{O}(\frac{(\delta^{\text{dev}} \delta^{\text{serv}})^2 \tilde L^2 \|z^{0} - z^* \|^2}{\varepsilon^2 (1 - \tau)}).
$
\end{theorem}
A full listing of the algorithm, as well as a full statement of the theorem with proof, can be found in  Appendix \ref{sec:MASHA2}.

The same way as in Section \ref{sec:masha1_main} we can consider only devices' or bidirectional compression.
In particular, in the line 2 of Table \ref{tab:comparison1} we put results for $\delta^{\text{serv}} = 1$, $\delta^{\text{dev}} = \delta$, $L_m = \tilde L = L$ and $1-\tau = \beta$. In the line 4 of Table \ref{tab:comparison1} there are results for $\delta^{\text{serv}} = \delta^{\text{dev}} = \delta$, $L_m = \tilde L = L$ and $1-\tau = \beta$.

\section{Experiments}

\subsection{Bilinear Saddle Point Problem}\label{sect:exp_bilinear}

We start our experiments with a distributed bilinear problem, i.e. the problem \eqref{eq:minmax} with 
\begin{align}
    \label{bilinear}
     g_m(x,y) \eqdef  x^\top A_m y + a^\top_m x + b^\top_m y + \frac{\lambda}{2} \| x\|^2 -  \frac{\lambda}{2} \|y\|^2,
\end{align}
 where $A_m \in \R^{d \times d}$, $a_m, b_m \in \R^d$. This problem is $\lambda$-strongly convex--strongly-concave and, moreover, all functions $g_m$ are $\|A_m \|_2$-smooth. Therefore, such a distributed problem is well suited for the primary comparison of our methods. We take $d=100$ and generate positive definite matrices $A_m$ and vectors $a_m, b_m$ randomly, $\lambda$ is chosen as $\max_m \|A_m\|_2 / 10^5$.

\begin{wrapfigure}[11]{r}{10cm}
\vspace{-0.7cm}
\caption{Comparison \algname{MASHA1} (Algorithm \ref{main_alg_qeg}) and \algname{MASHA2} (Algorithm \ref{main_alg_qfb}) with Error Feedback, QGD and Compressed Extra Gradient (\algname{CEG}) in iterations and in Mbytes for \eqref{bilinear}.}
\label{masha_exp1}
\centering
\vspace{-0.3cm}
\begin{minipage}[b]{0.7\textwidth}
\centering
\includegraphics[width=0.45\textwidth]{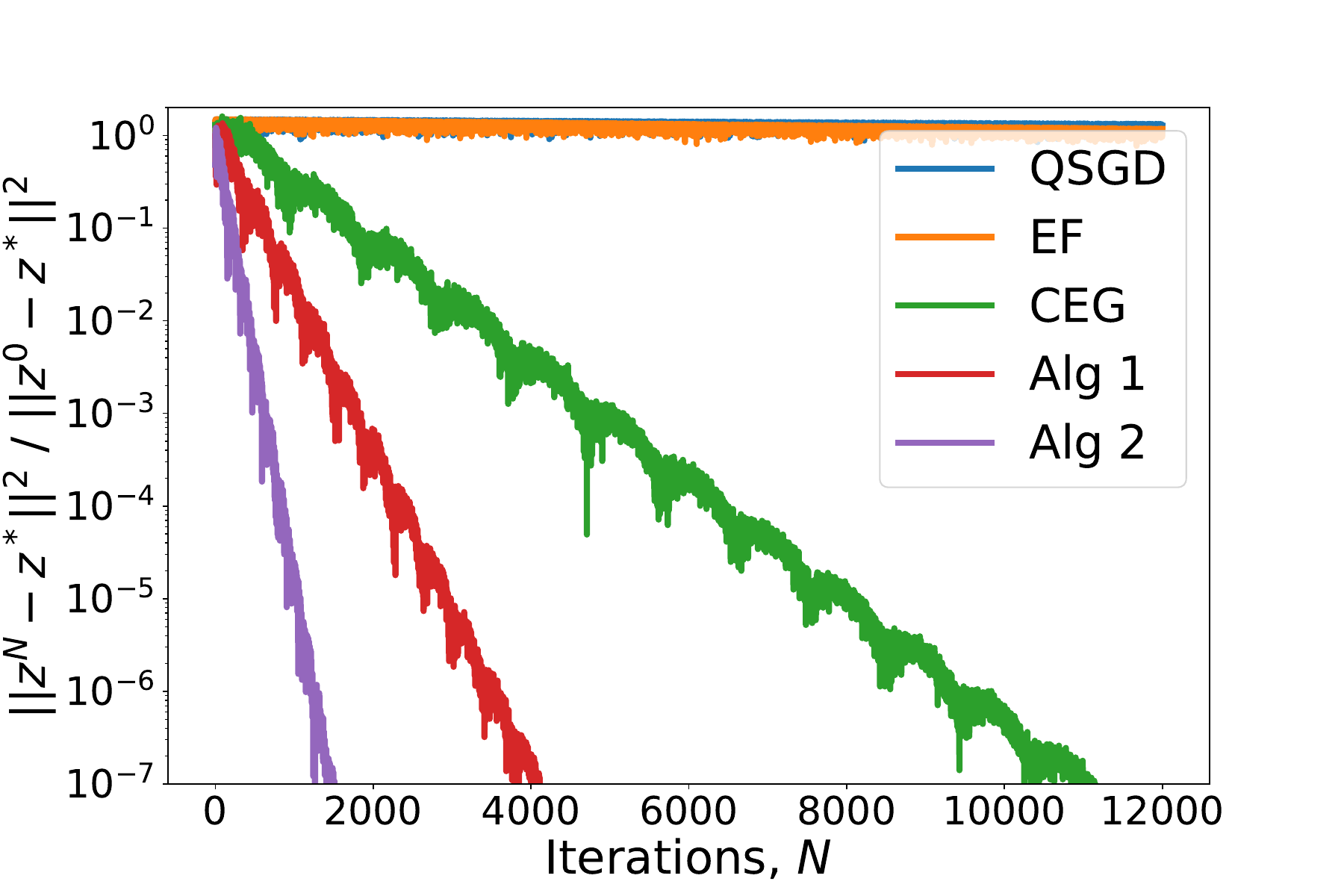}
\includegraphics[width=0.45\textwidth]{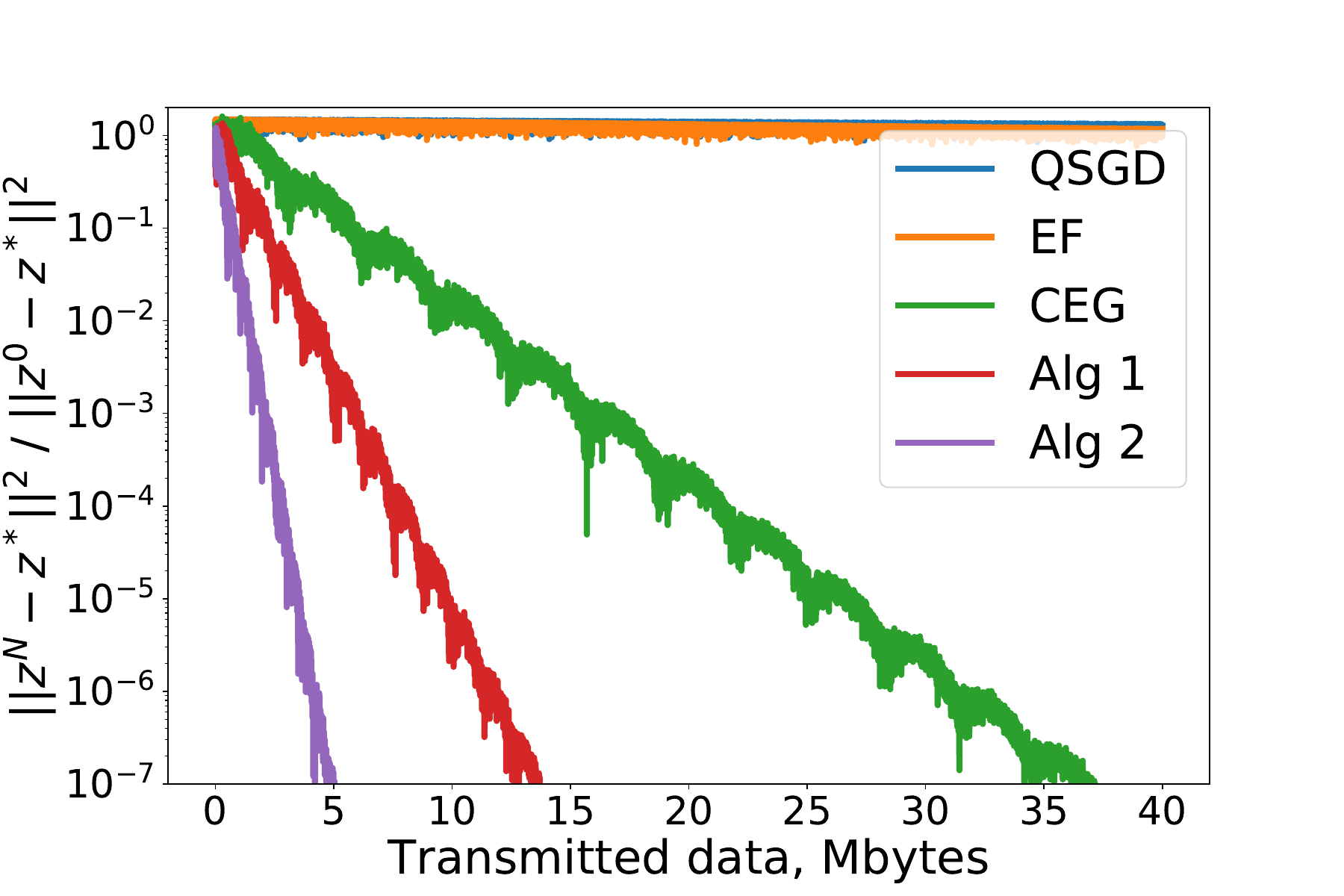}
\end{minipage}
\end{wrapfigure}

 The purpose of the experiment is to understand whether the \algname{MASHA1} and \algname{MASHA2} methods are superior to those in the literature. As a comparison, we take  QGD \cite{alistarh2017qsgd} with Random 30\%, classical Error Feedback \cite{stich2018sparsified} with Top 30\% compression, as well as \algname{CEG} (Section \ref{sec:ceg}) -- Compressed Extra Gradient, each step of which we use Random 30\%. In \algname{MASHA1} (Algorithm \ref{main_alg_qeg}) we also used Random 30\%, in \algname{MASHA2} (Algorithm \ref{main_alg_qfb}) -- Top 30\%. See Figure \ref{masha_exp1}. The stepsizes of all methods are chosen for best convergence.
 
We see on Figure \ref{masha_exp1} that methods based on gradient descent (QSGD and EF) converge slowly. This confirms that one needs to use method specifically designed  for saddle point problems (for example, the extragradient method), and not classical optimization methods. The much slower convergence of \algname{CEG} shows the efficiency of our approach in which we compress the differences $F_m(z^{k+1/2}) - F(w^k)$. \algname{MASHA2} wins \algname{MASHA1}. This shows that in practice a contractive compressor can perform better than an unbiased one with the same parameters.


\subsection{Adversarial Training of Transformers}\label{sect:exp_adv}

We now evaluate how compression performs for variational inequalities (and for saddle point problems, as a special case) in a more practically motivated scenario. 
Indeed, saddle point problems (special case of variational inequalities) have sample applications in machine learning, including {\em adversarial training}. And our goal is to show that compression provides important improvements for such large-scale problems as well. We train a {\em transformer-based masked language model}~\cite{transformer,bert,roberta} using a fleet of 16 low-cost preemptible workers with T4 GPU and low-bandwidth interconnect. For this task, we use the compute-efficient adversarial training regimen proposed for transformers by~\cite{zhu2019freelb,liu2020adversarial}. Formally, the adversarial formulation of the problem is the min-max problem
\begin{align*}
  \min \limits_{w} \max\limits_{\|\rho_n\|\leq e} \frac{1}{N} \sum\limits_{n=1}^N l(f(w, x_n + \rho_n, y_n)^2 + \frac{\lambda}{2} \| w\|^2,
\end{align*}
where $w$ are the weights of the model, $\{(x_n, y_n)\}_{n=1}^N$ are pairs of the training data, $\rho$ is the so-called adversarial noise which introduces a  perturbation in the data, and   $\lambda$ are the regularization parameters. To make our setup more realistic, we train ALBERT-large with layer sharing~\cite{albert}, which was recently shown to be much more communication-efficient during training~\cite{moshpit,dedloc}. We train our model on a combination of Bookcorpus and Wikipedia datasets with the same optimizer (LAMB) and parameters as in the original paper~\citep{albert}, use the adversarial training configuration of~\cite{zhu2019freelb}, and follow system design considerations for preemptible instances~\cite{moshpit}. In LAMB optimizer we change the original positive momentum to negative momentum, as in \algname{MASHA}. This means that we do not exactly use \algname{MASHA} in these experiments, but a combination of \algname{MASHA} and LAMB. In fact this approach is typical, e.g., in papers \cite{daskalakis2018training,gidel2018a,mertikopoulos2018optimistic,chavdarova2019reducing,pmlr-v89-liang19b}, the theoretical methods are combined with Adam.

In terms of communication, we consider 4 different setups for gradient compression: the ``baseline'' strategy with uncompressed gradients, full 8-bit quantization~\citep{compression_dettmers,compression_deep}, mixed 8-bit quantization, and Power compression~\citep{powersgd} with rank $r{=}8$. For mixed 8-bit quantization and Power  we only apply compression to gradient tensors with more than $2^{16}$ elements, sending smaller ones uncompressed. These small tensors represent layer biases and LayerNorm scales~\citep{layernorm} that collectively amount to $\leq1\%$ of the total gradient, but can be more difficult to compress than regular weight tensors. Finally, since Power is a biased compression algorithm, we use error feedback~\citep{error_feedback, EF21} with a modified formulation proposed by~\cite{powersgd}. For all experimental setups, we report learning curves in terms of the model training objective, similarly to~\citep{switch,moshpit}. To quantify the differences in training loss better, we also evaluate the downstream performance for each model on several popular tasks from~\citep{glue} after each model was trained on approximately 80 billion tokens. Finally, we measure the communication efficiency of each proposed strategy by measuring the average wall time per communication round when all 16 workers are active.

\begin{wrapfigure}[21]{r}{10cm}
\centering
\vspace{-0.5cm}
\captionof{figure}{\textbf{(upper left)} ALBERT training objective convergence rate with different compression algorithms; \textbf{(upper right)} ALBERT training objective convergence rate with different compression algorithms (zoomed); \textbf{(lower)}  Average wall time per communication round with standard deviation over 5 repetitions and downstream evaluation scores on GLUE benchmark tasks after at 80 billion training tokens (${\approx}10^4$ optimizer steps).}
\label{fig:albert}
\begin{minipage}[][][b]{0.7\textwidth}
\centering
\includegraphics[width=0.45\textwidth]{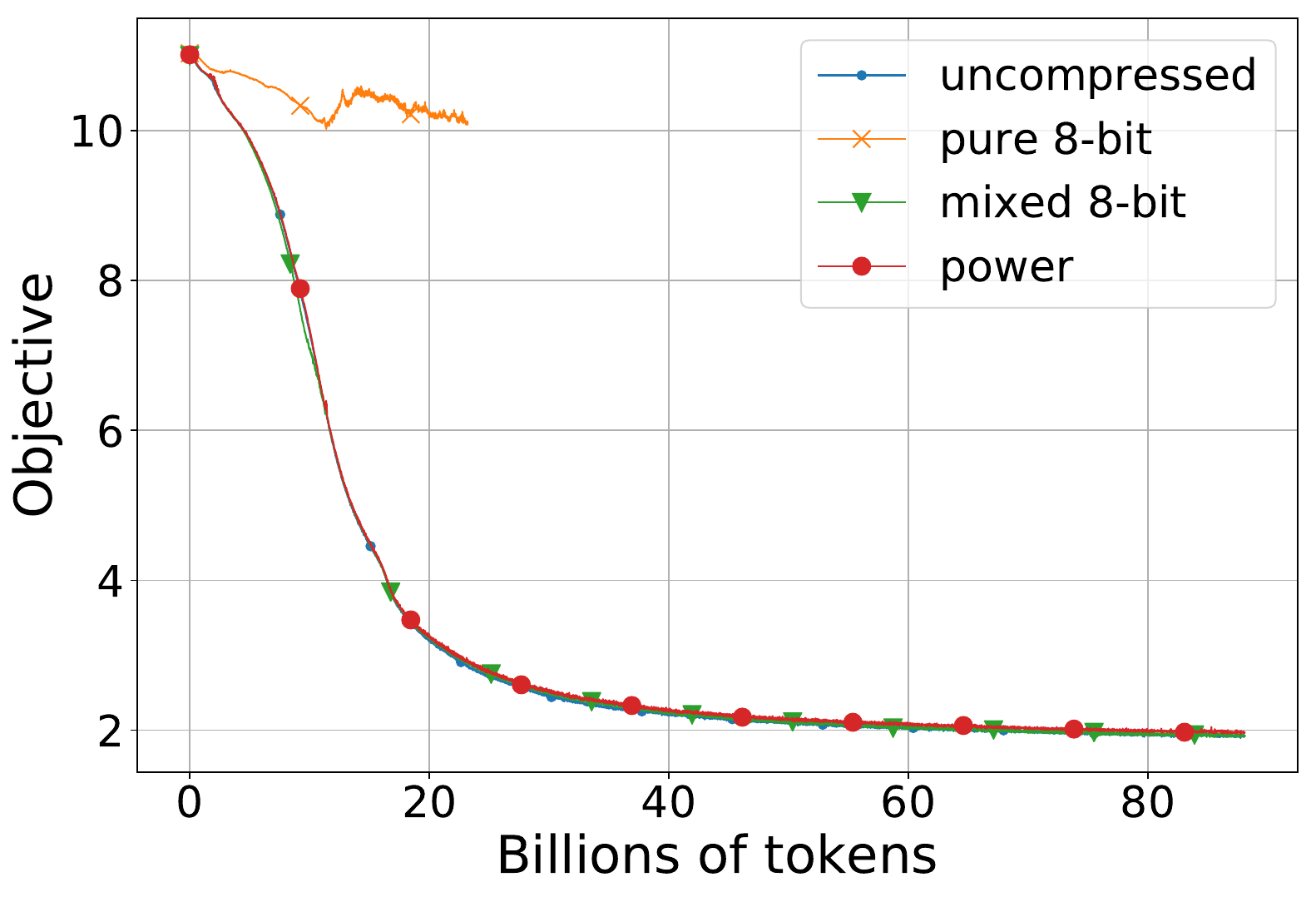}
\includegraphics[width=0.45\textwidth]{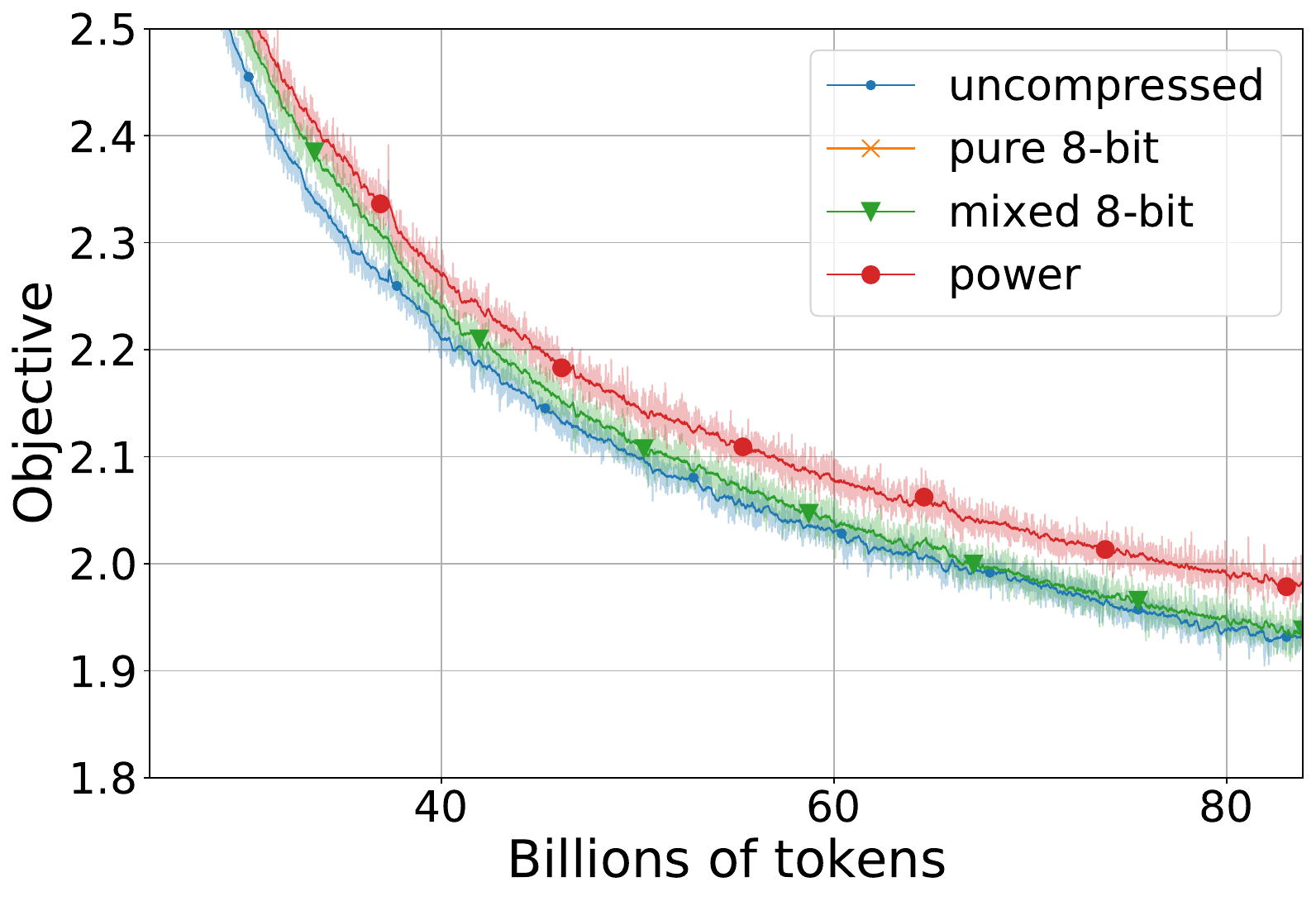}
\end{minipage}
\begin{minipage}{0.7\textwidth}
\centering
\renewcommand{\arraystretch}{1.0}
\renewcommand{\tabcolsep}{2pt}
\scriptsize
\begin{tabular}{l|c|c|c|c|c|c|c|c|c|c}
\toprule 
Setup & Avg time & CoLA & MNLI & MRPC & QNLI & QQP & RTE & SST2 & STS-B & WNLI \\ 
\midrule
Baseline & 8.79 $\pm$  0.03 &  45.2 & 81.1 &  83.0 & {88.3} & \textbf{89.0} & \textbf{67.8} & 85.5 & \textbf{89.4} & \textbf{18.3}   \\
Full 8-bit & 4.42 $\pm$ 0.07 &  N/A  & N/A  & N/A  & N/A  & N/A  & N/A  & N/A  & N/A  & N/A   \\
Mixed 8-bit & 4.61 $\pm$ 0.08 & \textbf{48.8} & \textbf{81.3} & \textbf{88.7} & 88.1 & 85.2 & 64.3 & 88.3 & 87.5 & 16.9  \\
Power &  \textbf{1.57 $\pm$ 0.05} & 43.9 & 80.5 & 85.6 & \textbf{88.6} & 86.0  & 47.2 & \textbf{88.5} & 88.5 & 16.9  \\
\bottomrule
\end{tabular}
\end{minipage}
\end{wrapfigure}

The learning curves in Figure \ref{fig:albert} (upper) follow a predictable pattern, with more extreme compression techniques  demonstrating  slower  per-iteration  convergence.   One  curious  exception  to  that is full 8-bit quantization, which was unable to achieve competitive training loss. The remaining three setups converge to similar loss values below 2.  Both the baseline and mixed 8-bit compression show similar values in terms of downstream performance, with Power compression showing mild degradation. But in terms of information transfer time, methods using compression (especially Power) are significantly superior to the method without compression. This makes it possible to use such techniques to increase the training time without sacrificing quality.

\section{Conclusion} \label{sec:concl}

\add{In this paper we present algorithms with unbiased and contractive compressions for solving distributed VIs and SPPs. Our algorithms are presented in deterministic, stochastic and federated versions. All basic algorithms and their modifications support bidirectional compression. Experiments confirm the efficiency of both our algorithms and the use of compression for solving large-scale VIs in general.

In future works it is important to address the issue of the necessity to forward uncompressed information in some iterations. Although full packages are rarely transmitted, this is a slight limitation of our approach.
Lower bounds for compression methods are also an interesting area of research. At the moment there are neither such results for VIs and SPPs, nor for minimizations. In Appendix \ref{sec:opt} we only hypothesize the optimality of our methods and back it up with analogies, provable lower estimates could complete the story with compressed methods.}

\section*{Acknowledgments}

This research of A. Beznosikov has been supported by The Analytical Center for the Government of the Russian Federation (Agreement No. 70-2021-00143 dd. 01.11.2021, IGK 000000D730321P5Q0002).


\bibliography{ltr}
\bibliographystyle{plain}

\newpage
\appendix
\onecolumn
\part*{APPENDIX}

\tableofcontents


\newpage

\section{Table with summary our results} \label{sec:table}

\renewcommand{\arraystretch}{1.6}
\begin{table*}[h]
    \centering
    \scriptsize
	\caption{ Summary of our \textbf{iteration complexity} results for finding an $\varepsilon$-solution for  problem \eqref{VI} in the deterministic (i.e., \eqref{MK}) with only device compression, deterministic with bidirectional (device-server) compression, stochastic (i.e., \eqref{MK}+\eqref{MK_fs}) and federated learning/partial participation (i.e., \eqref{MK}+\eqref{pp_setup}) setups. In the strongly-monotone (strongly convex - strongly convex) case, convergence is measured by the distance to the solution. In the monotone(convex-concave) case, convergence is measured in terms of the gap function \eqref{gap}. In non-monotone (non-convex-non-concave) case convergence is measured in terms of the norm of the operator.  {\em Notation:} $\mu$ = constant of strong monotonicity of the operator $F$, $L$ = maximum of local Lipschitz constants $L_m$,  $R$ = diameter (in Euclidean norm) of the optimization set, $R_0$ = initial distance to the solution, $\myred{q}$ = the variance parameter associated with an unbiased compressor (see \eqref{quant}); $\myblue{\delta} $ = the variance parameter associated with a contractive compressor (see \eqref{compr}); $\myred{\beta},\myblue{\beta}$ = expected density (the number of times the operator compresses information); $M$ = the number of parallel clients/nodes; $r$ = the size of the local dataset (see \eqref{MK_fs}); $b$ = the number of clients in Partial Participation (FL) setup. We have results with bidirectional compression also in stochastic and federated setups, but to simplify the bounds, we present bidirectional results only in the deterministic setup. 
	}
    \label{tab:comparison1}
    \vspace{-0.3cm}
\resizebox{\linewidth}{!}{
    \begin{tabular}{|c|c|c|c|c|}
    \cline{3-5}
    \multicolumn{2}{c|}{}
     & Strongly monotone  & Monotone & Non monotone \\
    \hline
    \multirow{4}{*}{\rotatebox[origin=c]{90}{Deter. (Device)}\rotatebox[origin=c]{90}{\eqref{VI} + \eqref{MK}}} 
    & \algnametiny{MASHA1}  & \multirow{2}{*}{$  \mathcal{\tilde O}\left(\myred{\beta} + \sqrt{\myred{\beta} +  \frac{\myred{q\beta}}{M}} \cdot \frac{L}{\mu}  \right)$} 
    & \multirow{2}{*}{$ \mathcal{ O}\left(\sqrt{\myred{\beta} +  \frac{\myred{q\beta}}{M}} \cdot \frac{L R^2}{ \varepsilon}\right)$} & \multirow{2}{*}{$ \mathcal{ O}\left(\left( \myred{\beta}+ \frac{\myred{q\beta }}{M}\right)\cdot \frac{L^2R^2}{\varepsilon^2}\right)$} \\ 
    & Alg \ref{main_alg_qeg} ~~Cor \ref{main_cor1_q} & & &
    \\ \cline{2-5}
    & \algnametiny{MASHA2}  & \multirow{2}{*}{$  \mathcal{\tilde O}\left(\myblue{\beta} + \myblue{\delta}\sqrt{\myblue{\beta}} \cdot \frac{L}{\mu}  \right)$} 
    & \multirow{2}{*}{$  \mathcal{O}\left( \myblue{\delta}\sqrt{\myblue{\beta}} \cdot \frac{L R^2}{ \varepsilon} \right)$} & \multirow{2}{*}{$  \mathcal{O}\left( \myblue{\delta^2}\myblue{\beta} \cdot \frac{L^2 R^2}{ \varepsilon^2} \right)$} \\ 
    & Alg \ref{main_alg_qfb} ~~Cor \ref{cor:masha2}& & & \\ 
    \hline 
    \multirow{4}{*}{\rotatebox[origin=c]{90}{Deter. (Bidirect.)}\rotatebox[origin=c]{90}{\eqref{VI} + \eqref{MK}}} 
    & \algnametiny{MASHA1}  & \multirow{2}{*}{$  \mathcal{\tilde O}\left(\myred{\beta} + \sqrt{\myred{q\beta} +  \frac{\myred{q^2\beta}}{M}} \cdot \frac{L}{\mu}  \right)$} 
    & \multirow{2}{*}{$ \mathcal{ O}\left(\sqrt{\myred{q\beta} +  \frac{\myred{q^2\beta}}{M}} \cdot \frac{L R^2}{ \varepsilon}\right)$} & \multirow{2}{*}{$ \mathcal{ O}\left(\left( \myred{q\beta}+ \frac{\myred{q^2\beta }}{M}\right)\cdot \frac{L^2R^2}{\varepsilon^2}\right)$} \\ 
    & Alg \ref{main_alg_qeg} ~~Cor \ref{cor:bi_masha1}& & &
    \\ \cline{2-5}
    & \algnametiny{MASHA2}  & \multirow{2}{*}{$  \mathcal{\tilde O}\left(\myblue{\beta} + \myblue{\delta^2}\sqrt{\myblue{\beta}} \cdot \frac{L}{\mu}  \right)$} 
    & \multirow{2}{*}{$  \mathcal{O}\left( \myblue{\delta^2}\sqrt{\myblue{\beta}} \cdot \frac{L R^2}{ \varepsilon} \right)$} & \multirow{2}{*}{$  \mathcal{O}\left( \myblue{\delta^4}\myblue{\beta} \cdot \frac{L^2 R^2}{ \varepsilon^2} \right)$} \\ 
    & Alg \ref{main_alg_qfb} ~~Cor \ref{cor:bi_masha2}& & & \\ 
    \hline  
    \multirow{4}{*}{\rotatebox[origin=c]{90}{Stoch. (F-S)}\rotatebox[origin=c]{90}{\eqref{VI} + \eqref{MK} + \eqref{MK_fs}}} 
    & \algnametiny{VR-MASHA1}  & \multirow{2}{*}{\tiny{$  \mathcal{\tilde O}\left(\myred{\beta} + r + \max\{\sqrt{\myred{\beta}};\sqrt{r}\}\sqrt{1 +  \frac{\myred{q}}{M}} \cdot \frac{L}{\mu}  \right)$}}
    & \multirow{2}{*}{\tiny{$ \mathcal{ O}\left(\max\{\sqrt{\myred{\beta}};\sqrt{r}\}\sqrt{1 +  \frac{\myred{q}}{M}} \cdot \frac{L R^2}{ \varepsilon}\right)$}} & \multirow{2}{*}{\tiny{$ \mathcal{ O}\left(\left( \max\{\myred{\beta};r\}\left(1 +  \frac{\myred{q}}{M}\right)\right)\cdot \frac{L^2R^2}{\varepsilon^2}\right)$}} \\ 
    & Alg \ref{alg_qeg_vr} ~~Cor \ref{cor:vr_masha1}& & &
    \\ \cline{2-5}
    & \algnametiny{VR-MASHA2}  & \multirow{2}{*}{$  \mathcal{\tilde O}\left(\myblue{\beta} + r + \max\{\sqrt{\myblue{\beta}};\sqrt{r}\} \myblue{\delta} \cdot \frac{L}{\mu}  \right)$} 
    & \multirow{2}{*}{$  \mathcal{O}\left( \max\{\sqrt{\myblue{\beta}};\sqrt{r}\} \myblue{\delta} \cdot \frac{L R^2}{ \varepsilon} \right)$} & \multirow{2}{*}{$  \mathcal{O}\left( \max\{\myblue{\beta};r\} \myblue{\delta^2} \cdot \frac{L^2 R^2}{ \varepsilon^2} \right)$} \\ 
    & Alg \ref{alg_qfbvr} ~~Cor \ref{cor:vr_masha2}& & & \\ 
    \hline 
    \multirow{4}{*}{\rotatebox[origin=c]{90}{FL (PP)}\rotatebox[origin=c]{90}{\eqref{VI} + \eqref{MK}+ \eqref{pp_setup}} } 
    & \algnametiny{PP-MASHA1}  & \multirow{2}{*}{$  \mathcal{\tilde O}\left(\frac{\myred{\beta}M}{b} + \sqrt{\frac{\myred{\beta}M}{b} +  \frac{\myred{q\beta }M}{b}} \cdot \frac{L}{\mu}  \right)$} 
    & \multirow{2}{*}{$ \mathcal{ O}\left(\sqrt{\frac{\myred{\beta}M}{b} +  \frac{\myred{q\beta }M}{b}}  \cdot \frac{L R^2}{ \varepsilon}\right)$} & \multirow{2}{*}{$ \mathcal{ O}\left(\left( \frac{\myred{\beta}M}{b} +  \frac{\myred{q\beta }M}{b} \right)\cdot \frac{L^2R^2}{\varepsilon^2}\right)$} \\ 
    & Alg \ref{alg_qeg_pp} ~~Cor \ref{cor:pp_masha1}& & &
    \\ \cline{2-5}
    & \algnametiny{PP-MASHA2}  & \multirow{2}{*}{$  \mathcal{\tilde O}\left(\frac{\myblue{\beta}M}{b} + \myblue{\delta}\sqrt{\frac{\myblue{\beta}M^3}{b^3}} \cdot \frac{L}{\mu}  \right)$} 
    & \multirow{2}{*}{$  \mathcal{O}\left( \myblue{\delta}\sqrt{\frac{\myblue{\beta}M^3}{b^3}} \cdot \frac{L R^2}{ \varepsilon} \right)$} & \multirow{2}{*}{$  \mathcal{O}\left( \myblue{\delta^2 \beta}\frac{M^3}{b^3} \cdot \frac{L^2 R^2}{ \varepsilon^2} \right)$} \\ 
    & Alg \ref{alg_qfb_pp} ~~Cor \ref{cor:pp_masha2}& & & \\ 
    \hline 
    \end{tabular}
    }
    \vspace{-0.3cm}
\end{table*}

\newpage

\section{Optimality of \algname{MASHA1} and \algname{MASHA2}} \label{sec:opt}

In this section, we discuss why the \algname{MASHA1} and \algname{MASHA2} convergence estimates cannot be improved (it means that the methods are optimal). We emphasise that this is only a hypothesis based on some analogies. As irrefutable proof we could use lower bounds, but there are no such lower bounds even for minimization problems (despite their wide research in the community).
The following considerations are also outlined in Table \ref{tab:comparison2}.

We consider the strongly convex/strongly monotone case.
For the deterministic minimization problem, lower and optimal upper bounds are given in \cite{nesterov2018lectures}. These bounds are $\mathcal{\tilde O}\left( \sqrt{\nicefrac{L}{\mu}} \right)$. Meanwhile, methods with unbiased (ADIANA \cite{pmlr-v119-li20g}) and contractive (ECLK \cite{qian2020error}) compression, but without compression, are also optimal for the deterministic minimization problem. 
Iteration complexity of ADIANA with compression is $\mathcal{\tilde O}\left(\sqrt{\nicefrac{q\beta}{M} + \beta} \cdot \sqrt{\nicefrac{L}{\mu} }\right)$. For ECLK complexities in iterations is $\mathcal{\tilde O}\left(\delta\sqrt{\beta} \cdot \sqrt{\nicefrac{L}{\mu} }\right)$. Note the interesting feature that the compression dependent multiplier can be improved, but only with a loss in the $\nicefrac{L}{\mu}$-multiplier. For example, DIANA \cite{DIANA} (unbiased) has $\mathcal{\tilde O}\left((\nicefrac{q}{M} + 1) \cdot \nicefrac{L}{\mu} \right)$ iteration complexity, or EF \cite{stich2019error} (contractive) has $\mathcal{\tilde O}\left(\delta \cdot \nicefrac{L}{\mu}\right)$ iteration complexity. \algname{MASHA1} and \algname{MASHA2} without compressions are optimal for Lipschitz continuous strongly monotone VIs \cite{zhang2019lower} and have a deterministic bound $\mathcal{\tilde O}\left( \nicefrac{L}{\mu} \right)$. \algname{MASHA1} and \algname{MASHA2} have the same compression dependency multipliers as ADIANA and ECLK. This suggests that the dependence of \algname{MASHA1} and \algname{MASHA2} on compression properties cannot be improved for variational inequalities without loss in $\nicefrac{L}{\mu}$.
In Section \ref{sec:ceg}, we prove the convergence of \algname{CEG} with unbiased compression, which achieves $\mathcal{\tilde O}\left((\nicefrac{q}{M} + 1) \cdot \nicefrac{L^2}{\mu^2} \right)$ iteration complexity.

As another argument, let us give an example of the situation with the VR approach (finite sum problem) for minimization problems and for VIs. For minimization, the lower bounds in the smooth strongly convex case are $\mathcal{\tilde O}\left(r + \sqrt{r \nicefrac{L}{\mu}} \right)$ \cite{woodworth2016tight}. The optimal method is \cite{allen2017katyusha}. SVRG \cite{NIPS2013_ac1dd209} has estimates $\mathcal{\tilde O}\left( r + \nicefrac{L}{\mu} \right)$ (better in $r$, worse in $\nicefrac{L}{\mu}$). What about variational inequalities? The lower bounds in the Lipschitz continuous strongly convex case are $\mathcal{\tilde O}\left(r + \sqrt{r} \nicefrac{L}{\mu} \right)$ \cite{han2021lower}. The optimal methods are \cite{alacaoglu2021stochastic}. Methods from \cite{palaniappan2016stochastic} have estimates $\mathcal{\tilde O}\left( r + \nicefrac{L^2}{\mu^2} \right)$. Following this logic, estimates for ADIANA and ECLK are transformed into estimates for \algname{MASHA1} and  \algname{MASHA2}.

The same situation with estimates is in the convex/monotone case.

\renewcommand{\arraystretch}{1.6}
\begin{table*}[h]
    \centering
    \scriptsize
	\caption{ Summary of iteration complexity results for minimization problems and variational inequalities in different setups: deterministic, stochastic, distributed with biased and contractive compressions. 
	{\em Notation:} $\mu$ = constant of strong convexity/monotonicity, $L$ = Lipschitz constant of the gradient/operator,  $q$ = the variance parameter associated with an unbiased compressor; $\delta $ = the variance parameter associated with a contractive compressor; $\beta$ = expected density (the number of times the operator compresses information); $M$ = the number of parallel clients/nodes; $r$ = the size of the local dataset. 
	}
    \label{tab:comparison2}
    \vspace{-0.1cm}
 \resizebox{\linewidth}{!}{
    \begin{tabular}{|c|c|c|c|c|c|c|c|c|}
    \cline{2-9}
    \multicolumn{1}{c|}{}
     & \multicolumn{2}{c|}{Deterministic}  & \multicolumn{3}{c|}{Stochastic (VR)} & \multicolumn{3}{c|}{Unbiased compression} \\
    \hline
    & Lower & Upper & Lower & Upper 1 & Upper 2 & Lower & Upper 1 & Upper 2\\
    \hline
    Minimization & $\sqrt{\frac{L}{\mu}}$ \cite{nesterov2018lectures} & $\sqrt{\frac{L}{\mu}}$ \cite{nesterov2018lectures} & $\sqrt{r} \cdot \sqrt{\frac{L}{\mu}}$ \cite{woodworth2016tight} & $\sqrt{r} \cdot \sqrt{\frac{L}{\mu}}$ \cite{allen2017katyusha} & $r +\frac{L}{\mu}$ \cite{NIPS2013_ac1dd209} & -- & $\sqrt{\beta + \frac{q\beta}{M}} \cdot \sqrt{\frac{L}{\mu}}$ \cite{pmlr-v119-li20g} & $\left(1 + \frac{q}{M}\right)\cdot \frac{L}{\mu}$  \cite{DIANA} \\
    \hline 
    VI/SPP & $\frac{L}{\mu}$ \cite{zhang2019lower} & $\frac{L}{\mu}$ \cite{gidel2018a} & $\sqrt{r} \cdot \frac{L}{\mu}$ \cite{han2021lower} & $\sqrt{r} \cdot \frac{L}{\mu}$ \cite{alacaoglu2021stochastic} & $r +\frac{L^2}{\mu^2}$ \cite{palaniappan2016stochastic} & -- & $\sqrt{\beta + \frac{q\beta}{M}} \cdot \frac{L}{\mu}$ (\textbf{Ours}) & $ \left(1 + \frac{q}{M}\right)\cdot \frac{L^2}{\mu^2}$ (\textbf{Ours})\\
    \hline
    \multicolumn{6}{c|}{}  & \multicolumn{3}{c|}{Contractive compression} \\
    \cline{7-9}
    \multicolumn{6}{c|}{} & Lower & Upper 1 & Upper 2\\
   \cline{7-9}
    \multicolumn{6}{c|}{} & -- & $\delta \sqrt{\beta} \cdot \sqrt{\frac{L}{\mu}}$ \cite{qian2020error} & $\delta \cdot \frac{L}{\mu}$ \cite{stich2019error} \\
    \cline{7-9} 
    \multicolumn{6}{c|}{}  & -- & $\delta \sqrt{\beta} \cdot \frac{L}{\mu}$ (\textbf{Ours})& --\\
    \cline{7-9}
    \end{tabular}
    }
    \vspace{-0.3cm}
\end{table*}

\newpage 
\section{Basic Facts}
\textbf{Upper bound for a squared sum.} For arbitrary integer $n\ge 1$ and arbitrary set of vectors $a_1,\ldots,a_n$ we have
\begin{equation}
    \left(\sum\limits_{i=1}^n a_i\right)^2 \le m\sum\limits_{i=1}^n a_i^2\label{eq:sqs}
\end{equation}

\section{\algnamenormal{MASHA1}: Handling Unbiased Compressors} \label{sec:masha1}

In this section, we provide additional information about Algorithm \ref{main_alg_qeg} -- \algname{MASHA1}. We give a full form of \algname{MASHA1} -- see Algorithm \ref{alg_qeg}. 

\begin{algorithm}[th]
	\caption{(Algorithm \ref{main_alg_qeg}) \algname{MASHA1}}
	\label{alg_qeg}
	\begin{algorithmic}[1]
\State
\noindent {\bf Parameters:}  Stepsize $\gamma>0$, parameter $\tau$, number of iterations $K$.\\
\noindent {\bf Initialization:} Choose  $z^0 = w^0 \in \mathcal{Z}$. \\
Server  sends to devices $z^0 = w^0$ and devices compute $F_m(w^0)$ and send to server and get $F(w^0)$
\For {$k=0,1, 2, \ldots, K-1$ }
\For {each device $m$ in parallel}
\State $\bar z^k = \tau z^k + (1 - \tau) w^k$ \label{alg1_bzk}
\State $z^{k+1/2} = \bar z_k - \gamma F(w^k)$ \label{alg1_zk1/2}
\State  Compute $F_m(z^{k+1/2})$ \& send  $Q^{\text{dev}}_m(F_m(z^{k+1/2}) - F_m(w^k))$ to server
\EndFor
\For {server}
\State Compute $Q^{\text{serv}} \left [\frac{1}{M} \sum \limits_{m=1}^M Q^{\text{dev}}_m(F_m(z^{k+1/2}) - F_m(w^k))\right]$ \& send to devices
\State Sends to devices one bit $b_k$: 1 with probability $1 - \tau$, 0 with with probability $\tau$ \label{alg1:1bit}
\EndFor
\For {each device $m$ in parallel}
\State  $z^{k+1} = z^{k+1/2} - \gamma  Q^{\text{serv}}\left[\frac{1}{M} \sum\limits_{m=1}^M Q^{\text{dev}}_m(F_m(z^{k+1/2}) - F_m(w^k))\right]$ \label{alg1_q_zk1}
\If {$b_k = 1$}
\State $w^{k+1} = z^{k}$ \label{alg1:wk_zk}
\State Compute $F_m(w^{k+1})$ \& send it to server; and get $F(w^{k+1})$ as a response from server
\Else 
\State $w^{k+1} = w^k$  \label{alg1:wk_wk}
\EndIf
\EndFor
\EndFor
	\end{algorithmic}
\end{algorithm}

The following theorem gives the convergence of \algname{MASHA1}.

\begin{theorem}[Theorem \ref{main_th_q}] \label{th_q}
Let distributed variational inequality \eqref{VI} + \eqref{MK} is solved by Algorithm \ref{alg_qeg} with unbiased compressor operators \eqref{quant}: on server with $q^{\text{serv}}$ parameter, on devices with $\{q^{\text{dev}}_m\}$. Let Assumption \ref{as3} and one case of Assumption \ref{as1} are satisfied. Then the following estimates holds 

$\bullet$ in strongly-monotone case with $\gamma \leq \min\left[\frac{\sqrt{1 - \tau}}{2 C_q}; \frac{1 - \tau}{2\mu}\right]$ \\(where $C_q = \sqrt{\frac{q^{\text{serv}}}{M^2} \sum_{m=1}^M (q_m^{\text{dev}} L_m^2 + (M-1) \tilde L^2)}$):
\begin{align*}
 \EE\left(\| z^{K} - z^* \|^2 + \|w^{K} - z^* \|^2\right) &\leq \left( 1 -\frac{\mu\gamma}{2} \right)^K \cdot 2\|z^{0} - z^* \|^2;
\end{align*}

$\bullet$ in monotone case with $\gamma \leq \frac{\sqrt{1 - \tau}}{2 C_q + 4 \tilde L}$:
\begin{align*}
\EE\left[\max_{z \in \mathcal{C}} \left[ \langle F(u),  \left(\frac{1}{K}\sum\limits_{k=0}^{K-1} z^{k+1/2}\right) - u  \rangle \right] \right]
\leq  \frac{2\max_{z \in \mathcal{C}}\left[\| z^0 - z \|^2\right] + 6\|z^{0} - z^* \|^2}{\gamma K};
\end{align*}

$\bullet$ in non-monotone case with $\gamma \leq \frac{\sqrt{1 - \tau}}{2 C_q}$:
\begin{align*}
    \EE\left(\frac{1}{K}\sum\limits_{k=0}^{K-1}\| F(w^k)\|^2\right)
    &\leq \frac{16\|z^{0} - z^* \|^2}{\gamma^2 K}.
\end{align*}
\end{theorem}

For the monotone case, we use the {\em gap function} as convergence criterion:
\begin{equation}
    \label{gap}
    \text{Gap} (z) \eqdef \sup_{u \in \mathcal{C}} \left[ \langle F(u),  z - u  \rangle \right].
\end{equation}
Here we do not take the maximum over the entire set $\R^d$ (as in the classical version), but over $\mathcal{C}$ -- a compact subset of $\R^d$. Thus, we can also consider unbounded sets $\R^d$. This is permissible, since such a version of the criterion is valid if the solution $z^{*}$ lies in $\mathcal{C}$; for details see the work of \cite{nesterov2007dual}.

Let us move on to the choice of $\tau$.

Let us start with the only devices compression, i.e. it is assumed that server-side compression is not required, because broadcasts from the server are cheap. As noted in the main part of the paper, then we  consider only sendings from devices to the server.  Note that the following expression $\sum_{m=1}^M (q_m^{\text{dev}} L_m^2 + (M-1) \tilde L^2)$ occurs in $C_q$. It means that we can choose $q^{\text{dev}}_m$ depending on $L_m$. Let us define $L_{\min} = \min_m L_m$ and $q_l = q$ for $l = \arg \min_m L_m$. If one put $q_m = q L_{\min} / L_m$, then we get $C_q =  \sqrt{\frac{1}{M} (q L_{\min}^2 + (M-1) \tilde L^2)}$.

At each iteration, the device sends to the server $\mathcal{O}\left( \tfrac{1}{M}\sum_{m=1}^M \frac{1}{\beta_m} + (1-\tau)\right)$ bits -- each time information compressed by $\beta_m$ (for device $m$) times and with probability $1-\tau$ the full package. Then the optimal choice $\tau$ is $1 - \tfrac{1}{\beta}$ with $\tfrac{1}{\beta} = \tfrac{1}{M}\sum_{m=1}^M \tfrac{1}{\beta_m}$.

\begin{corollary} [Corollary \ref{main_cor1_q}] \label{cor1_q}
Let distributed variational inequality \eqref{VI} + \eqref{MK} is solved by Algorithm \ref{alg_qeg} without compression on server ($q^{\text{serv}} = 1$) and with unbiased compressor operators \eqref{quant} on devices with $\{q^{\text{dev}}_m\}$ (as described in the previous paragraphs). Let Assumption \ref{as3} and one case of Assumption \ref{as1} are satisfied. Then the following estimates holds 

$\bullet$ in strongly-monotone case with $\gamma \leq \min\left[\frac{1}{2} \cdot \left(\sqrt{\frac{q \beta L^2_{\min}}{M} + \beta \tilde L^2 } \right)^{-1}; \frac{1}{2\mu \beta}\right]$:
\begin{align*}
 \EE\left(\| z^{K} - z^* \|^2 + \|w^{K} - z^* \|^2\right) &\leq \left( 1 -\frac{\mu\gamma}{2} \right)^K \cdot 2\|z^{0} - z^* \|^2;
\end{align*}

$\bullet$ in monotone case with $\gamma \leq \frac{1}{6} \cdot \left(\sqrt{\frac{q \beta L^2_{\min}}{M} + \beta \tilde L^2 } \right)^{-1}$:
\begin{align*}
\EE\left[\max_{z \in \mathcal{C}} \left[ \langle F(u),  \left(\frac{1}{K}\sum\limits_{k=0}^{K-1} z^{k+1/2}\right) - u  \rangle \right] \right]
\leq  \frac{2\max_{z \in \mathcal{C}}\left[\| z^0 - z \|^2\right] + 6\|z^{0} - z^* \|^2}{\gamma K};
\end{align*}

$\bullet$ in non-monotone case with $\gamma \leq \frac{1}{2} \cdot \left(\sqrt{\frac{q \beta L^2_{\min}}{M} + \beta \tilde L^2 } \right)^{-1}$:
\begin{align*}
    \EE\left(\frac{1}{K}\sum\limits_{k=0}^{K-1}\| F(w^k)\|^2\right)
    &\leq \frac{16\|z^{0} - z^* \|^2}{\gamma^2 K}.
\end{align*}
\end{corollary}
In the line 1 of  Table \ref{tab:comparison1} we put complexities to achieve $\varepsilon$-solution. For simplicity, we put $Q^{\text{dev}}_m = Q$ with $q^{\text{dev}}_m = q$ and $\beta^{\text{dev}}_m = \beta$, also $L_m = \tilde L = L$.

Next, we add server compression. Now the transfer from the server is important. Here and after, for simplicity, we put $Q^{\text{serv}} = Q^{\text{dev}}_m = Q$ with $q^{\text{dev}}_m = q$ and $\beta^{\text{dev}}_m = \beta$, also $L_m = \tilde L = L$.  One can also analyze the case with different $q_m$ and $L_m$, as is done in Corollary \ref{cor1_q}.

At each iteration, the device is sent to the server and the server to devices $\mathcal{O}\left( \frac{1}{\beta} + 1-\tau\right)$ bits. Then the optimal choice $\tau$ is still $1 - \frac{1}{\beta}$.

\begin{corollary} \label{cor:bi_masha1}
Let distributed variational inequality \eqref{VI} + \eqref{MK} is solved by Algorithm \ref{alg_qeg} with unbiased compressor operators \eqref{quant}: on server with $q^{\text{serv}} = q$ parameter, on devices with $\{q^{\text{dev}}_m = q\}$. Let Assumption \ref{as3} and one case of Assumption \ref{as1} are satisfied. Then the following estimates holds 

$\bullet$ in strongly-monotone case with $\gamma \leq \min\left[\frac{1}{2 L} \cdot \left(\sqrt{\frac{q^2 \beta}{M} + \beta } \right)^{-1}; \frac{1}{2\mu \beta}\right]$:
\begin{align*}
 \EE\left(\| z^{K} - z^* \|^2 + \|w^{K} - z^* \|^2\right) &\leq \left( 1 -\frac{\mu\gamma}{2} \right)^K \cdot 2\|z^{0} - z^* \|^2;
\end{align*}

$\bullet$ in monotone case with $\gamma \leq \frac{1}{6L} \cdot \left(\sqrt{\frac{q^2 \beta}{M} + \beta } \right)^{-1}$:
\begin{align*}
\EE\left[\max_{z \in \mathcal{C}} \left[ \langle F(u),  \left(\frac{1}{K}\sum\limits_{k=0}^{K-1} z^{k+1/2}\right) - u  \rangle \right] \right]
\leq  \frac{2\max_{z \in \mathcal{C}}\left[\| z^0 - z \|^2\right] + 6\|z^{0} - z^* \|^2}{\gamma K};
\end{align*}

$\bullet$ in non-monotone case with $\gamma \leq \frac{1}{2 L} \cdot \left(\sqrt{\frac{q^2 \beta}{M} + \beta } \right)^{-1}$:
\begin{align*}
    \EE\left(\frac{1}{K}\sum\limits_{k=0}^{K-1}\| F(w^k)\|^2\right)
    &\leq \frac{16\|z^{0} - z^* \|^2}{\gamma^2 K}.
\end{align*}
\end{corollary}
In the line 3 of  Table \ref{tab:comparison1} we put complexities to achieve $\varepsilon$-solution.

\subsection{Proof of the convergence of \algnamenormal{MASHA1}} \label{prof_th1}

\textbf{Proof of Theorem \ref{th_q}:} We start from the following equalities for any $z$:
\begin{align*}
    \|z^{k+1} - z \|^2
    &=
    \|z^{k+1/2} - z \|^2 + 2 \la  z^{k+1} -  z^{k+1/2},  z^{k+1/2} - z\ra + \|  z^{k+1} -  z^{k+1/2}\|^2,
\end{align*}
\begin{align*}
    \|z^{k+1/2} - z \|^2
    &=
    \| z^{k} - z \|^2 + 2 \la z^{k+1/2} -   z^{k}, z^{k+1/2} - z\ra - \|  z^{k+1/2} - z^{k}\|^2.
\end{align*}
Then we sum two inequalities:
\begin{align}
\label{t5650}
\|z^{k+1} - z \|^2
    &=
    \|z^{k} - z \|^2 + 2 \la  z^{k+1} -  z^{k},  z^{k+1/2} - z\ra  \nonumber\\
    &\hspace{0.4cm}+ \| z^{k+1} -  z^{k+1/2}\|^2 - \| z^{k+1/2} - z^{k}\|^2.
\end{align}
Using lines \ref{alg1_bzk}, \ref{alg1_zk1/2}, \ref{alg1_q_zk1}, we get
\begin{align*}
\|z^{k+1} - z \|^2
    &=
    \|z^{k} - z \|^2 \nonumber\\
    &\hspace{0.4cm}
    - 2 \la  \gamma  Q^{\text{serv}}\left[\frac{1}{M} \sum\limits_{m=1}^M Q^{\text{dev}}_m(F_m(z^{k+1/2}) - F_m(w^k))\right] + \gamma F(w^k),  z^{k+1/2} - z\ra  \nonumber\\
    &\hspace{0.4cm} + 2 \la \tau z^k + (1-\tau) w^k -  z^{k},  z^{k+1/2} - z\ra \nonumber\\
    &\hspace{0.4cm}+ \| z^{k+1} -  z^{k+1/2}\|^2 - \| z^{k+1/2} - z^{k}\|^2 \nonumber\\
    &=
    \|z^{k} - z \|^2 \nonumber\\
    &\hspace{0.4cm}- 2 \la  \gamma  Q^{\text{serv}}\left[\frac{1}{M} \sum\limits_{m=1}^M Q^{\text{dev}}_m(F_m(z^{k+1/2}) - F_m(w^k))\right] + \gamma F(w^k),  z^{k+1/2} - z\ra  \nonumber\\
    &\hspace{0.4cm} + 2 (1-\tau) \la w^k -  z^{k},  z^{k+1/2} - z\ra \nonumber\\
    &\hspace{0.4cm}+ \| z^{k+1} -  z^{k+1/2}\|^2 - \| z^{k+1/2} - z^{k}\|^2.
\end{align*}
The equality $2\la a,b \ra = \|a+b \|^2 - \| a\|^2 - \| b\|^2$ gives 
\begin{align}
\label{t437}
\|z^{k+1} - z \|^2
    &=
    \|z^{k} - z \|^2 \nonumber\\
    &\hspace{0.4cm}- 2 \la  \gamma  Q^{\text{serv}}\left[\frac{1}{M} \sum\limits_{m=1}^M Q^{\text{dev}}_m(F_m(z^{k+1/2}) - F_m(w^k))\right] + \gamma F(w^k),  z^{k+1/2} - z\ra  \nonumber\\
    &\hspace{0.4cm} + 2 (1-\tau) \la w^k -  z^{k+1/2},  z^{k+1/2} - z\ra + 2 (1-\tau) \la z^{k+1/2} -  z^{k},  z^{k+1/2} - z\ra \nonumber\\
    &\hspace{0.4cm}+ \| z^{k+1} -  z^{k+1/2}\|^2 - \| z^{k+1/2} - z^{k}\|^2 \nonumber\\
    &=
    \|z^{k} - z \|^2 \nonumber\\
    &\hspace{0.4cm}- 2 \la  \gamma  Q^{\text{serv}}\left[\frac{1}{M} \sum\limits_{m=1}^M Q^{\text{dev}}_m(F_m(z^{k+1/2}) - F_m(w^k))\right] + \gamma F(w^k),  z^{k+1/2} - z\ra  \nonumber\\
    &\hspace{0.4cm}+  (1-\tau)\| w^k - z \|^2 -  (1-\tau)\| w^k - z^{k+1/2} \|^2 -   (1-\tau)\| z^{k+1/2} - z\|^2 \nonumber\\
    &\hspace{0.4cm}+ (1-\tau)\| z^{k+1/2}  -  z^k\|^2 +(1-\tau) \| z^{k+1/2} - z \|^2 - (1-\tau) \| z^{k} -z\|^2 \nonumber\\
    &\hspace{0.4cm}+ \| z^{k+1} -  z^{k+1/2}\|^2 - \| z^{k+1/2} - z^{k}\|^2 \nonumber\\
    &=
    \tau \|z^{k} - z \|^2 +  (1-\tau)\| w^k - z \|^2 \nonumber\\
    &\hspace{0.4cm}- 2 \la  \gamma  Q^{\text{serv}}\left[\frac{1}{M} \sum\limits_{m=1}^M Q^{\text{dev}}_m(F_m(z^{k+1/2}) - F_m(w^k))\right] + \gamma F(w^k),  z^{k+1/2} - z\ra  \nonumber\\
    &\hspace{0.4cm}+ \| z^{k+1} -  z^{k+1/2}\|^2 -  (1-\tau)\| w^k - z^{k+1/2} \|^2  - \tau \| z^{k+1/2} - z^{k}\|^2.
\end{align}

We now consider the three cases of monotonicity separately.

\subsubsection{Strongly-monotone case}

Let substitute $z = z^*$, take full mathematical expectation and get
\begin{align*}
\EE\|z^{k+1} - z^* \|^2
    &=
    \tau \EE\|z^{k} - z^* \|^2 +  (1-\tau)\EE\| w^k - z^* \|^2 \nonumber\\
    &\hspace{0.4cm}- 2\gamma  \EE \left[\la   Q^{\text{serv}}\left[\frac{1}{M} \sum\limits_{m=1}^M Q^{\text{dev}}_m(F_m(z^{k+1/2}) - F_m(w^k))\right] + F(w^k),  z^{k+1/2} - z^*\ra\right]  \nonumber\\
    &\hspace{0.4cm}+ \EE\| z^{k+1} -  z^{k+1/2}\|^2 -  (1-\tau)\EE\| w^k - z^{k+1/2} \|^2  - \tau \EE\| z^{k+1/2} - z^{k}\|^2.
\end{align*}
With unbiasedness \eqref{quant} we have
\begin{align}
\label{t56501}
\EE\|z^{k+1} - z^* \|^2
    &=
    \tau \EE\|z^{k} - z^* \|^2 +  (1-\tau)\EE\| w^k - z^* \|^2 \nonumber\\
    &\hspace{0.4cm}- 2\gamma  \EE \left[\la   \EE_{Q^{\text{serv}},Q^{\text{dev}}} \left[Q^{\text{serv}}\left[\frac{1}{M} \sum\limits_{m=1}^M Q^{\text{dev}}_m(F_m(z^{k+1/2}) - F_m(w^k))\right] + F(w^k)\right],  z^{k+1/2} - z^*\ra\right]  \nonumber\\
    &\hspace{0.4cm}+ \EE\| z^{k+1} -  z^{k+1/2}\|^2 -  (1-\tau)\EE\| w^k - z^{k+1/2} \|^2  - \tau \EE\| z^{k+1/2} - z^{k}\|^2 \nonumber\\
    &=
    \tau \EE\|z^{k} - z^* \|^2 +  (1-\tau)\EE\| w^k - z^* \|^2 \nonumber\\
    &\hspace{0.4cm}- 2\gamma  \EE \left[\la   F(z^{k+1/2}),  z^{k+1/2} - z^*\ra\right]  \nonumber\\
    &\hspace{0.4cm}+ \EE\| z^{k+1} -  z^{k+1/2}\|^2 -  (1-\tau)\EE\| w^k - z^{k+1/2} \|^2  - \tau \EE\| z^{k+1/2} - z^{k}\|^2.
\end{align}
Let us work with $\EE\left[\| z^{k+1} -  z^{k+1/2}\|^2\right]$, with \eqref{quant} we get
\begin{align*}
\EE\left[\| z^{k+1} -  z^{k+1/2}\|^2\right] &= \gamma^2 \cdot \EE\left[\left\| Q^{\text{serv}} \left [\frac{1}{M} \sum_{m=1}^M Q^{\text{dev}}_m(F_m(z^{k+1/2}) - F_m(w^k))\right]\right\|^2\right]  \\
&\leq \gamma^2 \cdot\frac{q^{\text{serv}}}{M^2} \EE\left[ \left\| \sum_{m=1}^M Q^{\text{dev}}_m(F_m(z^{k+1/2}) - F_m(w^k))\right\|^2\right]  \\
&= \gamma^2 \cdot \frac{q^{\text{serv}}}{M^2} \sum_{m=1}^M \EE\left[\left \|  Q^{\text{dev}}_m(F_m(z^{k+1/2}) - F_m(w^k))\right\|^2\right] \\
&\hspace{0.4cm}+ \gamma^2 \cdot \frac{q^{\text{serv}}}{M^2} \sum_{m \neq l}\EE \left[\langle Q^{\text{dev}}_m (F_m(z^{k+1/2}) - F_m(w^k)) ; Q^{\text{dev}}_l (F_l(z^{k+1/2}) - F_l(w^k)) \rangle \right]
\end{align*}
Next we apply \eqref{quant} and Assumption \ref{as3} for the first term and independence and unbiasedness of $Q$ for the second term:
\begin{align}
\label{temp33}
\EE\left[\| z^{k+1} -  z^{k+1/2}\|^2\right] &\leq \gamma^2 \cdot\frac{q^{\text{serv}}}{M^2} \sum_{m=1}^M q_m^{\text{dev}} L_m^2 \EE\left[\left \|  z^{k+1/2} - w^k\right\|^2\right] \nonumber\\
&\hspace{0.4cm}+ \gamma^2 \cdot\frac{q^{\text{serv}}}{M^2} \sum_{m \neq l}\EE \left[\langle F_m(z^{k+1/2}) - F_m(w^k) ; F_l(z^{k+1/2}) - F_l(w^k) \rangle \right] \nonumber\\
&\leq \gamma^2 \cdot\frac{q^{\text{serv}}}{M^2} \sum_{m=1}^M q_m^{\text{dev}} L_m^2 \EE\left[\left \|  z^{k+1/2} - w^k\right\|^2\right] \nonumber\\
&\hspace{0.4cm}+ \gamma^2 \cdot\frac{q^{\text{serv}}}{2M^2} \sum_{m \neq l}\EE \left[\| F_m(z^{k+1/2}) - F_m(w^k) \|^2 +  \| F_l(z^{k+1/2}) - F_l(w^k) \|^2 \right] \nonumber\\
&\leq \gamma^2 \cdot\frac{q^{\text{serv}}}{M^2} \sum_{m=1}^M q_m^{\text{dev}} L_m^2 \EE\left[\left \|  z^{k+1/2} - w^k\right\|^2\right] \nonumber\\
&\hspace{0.4cm}+ \gamma^2 \cdot\frac{q^{\text{serv}}}{2M^2} \sum_{m \neq l}  \EE \left[L^2_m \| z^{k+1/2} - w^k\|^2 +  L_l^2\| z^{k+1/2} - w^k \|^2 \right] \nonumber\\
&= \gamma^2 \cdot\frac{q^{\text{serv}}}{M^2} \sum_{m=1}^M q_m^{\text{dev}} L_m^2 \EE\left[\left \|  z^{k+1/2} - w^k\right\|^2\right] \nonumber\\
&\hspace{0.4cm}
+ \gamma^2 \cdot\frac{q^{\text{serv}} (M-1)}{M}   \tilde L^2 \EE \left[\| z^{k+1/2} - w^k\|^2\right]
\nonumber\\
&= \gamma^2 \cdot\frac{q^{\text{serv}}}{M^2} \EE \left[\| z^{k+1/2} - w^k\|^2\right] \cdot \sum_{m=1}^M q_m^{\text{dev}} L_m^2 + (M-1) \tilde L^2
\end{align}
Let us define new constant $C_q = \sqrt{\frac{q^{\text{serv}}}{M^2} \sum_{m=1}^M (q_m^{\text{dev}} L_m^2 + (M-1) \tilde L^2)}$ and then connect \eqref{t56501} and \eqref{temp33}:
\begin{align}
\label{t565012}
\EE\|z^{k+1} - z^* \|^2
    &\leq
    \tau \EE\|z^{k} - z^* \|^2 +  (1-\tau)\EE\| w^k - z^* \|^2 \nonumber\\
    &\hspace{0.4cm}- 2\gamma  \EE \left[\la   F(z^{k+1/2}),  z^{k+1/2} - z^*\ra\right]  \nonumber\\
    &\hspace{0.4cm} -  (1-\tau - \gamma^2 C_q^2)\EE\| w^k - z^{k+1/2} \|^2  - \tau \EE\| z^{k+1/2} - z^{k}\|^2.
\end{align}
Then we use choice of $w^{k + 1}$ (lines \ref{alg1:1bit}, \ref{alg1:wk_zk}, \ref{alg1:wk_wk}) and get
\begin{align}
\label{twk1}
\EE\|w^{k+1}  - z^*\|^2 = \EE\left[ \EE_{w^{k+1}}\|w^{k+1}  - z^*\|^2\right] = \tau \EE\left\|w^{k} - z^*\right\|^2 + (1 - \tau) \EE\|z^{k} - z^*\|^2,
\end{align}
Summing \eqref{t565012} and \eqref{twk1}, we obtain
\begin{align}
\label{t565013}
\EE\|z^{k+1} - z^* \|^2 &+ \EE\|w^{k+1}  - z^*\|^2 \nonumber\\
    &\leq
    \EE\|z^{k} - z^* \|^2 +  \EE\| w^k - z^* \|^2 \nonumber\\
    &\hspace{0.4cm}- 2\gamma  \EE \left[\la   F(z^{k+1/2}),  z^{k+1/2} - z^*\ra\right]  \nonumber\\
    &\hspace{0.4cm} -  (1-\tau - \gamma^2 C_q^2)\EE\| w^k - z^{k+1/2} \|^2  - \tau \EE\| z^{k+1/2} - z^{k}\|^2.
\end{align}
The property of the solution \eqref{VI} gives
\begin{align*}
\EE\|z^{k+1} - z^* \|^2 &+ \EE\|w^{k+1}  - z^*\|^2 \nonumber\\
    &\leq
    \EE\|z^{k} - z^* \|^2 +  \EE\| w^k - z^* \|^2 \nonumber\\
    &\hspace{0.4cm}- 2\gamma  \EE \left[\la   F(z^{k+1/2}) - F(z^*),  z^{k+1/2} - z^*\ra\right]  \nonumber\\
    &\hspace{0.4cm} -  (1-\tau - \gamma^2 C_q^2)\EE\| w^k - z^{k+1/2} \|^2  - \tau \EE\| z^{k+1/2} - z^{k}\|^2.
\end{align*}
And by Assumption \ref{as1} in strong monotone case we have 
\begin{align*}
\EE\|z^{k+1} - z^* \|^2 &+ \EE\|w^{k+1}  - z^*\|^2 \nonumber\\
    &\leq
    \EE\|z^{k} - z^* \|^2 +  \EE\| w^k - z^* \|^2 - 2\gamma\mu  \EE\| z^{k+1/2} - z^*\|^2 \nonumber\\
    &\hspace{0.4cm} -  (1-\tau - \gamma^2 C_q^2)\EE\| w^k - z^{k+1/2} \|^2  - \tau \EE\| z^{k+1/2} - z^{k}\|^2.
\end{align*}
With $-\| a\|^2 \leq - \frac{1}{2} \|a+b \|^2 + \|b \|^2$ we deduce:
\begin{align}
\label{tt5657}
\EE(\|z^{k+1} - z^* \|^2 &+ \EE\|w^{k+1}  - z^*\|^2) \nonumber\\
    &\leq
    \left(1 - \frac{\mu\gamma}{2}\right)\EE\left(\|z^{k} - z^* \|^2 + \| w^k - z^* \|^2\right) \nonumber\\
    &\hspace{0.4cm} -  (1-\tau - \mu\gamma - \gamma^2 C_q^2)\EE\| w^k - z^{k+1/2} \|^2  - (\tau- \mu\gamma) \EE\| z^{k+1/2} - z^{k}\|^2.
\end{align}
It remains only to choose $\gamma \leq \min\left\{ \frac{\sqrt{1 - \tau}}{2 C_q}; \frac{1 - \tau}{2\mu}\right\}$ and get
\begin{align*}
\EE(\|z^{k+1} - z^* \|^2 + \EE\|w^{k+1}  - z^*\|^2)
&\leq \left( 1 -\frac{\mu\gamma}{2} \right) \cdot \EE\left(\|z^{k} - z^* \|^2 + \| w^k - z^* \|^2\right) .
\end{align*}
Running the recursion completes the proof.
\EndProof

\subsubsection{Monotone case} 

We start from \eqref{t437}:
\begin{align*}
2\gamma &\la F(z^{k+1/2}), z^{k+1/2} - z\ra
\\
 &= \tau \|z^{k} - z \|^2 - \|z^{k+1} - z \|^2 +  (1-\tau)\| w^k - z \|^2 \nonumber\\
    &\hspace{0.4cm}- 2 \gamma \la    Q^{\text{serv}}\left[\frac{1}{M} \sum\limits_{m=1}^M Q^{\text{dev}}_m(F_m(z^{k+1/2}) - F_m(w^k))\right] + F(w^k) - F(z^{k+1/2}),  z^{k+1/2} - z\ra  \nonumber\\
    &\hspace{0.4cm}+ \| z^{k+1} -  z^{k+1/2}\|^2 -  (1-\tau)\| w^k - z^{k+1/2} \|^2  - \tau \| z^{k+1/2} - z^{k}\|^2.
\end{align*}
Adding both sides $\| w^{k+1} - z\|^2 $ and making small rearrangement we have
\begin{align*}
2\gamma &\la F(z^{k+1/2}), z^{k+1/2} - z\ra  \\
&\leq \left[\| z^k - z \|^2 + \| w^k - z \|^2\right] - \left[\| z^{k+1} - z\|^2 + \| w^{k+1} - z\|^2\right]  \notag\\
&\hspace{0.4cm}-\tau \| w^k - z \|^2 - (1 - \tau) \| z^{k} - z\|^2 + \| w^{k+1} - z\|^2 \notag\\
&\hspace{0.4cm} - 2\gamma \langle Q^{\text{serv}}\left[\frac{1}{M} \sum\limits_{m=1}^M Q^{\text{dev}}_m(F_m(z^{k+1/2}) - F_m(w^k))\right] + F(w^k) - F(z^{k+1/2}),  z^{k+1/2} - z  \rangle \notag\\
&\hspace{0.4cm}- \tau \| z^{k+1/2} - z^k\|^2 - (1 -\tau) \| z^{k+1/2} - w^k \|^2 + \| z^{k+1} -  z^{k+1/2}\|^2.
\end{align*}
Then we sum up over $k = 0, \ldots, K - 1$, take maximum of both sides over $z \in \mathcal{C}$, after take expectation and get
\begin{align*}
2\gamma \cdot &\EE\left[\max_{z \in \mathcal{C}} \sum\limits_{k=0}^{K-1}\la F(z^{k+1/2}), z^{k+1/2} - z\ra \right]\leq \max_{z \in \mathcal{C}}\left[\| z^0 - z \|^2 + \| w^0 - z \|^2\right]  \notag\\
&\hspace{0.4cm}+\EE\left[\max_{z \in \mathcal{C}} \sum\limits_{k=0}^{K-1} \left[ -\tau \| w^k - z \|^2 - (1 - \tau) \| z^{k} - z\|^2 + \| w^{k+1} - z\|^2\right] \right] \notag\\
&\hspace{0.4cm}- \sum\limits_{k=0}^{K-1} \left[\tau \EE\left[\| z^{k+1/2} - z^k\|^2 \right] + (1 -\tau) \EE\left[\| z^{k+1/2} - w^k \|^2 \right] -  \EE\left[\| z^{k+1} -  z^{k+1/2}\|^2 \right]\right]\notag\\ 
& \hspace{0.4cm}+ 2\gamma \EE\left[\max_{z \in \mathcal{C}} \sum\limits_{k=0}^{K-1} \left[ \langle Q^{\text{serv}}\left[\frac{1}{M} \sum\limits_{m=1}^M Q^{\text{dev}}_m(F_m(z^{k+1/2}) - F_m(w^k))\right] + F(w^k) - F(z^{k+1/2}),  z - z^{k+1/2} \rangle \right] \right].
\end{align*}
Applying \eqref{temp33} for $\EE\left[\| z^{k+1} -  z^{k+1/2}\right]$, we get
\begin{align*}
2\gamma \cdot &\EE\left[\max_{z \in \mathcal{C}} \sum\limits_{k=0}^{K-1}\la F(z^{k+1/2}), z^{k+1/2} - z\ra \right]\leq \max_{z \in \mathcal{C}}\left[\| z^0 - z \|^2 + \| w^0 - z \|^2\right]  \notag\\
&\hspace{0.4cm}+\EE\left[\max_{z \in \mathcal{C}} \sum\limits_{k=0}^{K-1} \left[ -\tau \| w^k - z \|^2 - (1 - \tau) \| z^{k} - z\|^2 + \| w^{k+1} - z\|^2\right] \right] \notag\\
&\hspace{0.4cm}- \sum\limits_{k=0}^{K-1} \left[\tau \EE\left[\| z^{k+1/2} - z^k\|^2 \right] + (1 -\tau - \gamma^2 C_q^2) \EE\left[\| z^{k+1/2} - w^k \|^2 \right] \right]\notag\\ 
& \hspace{0.4cm}+ 2\gamma \EE\left[\max_{z \in \mathcal{C}} \sum\limits_{k=0}^{K-1} \left[ \langle Q^{\text{serv}}\left[\frac{1}{M} \sum\limits_{m=1}^M Q^{\text{dev}}_m(F_m(z^{k+1/2}) - F_m(w^k))\right] + F(w^k) - F(z^{k+1/2}),  z - z^{k+1/2} \rangle \right] \right].
\end{align*}
With $\gamma \leq \frac{\sqrt{1 -\tau}}{2C_q}$ we get
\begin{align}
\label{temp4041}
2\gamma \cdot &\EE\left[\max_{z \in \mathcal{C}} \sum\limits_{k=0}^{K-1}\la F(z^{k+1/2}), z^{k+1/2} - z\ra \right]\leq \max_{z \in \mathcal{C}}\left[\| z^0 - z \|^2 + \| w^0 - z \|^2\right]  \notag\\
&\hspace{0.4cm}+\EE\left[\max_{z \in \mathcal{C}} \sum\limits_{k=0}^{K-1} \left[ -\tau \| w^k - z \|^2 - (1 - \tau) \| z^{k} - z\|^2 + \| w^{k+1} - z\|^2\right] \right] \notag\\
& \hspace{0.4cm}+ 2\gamma \EE\left[\max_{z \in \mathcal{C}} \sum\limits_{k=0}^{K-1} \left[ \langle Q^{\text{serv}}\left[\frac{1}{M} \sum\limits_{m=1}^M Q^{\text{dev}}_m(F_m(z^{k+1/2}) - F_m(w^k))\right] + F(w^k) - F(z^{k+1/2}),  z - z^{k+1/2} \rangle \right] \right].
\end{align}
To finish the proof we need to estimate terms in two last lines. We begin with $\E\left[ \max\limits_{z \in \mathcal{C}}\sum\limits_{k=0}^{K-1} \langle F(z^{k+1/2}) - Q^{\text{serv}}\left[\frac{1}{M} \sum\limits_{m=1}^M Q^{\text{dev}}_m(F_m(z^{k+1/2}) - F_m(w^k))\right] - F(w^k),  z^{k+1/2} - z \rangle\right]$. Let define sequence $v$: $v^0 = z^{0}$, $v^{k+1} = v^k-\gamma \delta_k$ with $\delta^k = F(z^{k+1/2}) - Q^{\text{serv}}\left[\frac{1}{M} \sum\limits_{m=1}^M Q^{\text{dev}}_m(F_m(z^{k+1/2}) - F_m(w^k))\right] - F(w^k)$. Then we have
\begin{align}
    \label{t4441}
    \sum\limits_{k=0}^{K-1} \langle \delta^k, z^{k+1/2} - u \rangle = \sum\limits_{k=0}^{K-1} \langle \delta^k, z^{k+1/2} - v^k \rangle + 
    \sum\limits_{k=0}^{K-1} \langle \delta^k,  v^k - z \rangle . 
\end{align}
By the definition of $v^{k+1}$, we have 
\begin{align*}
    \langle \gamma\delta^k, v^k  - z \rangle &= \langle \gamma\delta^k, v^k - v^{k+1} \rangle  + \langle v^{k+1} - v^k, z - v^{k+1} \rangle \nonumber\\
&= \langle \gamma\delta^k, v^k - v^{k+1} \rangle + \frac{1}{2}\|v^k - z\|^2 -  \frac{1}{2}\|v^{k+1} - z\|^2 - \frac{1}{2}\| v^k - v^{k+1}\|^2
\nonumber\\
&= \frac{\gamma^2}{2} \|\delta^k\|^2  + \frac{1}{2}\|v^k - v^{k+1}\|^2 + \frac{1}{2}\|v^k - z\|^2 -  \frac{1}{2}\|v^{k+1} - z\|^2 - \frac{1}{2}\| v^k - v^{k+1}\|^2
\nonumber\\
&= \frac{\gamma^2}{2} \|\delta^k\|^2  + \frac{1}{2}\|v^k - z\|^2 -  \frac{1}{2}\|v^{k+1} - z\|^2 .
\end{align*}
With \eqref{t4441} it gives
\begin{align*}
    \sum\limits_{k=0}^{K-1} \langle \delta^k, z^{k+1/2} - z \rangle &\leq \sum\limits_{k=0}^{K-1} \langle \delta^k, z^{k+1/2} - v^k \rangle + 
    \frac{1}{\gamma}\sum\limits_{k=0}^{K-1} \left(\frac{\gamma^2}{2} \|\delta^k\|^2  + \frac{1}{2}\|v^k - z\|^2 -  \frac{1}{2}\|v^{k+1} - z\|^2 \right) \nonumber\\
    &\leq \sum\limits_{k=0}^{K-1} \langle \delta^k, z^{k+1/2} - v^k \rangle + 
    \frac{\gamma}{2}\sum\limits_{k=0}^{K-1} \|\delta^k\|^2 + \frac{1}{2\gamma}\|v^0 - z\|^2. 
\end{align*}
We take the maximum on  $z$ and get
\begin{align*}
     \max_{z \in \mathcal{C}} \sum\limits_{k=0}^{K-1} \langle \delta^k, z^{k+1/2} - z \rangle &\leq \sum\limits_{k=0}^{K-1} \langle \delta^k, z^{k+1/2} - v^k \rangle + \frac{1}{2\gamma} \max_{z \in \mathcal{C}}  \|v^0 - z\|^2 \\
     &\hspace{0.4cm}+
    \frac{\gamma}{2}\sum\limits_{k=0}^{K-1} \|F(z^{k+1/2}) - Q^{\text{serv}}\left[\frac{1}{M} \sum\limits_{m=1}^M Q^{\text{dev}}_m(F_m(z^{k+1/2}) - F_m(w^k))\right] - F(w^k)\|^2.  
\end{align*}
Taking the full expectation, we get
\begin{align}
\label{temp404}
     \EE&\left[ \max_{z \in \mathcal{C}} \sum\limits_{k=0}^{K-1} \langle \delta^k, z^{k+1/2} - z \rangle\right] \leq \EE\left[\sum\limits_{k=0}^{K-1} \langle \delta^k, z^{k+1/2} - v^k \rangle\right] +\frac{1}{2\gamma}\max_{z \in \mathcal{C}} \|v^0 - z\|^2 \nonumber\\
    &\hspace{0.4cm}+ 
    \frac{\gamma}{2}\sum\limits_{k=0}^{K-1} \EE\left[\|F(z^{k+1/2}) - Q^{\text{serv}}\left[\frac{1}{M} \sum\limits_{m=1}^M Q^{\text{dev}}_m(F_m(z^{k+1/2}) - F_m(w^k))\right] - F(w^k)\|^2 \right]  \nonumber\\
    &= \EE\left[\sum\limits_{k=0}^{K-1} \langle \EE\left[F(z^{k+1/2}) - Q^{\text{serv}}\left[\frac{1}{M} \sum\limits_{m=1}^M Q^{\text{dev}}_m(F_m(z^{k+1/2}) - F_m(w^k))\right] - F(w^k) ~|~ z^{k+1/2} - v^k \right], z^{k+1/2} - v^k \rangle\right] \nonumber\\
    &\hspace{0.4cm}+ 
    \frac{\gamma}{2}\sum\limits_{k=0}^{K-1} \EE\left[\|F(z^{k+1/2}) - Q^{\text{serv}}\left[\frac{1}{M} \sum\limits_{m=1}^M Q^{\text{dev}}_m(F_m(z^{k+1/2}) - F_m(w^k))\right] - F(w^k)\|^2 \right]\nonumber\\
    &\hspace{0.4cm}+  \frac{1}{2\gamma}\max_{z \in \mathcal{C}} \|v^0 - z\|^2 \nonumber\\
    &= \frac{\gamma}{2}\sum\limits_{k=0}^{K-1} \EE\left[\|F(z^{k+1/2}) - Q^{\text{serv}}\left[\frac{1}{M} \sum\limits_{m=1}^M Q^{\text{dev}}_m(F_m(z^{k+1/2}) - F_m(w^k))\right] - F(w^k)\|^2 \right] \nonumber\\
    &\hspace{0.4cm}+  \frac{1}{2\gamma}\max_{z \in \mathcal{C}} \|v^0 - z\|^2.
\end{align}
Now let us estimate $\EE\left[\max\limits_{z \in \mathcal{C}} \sum\limits_{k=0}^{K-1} \left[ -\tau \| w^k - z \|^2 - (1 - \tau) \| z^{k} + z\|^2 + \| w^{k+1} - z\|^2\right] \right]$, for this we note that
\begin{align*}
    \EE&\left[\max_{z \in \mathcal{C}} \sum\limits_{k=0}^{K-1} \left[ -\tau \| w^k - z \|^2 - (1 - \tau) \| z^{k} - z\|^2 + \| w^{k+1} - z\|^2\right] \right] \\
    &= \EE\left[\max_{z \in \mathcal{C}} \sum\limits_{k=0}^{K-1} \left[ -2 \langle (1 - \tau) z^{k} + \tau w^k - w^{k+1}, z\rangle - (1 -\tau)\|z^{k} \|^2 - \tau \| w^k\|^2 + \|w^{k+1} \|^2\right] \right] \\
    &= \EE\left[\max_{z \in \mathcal{C}} \sum\limits_{k=0}^{K-1} \left[ -2 \langle (1 - \tau) z^{k} + \tau w^k - w^{k+1}, z\rangle\right]\right] \\
    &\hspace{0.4cm} +\EE\left[\sum\limits_{k=0}^{K-1} -(1 -\tau)\|z^{k} \|^2 - \tau \| w^k\|^2 + \|w^{k+1} \|^2\right] .
\end{align*}
One can note that by definition $w^{k+1}$: $\EE\left[ (1 -\tau)\|z^{k} \|^2 + \tau \| w^k\|^2 - \|w^{k+1} \|^2\right] = 0$, then 
\begin{align*}
    \EE&\left[\max_{z \in \mathcal{C}} \sum\limits_{k=0}^{K-1} \left[ -\tau \| w^k - z \|^2 - (1 - \tau) \| z^{k} - z\|^2 + \| w^{k+1} - z\|^2\right] \right] \\
    &\hspace{4cm}= 2\EE\left[\max_{z \in \mathcal{C}} \sum\limits_{k=0}^{K-1} \langle (1 - \tau) z^{k} + \tau w^k - w^{k+1}, -z\rangle\right] 
    \\
    &\hspace{4cm}= 2\EE\left[\max_{z \in \mathcal{C}} \sum\limits_{k=0}^{K-1} \langle (1 - \tau) z^{k} + \tau w^k - w^{k+1}, z\rangle\right] .
\end{align*}
Further, one can carry out the reasoning similarly to chain for \eqref{temp404}:
\begin{align}
\label{temp4042}
    \EE&\left[\max_{z \in \mathcal{C}} \sum\limits_{k=0}^{K-1} \left[ \tau \| w^k - z \|^2 + (1 - \tau) \| z^{k} - z\|^2 - \| w^{k+1} - z\|^2\right] \right] \notag\\
    &\leq \sum\limits_{k=0}^{K-1} \EE\left[\|(1 - \tau) z^{k+1} + \tau w^k - w^{k+1}\|^2 \right]+ \max_{z \in \mathcal{C}} \|v^0 - z\|^2 \notag\\
    &= \sum\limits_{k=0}^{K-1} \EE\left[\|\EE_{w^{k+1}} [w^{k+1}] - w^{k+1}\|^2 \right]+ \max_{z \in \mathcal{C}} \|v^0 - z\|^2 \notag\\
    &= \sum\limits_{k=0}^{K-1} \EE\left[-\|\EE_{w^{k+1}} [w^{k+1}] \|^2  + \EE_{w^{k+1}} \| w^{k+1}\|^2 \right]+ \max_{z \in \mathcal{C}} \|v^0 - z\|^2 \notag\\
    &= \sum\limits_{k=0}^{K-1} \EE\left[-\|(1 - \tau) z^{k} + \tau w^k \|^2  + (1 - \tau)\| z^{k}\|^2 + \tau\| w^{k}\|^2 \right]+ \max_{z \in \mathcal{C}} \|v^0 - z\|^2 \notag\\
    &= \sum\limits_{k=0}^{K-1} \tau(1 - \tau)\EE\left[\| z^{k} - w^k \|^2 \right]+ \max_{z \in \mathcal{C}} \|v^0 - z\|^2.
\end{align}
Substituting \eqref{temp404} and \eqref{temp4042} in \eqref{temp4041} we get
\begin{align}
\label{temp34}
2\gamma &\EE\left[\max_{z \in \mathcal{C}} \sum\limits_{k=0}^{K-1}\la F(z^{k+1/2}), z^{k+1/2} - z\ra  \right]\leq  \max_{z\in \mathcal{C}}\left[3\| z^0 - z\|^2 + \| w^0 - z \|^2\right] \notag\\
&\hspace{0.4cm}+\sum\limits_{k=0}^{K-1}\tau(1 - \tau)\EE\left[\| z^{k} - w^k \|^2 \right] \notag\\
&\hspace{0.4cm}
+ \gamma^2\EE\left[\|F(z^{k+1/2}) - Q^{\text{serv}}\left[\frac{1}{M} \sum\limits_{m=1}^M Q^{\text{dev}}_m(F_m(z^{k+1/2}) - F_m(w^k))\right] - F(w^k)\|^2 \right].
\end{align}
Next we work separately with $\EE\left[\|F(z^{k+1/2}) - Q^{\text{serv}}\left[\frac{1}{M} \sum\limits_{m=1}^M Q^{\text{dev}}_m(F_m(z^{k+1/2}) - F_m(w^k))\right] - F(w^k)\|^2 \right]$:
\begin{align*}
\EE&\left[\left\|Q^{\text{serv}} \left [\frac{1}{M} \sum_{m=1}^M Q^{\text{dev}}_m(F_m(z^{k+1/2}) - F_m(w^k))\right] + F(w^k) - F(z^{k+1/2})\right\|^2 \right] \\
&= \EE\left[\left\|Q^{\text{serv}} \left [\frac{1}{M} \sum_{m=1}^M Q^{\text{dev}}_m(F_m(z^{k+1/2}) - F_m(w^k))\right]\right\|^2 \right] + \EE\left[\|F(z^{k+1/2}) - F(w^k)\|^2 \right] \\
&\hspace{0.4cm}+\EE\left[\langle Q^{\text{serv}} \left [\frac{1}{M} \sum_{m=1}^M Q^{\text{dev}}_m(F_m(z^{k+1/2}) - F_m(w^k))\right]; F(z^{k+1/2}) - F(w^k) \rangle\right].
\end{align*}
With \eqref{temp33} we get
\begin{align}
\label{temp35}
\EE&\left[\left\|Q^{\text{serv}} \left [\frac{1}{M} \sum_{m=1}^M Q^{\text{dev}}_m(F_m(z^{k+1/2}) - F_m(w^k))\right] + F(w^k) - F(z^{k+1/2})\right\|^2 \right]
\nonumber\\
&\leq C_q^2 \EE\left[\left\|z^{k+1/2} - w^k \right\|^2 \right] + \EE\left[\|F(z^{k+1/2}) - F(w^k)\|^2 \right] \nonumber\\
&\hspace{0.4cm}+\EE\left[\langle \frac{1}{M} \sum_{m=1}^M Q^{\text{dev}}_m(F_m(z^{k+1/2}) - F_m(w^k)); F(z^{k+1/2}) - F(w^k) \rangle\right] \nonumber\\
&= C_q^2 \EE\left[\left\|z^{k+1/2} - w^k \right\|^2 \right] + 2\EE\left[\|F(z^{k+1/2}) - F(w^k)\|^2 \right] \nonumber\\
&\leq C_q^2 \EE\left[\left\|z^{k+1/2} - w^k \right\|^2 \right] + \frac{2}{M} \sum\limits_{m=1}^M L_m^2 \cdot \EE\left[\left\|  z^{k+1/2} - w^k\right\|^2 \right].
\end{align}
With Assumption \ref{as3} and notation $ \tilde L^2 = \frac{1}{M} \sum\limits_{m=1}^M L_m^2$ from \eqref{temp34} and \eqref{temp35} we have
\begin{align*}
2\gamma &\EE\left[\max_{z \in \mathcal{C}} \sum\limits_{k=0}^{K-1}\la F(z^{k+1/2}), z^{k+1/2} - z\ra  \right]\leq  \max_{z \in \mathcal{C}}\left[3\| z^0 - z\|^2 + \| w^0 - z \|^2\right] \notag\\
&\hspace{0.4cm}+\sum\limits_{k=0}^{K-1} \left[\tau(1 - \tau)\EE\left[\| z^{k} - w^k \|^2 \right] + \gamma^2 (C_q^2 + 2 \tilde L^2) \EE\left[\| z^{k+1/2} - w^k \|^2 \right] \right].
\end{align*}
With $\gamma \leq \frac{\sqrt{1 - \tau}}{2\sqrt{C_q^2 + 2 \tilde L^2}}$ we deduce to
\begin{align*}
2\gamma \cdot &\EE\left[\max_{z \in \mathcal{C}} \sum\limits_{k=0}^{K-1}\la F(z^{k+1/2}), z^{k+1/2} - z\ra  \right]\leq  \max_{z \in \mathcal{C}}\left[3\| z^0 - z \|^2 + \| w^0 - z \|^2\right] \notag\\
&\hspace{0.4cm}+(1 - \tau) \sum\limits_{k=0}^{K-1} \left[\EE\left[\| z^{k+1} - w^k \|^2 \right] + \EE\left[\| z^{k+1/2} -w^k \|^2\right] \right] \\
&\leq  \max_{z \in \mathcal{C}}\left[3 \| z^0 - z \|^2 + \| w^0 - z \|^2\right]  \notag\\
&\hspace{0.4cm}+3(1 - \tau) \sum\limits_{k=0}^{K-1} \left[\EE\left[\| z^{k} - z^{k+1/2} \|^2 \right] + \EE\left[\| z^{k+1/2} -w^k \|^2\right] \right] .
\end{align*}
Let us go back to \eqref{tt5657} with $\mu = 0$, $\gamma \leq \frac{\sqrt{1-\tau}}{2C_q}$ and get that
\begin{align*}
\EE(\|z^{k+1} - z^* \|^2 &+ \EE\|w^{k+1}  - z^*\|^2) \nonumber\\
    &\leq
    \EE\left(\|z^{k} - z^* \|^2 + \| w^k - z^* \|^2\right) \nonumber\\
    &\hspace{0.4cm} - \frac{1-\tau}{2}\left(\EE\| w^k - z^{k+1/2} \|^2 + \EE\| z^{k+1/2} - z^{k}\|^2\right).
\end{align*}

Hence substituting this we go to the end of the proof:
\begin{align*}
2\gamma \cdot &\EE\left[\max_{z \in \mathcal{C}} \sum\limits_{k=0}^{K-1}\la F(z^{k+1/2}), z^{k+1/2} - z\ra  \right]
\leq  \max_{z \in \mathcal{C}}\left[3 \| z^0 - z\|^2 + \| w^0 - z \|^2\right] 
\nonumber\\
    &\hspace{0.4cm}
+6 \sum\limits_{k=0}^{K-1} \left[ \EE\left(\|z^{k} - z^* \|^2 + \| w^k - z^* \|^2\right) - \EE(\|z^{k+1} - z^* \|^2 + \EE\|w^{k+1}  - z^*\|^2) \right] \\
&\leq  \max_{z \in \mathcal{C}}\left[3 \| z^0 - u \|^2 + \| w^0 - z \|^2\right] + 6 \left(\|z^{0} - z^* \|^2 + \| w^0 - z^* \|^2\right)\\
&\leq  \max_{z \in \mathcal{C}}\left[4\| z^0 - z \|^2\right] + 12 \|z^{0} - z^* \|^2.
\end{align*}
It remains to slightly correct the convergence criterion by monotonicity of $F$:
\begin{align*}
\EE&\left[\max_{z \in \mathcal{C}} \sum\limits_{k=0}^{K-1} \left[ \langle F(z^{k+1/2}),  z^{k+1/2} - z \rangle \right]  \right] \\
&\geq \EE\left[\max_{z \in \mathcal{C}} \sum\limits_{k=0}^{K-1} \left[ \langle F(u),  z^{k+1/2} - u  \rangle \right]  \right].
\end{align*}
where we additionally use $\bar z^{K} = \frac{1}{K}\sum\limits_{k=0}^{K-1} z^{k+1/2} $. This brings us to
\begin{align*}
\EE\left[\max_{z \in \mathcal{C}} \left[ \langle F(u),  \left(\frac{1}{K}\sum\limits_{k=0}^{K-1} z^{k+1/2}\right) - u  \rangle \right] \right]
\leq  \frac{2\max_{z \in \mathcal{C}}\left[\| z^0 - z \|^2\right] + 6\|z^{0} - z^* \|^2}{\gamma K}.
\end{align*}
\EndProof

\subsubsection{Non-monotone case}

We start from \eqref{t565013} 
\begin{align*}
\EE\|z^{k+1} - z^* \|^2 &+ \EE\|w^{k+1}  - z^*\|^2 \nonumber\\
    &\leq
    \EE\|z^{k} - z^* \|^2 +  \EE\| w^k - z^* \|^2 \nonumber\\
    &\hspace{0.4cm}- 2\gamma  \EE \left[\la   F(z^{k+1/2}),  z^{k+1/2} - z^*\ra\right]  \nonumber\\
    &\hspace{0.4cm} -  (1-\tau - \gamma^2 C_q^2)\EE\| w^k - z^{k+1/2} \|^2  - \tau \EE\| z^{k+1/2} - z^{k}\|^2.
\end{align*}
And then use non-monotone case of Assumption \ref{as1}:
\begin{align*}
\EE\|z^{k+1} - z^* \|^2 &+ \EE\|w^{k+1}  - z^*\|^2 \nonumber\\
    &\leq
    \EE\|z^{k} - z^* \|^2 +  \EE\| w^k - z^* \|^2 \nonumber\\
    &\hspace{0.4cm} -  (1-\tau - \gamma^2 C_q^2)\EE\| w^k - z^{k+1/2} \|^2  - \tau \EE\| z^{k+1/2} - z^{k}\|^2.
\end{align*}
With $\tau \geq \frac{1}{2}$ we get
\begin{align*}
\EE\|z^{k+1} - z^* \|^2 &+ \EE\|w^{k+1}  - z^*\|^2 \nonumber\\
    &\leq
    \EE\|z^{k} - z^* \|^2 +  \EE\| w^k - z^* \|^2 -  (1-\tau - \gamma^2 C_q^2)\EE\| w^k - z^{k+1/2} \|^2 \nonumber\\
    &\hspace{0.4cm}   - \frac{1}{4} \EE\| z^{k+1/2} - z^{k}\|^2 - \frac{1}{4} \EE\| z^{k+1/2} - z^{k}\|^2
    \nonumber\\
    &=
    \EE\|z^{k} - z^* \|^2 +  \EE\| w^k - z^* \|^2 -  (1-\tau - \gamma^2 C_q^2)\EE\| w^k - z^{k+1/2} \|^2 \nonumber\\
    &\hspace{0.4cm}   - \frac{1}{4} \EE\| z^{k+1/2} - z^{k}\|^2 - \frac{1}{4} \EE\| (1-\tau)(w^k -z^k) - \gamma F(w^k) \|^2.
\end{align*}
Using $-\| a\|^2 \leq -\frac{1}{2}\| a + b\|^2 + \| b\|^2$ gives
\begin{align*}
\EE\|z^{k+1} - z^* \|^2 &+ \EE\|w^{k+1}  - z^*\|^2 \nonumber\\
    &\leq
    \EE\|z^{k} - z^* \|^2 +  \EE\| w^k - z^* \|^2 -  (1-\tau - \gamma^2 C_q^2)\EE\| w^k - z^{k+1/2} \|^2 \nonumber\\
    &\hspace{0.4cm}   - \frac{1}{4} \EE\| z^{k+1/2} - z^{k}\|^2 - \frac{\gamma^2}{8} \EE\| F(w^k) \|^2 + \frac{(1-\tau)^2}{4} \EE\| w^k -z^k \|^2 \nonumber\\
    &\leq
    \EE\|z^{k} - z^* \|^2 +  \EE\| w^k - z^* \|^2 -  (1-\tau - \gamma^2 C_q^2)\EE\| w^k - z^{k+1/2} \|^2 \nonumber\\
    &\hspace{0.4cm}   - \frac{1}{4} \EE\| z^{k+1/2} - z^{k}\|^2 - \frac{\gamma^2}{8} \EE\| F(w^k) \|^2 \nonumber\\
    &\hspace{0.4cm} + \frac{(1-\tau)^2}{2} \EE\| w^k -z^{k+1/2} \|^2 + \frac{(1-\tau)^2}{2} \EE\| z^{k+1/2} -z^k \|^2 \nonumber\\
    &\leq
    \EE\|z^{k} - z^* \|^2 +  \EE\| w^k - z^* \|^2 -  (1-\tau - \gamma^2 C_q^2)\EE\| w^k - z^{k+1/2} \|^2 \nonumber\\
    &\hspace{0.4cm}   - \frac{1}{4} \EE\| z^{k+1/2} - z^{k}\|^2 - \frac{\gamma^2}{8} \EE\| F(w^k) \|^2 \nonumber\\
    &\hspace{0.4cm} + \frac{1-\tau}{4} \EE\| w^k -z^{k+1/2} \|^2 + \frac{1}{8} \EE\| z^{k+1/2} -z^k \|^2 \nonumber\\
    &\leq
    \EE\|z^{k} - z^* \|^2 +  \EE\| w^k - z^* \|^2 -  \left(\frac{1-\tau}{2} - \gamma^2 C_q^2\right)\EE\| w^k - z^{k+1/2} \|^2 \nonumber\\
    &\hspace{0.4cm}   - \frac{\gamma^2}{8} \EE\| F(w^k) \|^2.
\end{align*}
Choice of  $\gamma \leq \frac{\sqrt{1 - \tau}}{2 C_q}
$ gives 
\begin{align*}
\EE\|z^{k+1} - z^* \|^2 &+ \EE\|w^{k+1}  - z^*\|^2 \leq
    \EE\|z^{k} - z^* \|^2 +  \EE\| w^k - z^* \|^2    - \frac{\gamma^2}{8} \EE\| F(w^k) \|^2.
\end{align*}
Summing over all $k$ from $0$ to $K-1$ gives
\begin{align*}
 \frac{1}{K}\sum\limits_{k=0}^{K-1} \EE\| F(w^k) \|^2 
    &\leq
    \frac{8 \EE(\|z^{0} - z^* \|^2 +  \| w^0 - z^* \|^2)}{\gamma^2 K}.
\end{align*}
\EndProof

\newpage

\section{\algnamenormal{MASHA2}: Handling Contractive Compressors} \label{sec:MASHA2}

In this section, we provide additional information about Algorithm \ref{main_alg_qfb} -- \algname{MASHA2}. We give a full form of \algname{MASHA2} -- see Algorithm \ref{alg_qfb}.

Similarly with \algname{MASHA 1}, this Algorithm, locally each device stores three vectors: a current point $z^k$, a reference point $w^k$ and a values $F(w^k)$ at this point. At each iteration, performs compressed communications from devices to the server (line \ref{alg2:com_d_s}) and from the server to devices (line \ref{alg2:com_s_d}). There is also one bit $b_k$ forwarding from the server (line \ref{alg2:1bit}). Additionally, communication can occur when $b_k$ is equal to $1$ (with a small probability of $1-\tau$) -- in this case, each device $m$ updates point $w^{k+1} = z^k$, computes $F_m$ at this point, sends $F_m(w^{k+1})$ to the server without compression, the server calculates $F(w^{k+1})$ and sends it to devices also without compression. \algname{MASHA 2}, similarly with \algname{MASHA 1}, uses communications without compression, but very rarely (about once every $\frac{1}{1-\tau}$ iterations). Because, when $b_k = 0$, $w^{k+1} = w^k$ and all devices have locally value $F(w^{k+1}) = F(w^k)$ obtained sometime in previous communications (when $b=1$).
\begin{algorithm}[th]
	\caption{\algname{MASHA2}}
	\label{alg_qfb}
	\begin{algorithmic}[1]
\State
\noindent {\bf Parameters:}  Stepsize $\gamma>0$, parameter $\tau$, number of iterations $K$.\\
\noindent {\bf Initialization:} Choose  $z^0 = w^0 \in \mathcal{Z}$, $e^0_m = 0$, $e^0 = 0$. \\
Server  sends to devices $z^0 = w^0$ and devices compute $F_m(w^0)$ and send to server and get $F(w^0)$
\For {$k=0,1, 2, \ldots, K-1$ }
\For {each device $m$ in parallel}
\State $\bar z^k = \tau z^k + (1 - \tau) w^k$ \label{alg2:barz}
\State $z^{k+1/2} = \bar z_k - \gamma F(w^k)$ \label{alg2:z1/2}
\State  Compute $F_m(z^{k+1/2})$ and send to server $C^{\text{dev}}_m(\gamma F_m(z^{k+1/2}) - \gamma F_m(w^k) + e^k_m)$ \label{alg2:com_d_s}
\State $e^{k+1}_m = e^k_m + \gamma F_m(z^{k+1/2}) - \gamma F_m(w^k) - C^{\text{dev}}_m(\gamma F_m(z^{k+1/2}) - \gamma F_m(w^k) + e^k_m)$
\EndFor
\For {server}
\State Compute $g^k = C^{\text{serv}} \left [\frac{1}{M} \sum\limits_{m=1}^M C^{\text{dev}}_m(\gamma  F_m(z^{k+1/2}) - \gamma  F_m(w^k)+ e^k_m ) + e^k\right] $  \& send to devices \label{alg2:com_s_d}
\State $e^{k+1} = e^k + \frac{1}{M} \sum\limits_{m=1}^M C^{\text{dev}}_m(\gamma  F_m(z^{k+1/2}) - \gamma  F_m(w^k)+ e^k_m ) - g^k$
\State Sends to devices one bit $b_k$: 1 with probability $1 - \tau$, 0 with with probability $\tau$ \label{alg2:1bit}
\EndFor
\For {each device $m$ in parallel}
\State $z^{k+1} = z^{k+1/2} - C^{\text{serv}} \left [\frac{1}{M} \sum\limits_{m=1}^M C^{\text{dev}}_m(\gamma  F_m(z^{k+1/2}) - \gamma  F_m(w^k)+ e^k_m ) + e^k\right] $
\If {$b_k = 1$}
\State $w^{k+1} = z^{k}$ \label{alg2:wk_zk}
\State Compute $F_m(w^{k+1})$ and it send to server; and get $F(w^{k+1})$
\Else 
\State $w^{k+1} = w^k$ \label{alg2:wk_wk}
\EndIf
\EndFor
\EndFor
	\end{algorithmic}
\end{algorithm}

Let us introduce the useful notation:
$$
\hat z^{k} = z^k - e^k - \frac{1}{M} \sum\limits_{m=1}^M e^{k}_m, \quad \hat z^{k+1/2} = z^{k+1/2} - e^k - \frac{1}{M} \sum\limits_{m=1}^M e^{k}_m, \quad \hat w^{k} = w^k - e^k - \frac{1}{M} \sum\limits_{m=1}^M e^{k}_m.
$$
It is easy to verify that such sequences have a very useful property:
\begin{align}
    \label{t88}
    \hat z^{k+1} &= z^{k+1} - e^{k+1} - \frac{1}{M} \sum\limits_{m=1}^M e^{k+1}_m \nonumber\\
    &= z^{k+1/2} - C^{\text{serv}} \left [\frac{1}{M} \sum\limits_{m=1}^M C^{\text{dev}}_m(\gamma  F_m(z^{k+1/2}) - \gamma  F_m(w^k)+ e^k_m ) + e^k\right] \nonumber\\
    &\hspace{0.4cm}-e^k - \frac{1}{M} \sum\limits_{m=1}^M C^{\text{dev}}_m(\gamma  F_m(z^{k+1/2}) - \gamma  F_m(w^k)+ e^k_m ) 
    \nonumber\\
    &\hspace{0.4cm}+ C^{\text{serv}} \left [\frac{1}{M} \sum\limits_{m=1}^M C^{\text{dev}}_m(\gamma  F_m(z^{k+1/2}) - \gamma  F_m(w^k)+ e^k_m ) + e^k\right]
    \nonumber\\
    &\hspace{0.4cm}- \frac{1}{M} \sum\limits_{m=1}^M \left[e^k_m + \gamma \cdot F_m(z^{k+1/2}) - \gamma \cdot F_m(w^k) - C^{\text{dev}}_m(\gamma \cdot F_m(z^{k+1/2}) - \gamma \cdot F_m(w^k) + e^k_m)\right] \nonumber\\
    &= z^{k+1/2} - e^k - \frac{1}{M} \sum_{m=1}^M e^k_m - \gamma \cdot (F(z^{k+1/2}) - F(w^k))  \nonumber \\
    &= \hat z^{k+1/2} - \gamma \cdot (F(z^{k+1/2}) - F(w^k)).
\end{align}
The following theorem gives the convergence of \algname{MASHA2}. 

\begin{theorem} \label{th_comp}
Let distributed variational inequality \eqref{VI} + \eqref{MK} is solved by Algorithm \ref{alg_qfb} with $\tau \geq \frac{3}{4}$ and biased compressor operators \eqref{compr}: on server with $\delta^{\text{serv}}$ parameter, on devices with $\delta^{\text{dev}}$. Let Assumption \ref{as3} and one case of Assumption \ref{as1} are satisfied. Then the following estimates holds 

$\bullet$ in strongly-monotone case with $\gamma \leq \min\left[ \frac{1-\tau}{8\mu}; \frac{\sqrt{1-\tau}}{2L + 165 \delta^{\text{serv}}\delta^{\text{dev}} \tilde L}\right]$:
\begin{align*}
 \EE\left(\|\hat z^{K} - z^* \|^2 + \|w^{K} - z^* \|^2\right) &\leq \left(1-  \frac{\mu \gamma}{2}\right)^K \cdot 2\|z^{0} - z^* \|^2 ;
\end{align*}

$\bullet$ in monotone case with $\gamma \leq \frac{\sqrt{1-\tau}}{2L + 165 \delta^{\text{serv}}\delta^{\text{dev}} \tilde L}$:
\begin{align*}
    \EE\left[\max_{z \in \mathcal{C}} \la F(z) , \left(\frac{1}{K}\sum\limits_{k=0}^{K-1} z^{k+1/2}\right) - z\ra \right] 
    &\leq \frac{2\max_{z \in \mathcal{C}}\| z^{0} - z\|^2 + 4\| z^{0} - z^*\|^2}{\gamma K};
\end{align*}

$\bullet$ in non-monotone case with $\gamma \leq \frac{\sqrt{1-\tau}}{2L + 165\delta^{\text{serv}}\delta^{\text{dev}}\tilde L}$:
\begin{align*}
    \EE\left(\frac{1}{K}\sum\limits_{k=0}^{K-1}\| F(w^k)\|^2\right)
    &\leq \frac{32\EE\|z^{0} - z^*\|^2 }{\gamma^2 K}.
\end{align*}
\end{theorem}

Let us start with the only devices compression. For simplicity, we put $\tilde L = L$. We use  the same reasoning as in Section \ref{sec:masha1}. At each iteration, the device sends to the server $\mathcal{O}\left( \frac{1}{\beta} + 1-\tau\right)$ bits. Then the optimal choice $\tau$ is $1 - \frac{1}{\beta}$.

\begin{corollary} \label{cor:masha2} 
Let distributed variational inequality \eqref{VI} + \eqref{MK} is solved by Algorithm \ref{alg_qfb} without compression on server ($\delta^{\text{serv}} = 1$) and with  biased compressor operators \eqref{compr} on devices with $\delta^{\text{dev}} = \delta$. Let Assumption \ref{as3} and one case of Assumption \ref{as1} are satisfied. Then the following estimates holds 

$\bullet$ in strongly-monotone case with $\gamma \leq \min\left[ \frac{1}{8\mu \beta}; \frac{1}{167 \delta \sqrt{\beta} L}\right]$:
\begin{align*}
 \EE\left(\|\hat z^{K} - z^* \|^2 + \|w^{K} - z^* \|^2\right) &\leq \left(1-  \frac{\mu \gamma}{2}\right)^K \cdot 2\|z^{0} - z^* \|^2 ;
\end{align*}

$\bullet$ in monotone case with $\gamma \leq \frac{1}{167 \delta \sqrt{\beta}  L}$:
\begin{align*}
    \EE\left[\max_{z \in \mathcal{C}} \la F(z) , \left(\frac{1}{K}\sum\limits_{k=0}^{K-1} z^{k+1/2}\right) - z\ra \right] 
    &\leq \frac{2\max_{z \in \mathcal{C}}\| z^{0} - z\|^2 + 4\| z^{0} - z^*\|^2}{\gamma K};
\end{align*}

$\bullet$ in non-monotone case with $\gamma \leq \frac{1}{167\delta  \sqrt{\beta} L}$:
\begin{align*}
    \EE\left(\frac{1}{K}\sum\limits_{k=0}^{K-1}\| F(w^k)\|^2\right)
    &\leq \frac{32\EE\|z^{0} - z^*\|^2 }{\gamma^2 K}.
\end{align*}
\end{corollary}
In the line 2 of  Table \ref{tab:comparison1} we put complexities to achieve $\varepsilon$-solution.

Next, we add server compression. Now the transfer from the server is important. For simplicity, we put $Q^{\text{serv}} = Q^{\text{dev}}_m = Q$ with $q^{\text{dev}}_m = q$ and $\beta^{\text{dev}}_m = \beta$, also $L_m = \tilde L = L$. At each iteration, the device is sent to the server and the server to devices $\mathcal{O}\left( \frac{1}{\beta} + 1-\tau\right)$ bits. Then the optimal choice $\tau$ is still $1 - \frac{1}{\beta}$.

\begin{corollary} \label{cor:bi_masha2}
Let distributed variational inequality \eqref{VI} + \eqref{MK} is solved by Algorithm \ref{alg_qfb} with $\tau \geq \frac{3}{4}$ and biased compressor operators \eqref{compr}: on server with $\delta^{\text{serv}} = \delta$ parameter, on devices with $\delta^{\text{dev}}  = \delta$. Let Assumption \ref{as3} and one case of Assumption \ref{as1} are satisfied. Then the following estimates holds 

$\bullet$ in strongly-monotone case with $\gamma \leq \min\left[ \frac{1}{8\mu \beta}; \frac{1}{167 \delta^2 \sqrt{\beta} L}\right]$:
\begin{align*}
 \EE\left(\|\hat z^{K} - z^* \|^2 + \|w^{K} - z^* \|^2\right) &\leq \left(1-  \frac{\mu \gamma}{2}\right)^K \cdot 2\|z^{0} - z^* \|^2 ;
\end{align*}

$\bullet$ in monotone case with $\gamma \leq \frac{1}{167 \delta^2 \sqrt{\beta}  L}$:
\begin{align*}
    \EE\left[\max_{z \in \mathcal{C}} \la F(z) , \left(\frac{1}{K}\sum\limits_{k=0}^{K-1} z^{k+1/2}\right) - z\ra \right] 
    &\leq \frac{2\max_{z \in \mathcal{C}}\| z^{0} - z\|^2 + 4\| z^{0} - z^*\|^2}{\gamma K};
\end{align*}

$\bullet$ in non-monotone case with $\gamma \leq \frac{1}{167\delta^2  \sqrt{\beta} L}$:
\begin{align*}
    \EE\left(\frac{1}{K}\sum\limits_{k=0}^{K-1}\| F(w^k)\|^2\right)
    &\leq \frac{32\EE\|z^{0} - z^*\|^2 }{\gamma^2 K}.
\end{align*}
\end{corollary}
In the line 4 of  Table \ref{tab:comparison1} we put complexities to achieve $\varepsilon$-solution.

\subsection{Proof of the convergence of \algnamenormal{MASHA2}} \label{prof_th2}

\textbf{Proof of Theorem \ref{th_comp}:} We start from the following equalities for any $z$:
\begin{align*}
    \|\hat z^{k+1} - z \|^2
    &=
    \|z^{k+1/2} - z \|^2 + 2 \la \hat z^{k+1} -  z^{k+1/2},  z^{k+1/2} - z\ra + \| \hat z^{k+1} -  z^{k+1/2}\|^2,
\end{align*}
\begin{align*}
    \|z^{k+1/2} - z \|^2
    &=
    \| \hat z^{k} - z \|^2 + 2 \la z^{k+1/2} -  \hat z^{k}, z^{k+1/2} - z\ra - \|  z^{k+1/2} - \hat z^{k}\|^2.
\end{align*}
Summing up, we obtain 
\begin{align}
\label{t909}
    \|\hat z^{k+1} - z \|^2
    &=
    \|\hat z^{k} - z \|^2 + 2 \la \hat z^{k+1} -  \hat z^{k},  z^{k+1/2} - z\ra  \nonumber\\
    &\hspace{0.4cm}+ \| \hat z^{k+1} -  z^{k+1/2}\|^2 - \| z^{k+1/2} - \hat z^{k}\|^2.
\end{align}
Using that \eqref{eq:sqs} and \eqref{t88}, we get 
\begin{align}
\label{t77}
\| \hat z^{k+1} - z^{k+1/2}\|^2 
&\leq 2 \| \hat z^{k+1} - \hat z^{k+1/2}\|^2 + 2 \| \hat z^{k+1/2} - z^{k+1/2}\|^2 \nonumber\\
&= 2\gamma^2 \cdot \| F(z^{k+1/2}) - F(w^k)\|^2 + 2 \left\| e^k - \frac{1}{M} \sum_{m=1}^M e^k_m\right\|^2 \nonumber\\
&\leq 2  \gamma^2 L^2 \cdot \| z^{k+1/2} - w^{k}\|^2 + 4 \| e^k\|^2+ \frac{4}{M} \sum\limits_{m=1}^M \left\| e^{k}_m \right\|^2 \nonumber\\
&\leq 2  \gamma^2 L^2 \cdot \| z^{k+1/2} - w^{k}\|^2 + 4 \| e^k\|^2 +  \frac{4}{M} \sum\limits_{m=1}^M \left\| e^{k}_m \right\|^2.
\end{align}
Additionally, here we use that $F$ is $L$-Lipschitz (Assumption \ref{as3}). Next, \eqref{t909} with \eqref{t77} gives
\begin{align}
\label{t66}
    \|\hat z^{k+1} - z \|^2
    &\leq
    \|\hat z^{k} - z \|^2 + 2 \la \hat z^{k+1} - \hat z^{k}, z^{k+1/2} - z\ra \nonumber\\
    &\hspace{0.4cm}+ 2  \gamma^2 L^2 \cdot \| z^{k+1/2} - w^{k}\|^2 + 4 \| e^k\|^2 +  \frac{4}{M} \sum\limits_{m=1}^M \left\| e^{k}_m \right\|^2 \nonumber\\
    &\hspace{0.4cm}- \| z^{k+1/2} - \hat z^{k}\|^2.
\end{align}
Now we consider the inner product $\la \hat z^{k+1} - \hat z^{k}, z^{k+1/2} - z\ra$. Using that
\begin{align}
\label{t686}
\hat z^{k+1}-\hat z^k &= \hat z^{k+1} - \hat z^{k+1/2} + \hat z^{k+1/2} - \hat z^k = - \gamma \cdot (F(z^{k+1/2}) - F(w^k)) + z^{k+1/2} - z^k \notag\\
&= - \gamma \cdot F(z^{k+1/2}) + \bar z^{k} -  z^k,
\end{align}
and using the definition of $\bar z^k$ (line \ref{alg2:barz}), we get
\begin{align*}
    2 \la \hat z^{k+1} - \hat z^{k}, z^{k+1/2} - z\ra &=
    2 \la - \gamma \cdot F(z^{k+1/2}) +  \bar z^{k} -   z^k ,z^{k+1/2} - z\ra \\
    &= -2 \gamma \la F(z^{k+1/2}) ,z^{k+1/2} - z\ra  + 2 \la  \bar z^{k} -  z^k ,z^{k+1/2} - z\ra
    \\
    &= -2 \gamma \la F(z^{k+1/2}) ,z^{k+1/2} - z\ra  + 2 (1-\tau)\la  w^k -  z^k ,z^{k+1/2} - z\ra.
\end{align*}
Substituting in \eqref{t66}, we obtain
\begin{align*}
    \|\hat z^{k+1} - z\|^2
    &\leq
    \|\hat z^{k} - z\|^2 -2 \gamma \la F(z^{k+1/2}) ,z^{k+1/2} - z\ra  + 2 (1-\tau)\la  w^k -  z^k ,z^{k+1/2} - z\ra \nonumber\\
    &\hspace{0.4cm}+ 2  \gamma^2 L^2 \cdot \| z^{k+1/2} - w^{k}\|^2 + 4 \| e^k\|^2 +  \frac{4}{M} \sum\limits_{m=1}^M \left\| e^{k}_m \right\|^2 \nonumber\\
    &\hspace{0.4cm}- \| z^{k+1/2} - \hat z^{k}\|^2.
\end{align*}
The equality $2\la a,b \ra = \|a+b \|^2 - \| a\|^2 - \| b\|^2$ gives 
\begin{align*}
    \|\hat z^{k+1} - z\|^2
    &\leq
    \|\hat z^{k} - z\|^2 -2 \gamma \la F(z^{k+1/2}) ,z^{k+1/2} - z\ra \nonumber\\
    &\hspace{0.4cm}+ 2 (1-\tau)\la w^k - z^{k+1/2} ,z^{k+1/2} - z\ra \nonumber\\
    &\hspace{0.4cm}+ 2 (1-\tau)\la z^{k+1/2} - z^k  ,z^{k+1/2} - z\ra \nonumber\\
    &\hspace{0.4cm}+2  \gamma^2 L^2 \cdot \| z^{k+1/2} - w^{k}\|^2 + 4 \| e^k\|^2 +  \frac{4}{M} \sum\limits_{m=1}^M \left\| e^{k}_m \right\|^2 - \| z^{k+1/2} - \hat z^{k}\|^2 \nonumber\\
    &= \|\hat z^{k} - z \|^2 -2 \gamma \la F(z^{k+1/2}) ,z^{k+1/2} - z\ra \nonumber\\
    &\hspace{0.4cm}+  (1-\tau)\| w^k - z \|^2 -  (1-\tau)\| w^k - z^{k+1/2} \|^2 -   (1-\tau)\| z^{k+1/2} - z\|^2 \nonumber\\
    &\hspace{0.4cm}+ (1-\tau)\| z^{k+1/2}  -  z^k\|^2 +(1-\tau) \| z^{k+1/2} - z \|^2 - (1-\tau) \| z^{k} -z\|^2 \nonumber\\
    &\hspace{0.4cm}+ 2  \gamma^2 L^2 \cdot \| z^{k+1/2} - w^{k}\|^2 + 4 \| e^k\|^2 +  \frac{4}{M} \sum\limits_{m=1}^M \left\| e^{k}_m \right\|^2 - \| z^{k+1/2} - \hat z^{k}\|^2 \nonumber\\
    &= \|\hat z^{k} - z \|^2 - (1 - \tau)\|z^{k} - z^* \|^2 + (1 - \tau)\| w^k  - z\|^2  \nonumber\\
    &\hspace{0.4cm}-2 \gamma \la F(z^{k+1/2}) ,z^{k+1/2} - z\ra -  (1-\tau)\| w^k - z^{k+1/2} \|^2 \nonumber\\
    &\hspace{0.4cm} +  2\gamma^2 L^2\| w^k - z^{k+1/2} \|^2 +  4 \| e^k\|^2 +  \frac{4}{M} \sum\limits_{m=1}^M \left\| e^{k}_m \right\|^2 \nonumber\\
    &\hspace{0.4cm}
    - \| z^{k+1/2} - \hat z^{k}\|^2 +  (1-\tau)\| z^{k+1/2}  -  z^k\|^2\nonumber\\
    &\leq \|\hat z^{k} - z\|^2 - (1 - \tau)\|z^{k} - z \|^2 + (1 - \tau)\| w^k  - z\|^2  \nonumber\\
    &\hspace{0.4cm}-2 \gamma \la F(z^{k+1/2}) ,z^{k+1/2} - z^*\ra -  (1-\tau)\| w^k - z^{k+1/2} \|^2 \nonumber\\
    &\hspace{0.4cm} +  2\gamma^2 L^2\| w^k - z^{k+1/2} \|^2 +  4 \| e^k\|^2 +  \frac{4}{M} \sum\limits_{m=1}^M \left\| e^{k}_m \right\|^2 \nonumber\\
    &\hspace{0.4cm}
    - \frac{1}{2}\| z^{k+1/2} - z^{k}\|^2 + \| z^{k} - \hat z^{k}\|^2 +  (1-\tau)\| z^{k+1/2}  -  z^k\|^2 .
\end{align*}
With definition of $\hat z^k$ we get
\begin{align}
\label{t22}
    \|\hat z^{k+1} - z\|^2
    &\leq \|\hat z^{k} - z\|^2 - (1 - \tau)\|z^{k} - z \|^2 + (1 - \tau)\| w^k  - z\|^2  \nonumber\\
    &\hspace{0.4cm}-2 \gamma \la F(z^{k+1/2}) ,z^{k+1/2} - z\ra -  (1-\tau - 2\gamma^2 L^2)\| w^k - z^{k+1/2} \|^2 \nonumber\\
    &\hspace{0.4cm} +  6 \| e^k\|^2 +  \frac{6}{M} \sum\limits_{m=1}^M \left\| e^{k}_m \right\|^2 
    - \left(\tau - \frac{1}{2}\right)\| z^{k+1/2} - z^{k}\|^2.
\end{align}
Next we will consider three cases of monotonicity separately.

\subsubsection{Strongly-monotone}

We continue with \eqref{t22} by putting $z = z^*$ and using optimality condition:
$\la F(z^*), z^{k+1/2} - z^*\ra \leq 0$.
\begin{align*}
    \|\hat z^{k+1} - z^*\|^2
    &\leq \|\hat z^{k} - z^*\|^2 - (1 - \tau)\|z^{k} - z^* \|^2 + (1 - \tau)\| w^k  - z^*\|^2  \nonumber\\
    &\hspace{0.4cm}-2 \gamma \la F(z^{k+1/2}) - F(z^*) ,z^{k+1/2} - z^*\ra -  (1-\tau - 2\gamma^2 L^2)\| w^k - z^{k+1/2} \|^2 \nonumber\\
    &\hspace{0.4cm} +  6 \| e^k\|^2 +  \frac{6}{M} \sum\limits_{m=1}^M \left\| e^{k}_m \right\|^2 
    - \left(\tau - \frac{1}{2}\right)\| z^{k+1/2} - z^{k}\|^2.
\end{align*}
Taking a full mathematical expectation, we obtain
\begin{align}
    \label{t610}
    \EE\|\hat z^{k+1} - z^*\|^2
    &\leq \EE\|\hat z^{k} - z^*\|^2 - (1 - \tau)\EE\|z^{k} - z^* \|^2 + (1 - \tau)\EE\| w^k  - z^*\|^2  \nonumber\\
    &\hspace{0.4cm}-2 \gamma \EE\left[\la F(z^{k+1/2}) - F(z^*) ,z^{k+1/2} - z^*\ra\right] -  (1-\tau - 2\gamma^2 L^2)\EE\| w^k - z^{k+1/2} \|^2 \nonumber\\
    &\hspace{0.4cm} +  6 \EE\| e^k\|^2 +  \frac{6}{M} \sum\limits_{m=1}^M \EE\left\| e^{k}_m \right\|^2 
    - \left(\tau - \frac{1}{2}\right)\EE\| z^{k+1/2} - z^{k}\|^2.
\end{align}
Next, we take into account strong-monotonicity (Assumption \ref{as1} (SM)):
\begin{align*}
    \EE\|\hat z^{k+1} - z^*\|^2
    &\leq \EE\|\hat z^{k} - z^*\|^2 - (1 - \tau)\EE\|z^{k} - z^* \|^2 + (1 - \tau)\EE\| w^k  - z^*\|^2  \nonumber\\
    &\hspace{0.4cm}-2 \gamma \mu \EE\|z^{k+1/2} - z^*\|^2 -  (1-\tau - 2\gamma^2 L^2)\EE\| w^k - z^{k+1/2} \|^2 \nonumber\\
    &\hspace{0.4cm} +  6 \EE\| e^k\|^2 +  \frac{6}{M} \sum\limits_{m=1}^M \EE\left\| e^{k}_m \right\|^2 
    - \left(\tau - \frac{1}{2}\right)\EE\| z^{k+1/2} - z^{k}\|^2.
\end{align*}
Taking a full mathematical expectation, we obtain
\begin{align*}
    \EE\|\hat z^{k+1} - z^*\|^2
    &\leq \EE\|\hat z^{k} - z^*\|^2 - (1 - \tau)\EE\|z^{k} - z^* \|^2 + (1 - \tau)\EE\| w^k  - z^*\|^2  \nonumber\\
    &\hspace{0.4cm}-2 \gamma \mu \EE\|z^{k+1/2} - z^*\|^2 -  (1-\tau - 2\gamma^2 L^2)\EE\| w^k - z^{k+1/2} \|^2 \nonumber\\
    &\hspace{0.4cm} +  6 \EE\| e^k\|^2 +  \frac{6}{M} \sum\limits_{m=1}^M \EE\left\| e^{k}_m \right\|^2 
    - \left(\tau - \frac{1}{2}\right)\EE\| z^{k+1/2} - z^{k}\|^2.
\end{align*}
Then we use choice of $w^{k + 1}$ (lines \ref{alg2:1bit}, \ref{alg2:wk_zk}, \ref{alg2:wk_wk}) and get
\begin{align}
\label{twk}
\EE\|w^{k+1}  - z^*\|^2 = \EE\left[ \EE_{w^{k+1}}\|w^{k+1}  - z^*\|^2\right] = \tau \EE\left\|w^{k} - z^*\right\|^2 + (1 - \tau) \EE\|z^{k} - z^*\|^2,
\end{align}
Summing up the two previous expressions gives
\begin{align*}
    \EE\|\hat z^{k+1} - z^*\|^2 &+ \EE\|w^{k+1}  - z^*\|^2 \\
    &\leq \EE\|\hat z^{k} - z^*\|^2  + \EE\| w^k  - z^*\|^2 -2 \gamma \mu \EE\|z^{k+1/2} - z^*\|^2  \nonumber\\
    &\hspace{0.4cm} -  (1-\tau - 2\gamma^2 L^2)\EE\| w^k - z^{k+1/2} \|^2 - \left(\tau - \frac{1}{2}\right)\EE\| z^{k+1/2} - z^{k}\|^2 \nonumber\\
    &\hspace{0.4cm} +  6 \EE\| e^k\|^2 +  \frac{6}{M} \sum\limits_{m=1}^M \EE\left\| e^{k}_m \right\|^2.
\end{align*}
Then we can weight previous expression by $p^k$ and get
\begin{align}
\label{t1000}
 \sum\limits_{k=0}^{K-1} &p^k \EE\|\hat z^{k+1} - z^* \|^2 + \sum\limits_{k=0}^{K-1} p^k \EE\|w^{k+1} - z^* \|^2 \nonumber\\
 &\leq \sum\limits_{k=0}^{K-1} p^k \EE\|\hat z^{k} - z^* \|^2 + \sum\limits_{k=0}^{K-1} p^k \EE\|w^{k} - z^* \|^2 - 2 \gamma \mu \sum\limits_{k=0}^{K-1} p^k \EE\| z^{k+1/2} - z^*\|^2 \nonumber\\
    & \hspace{0.4cm}-  (1-\tau - 2\gamma^2 L^2 ) \cdot \sum\limits_{k=0}^{K-1} p^k \EE\| w^k - z^{k+1/2} \|^2 - \left(\tau - \frac{1}{2}\right)\cdot \sum\limits_{k=0}^{K-1} p^k\EE\| z^{k+1/2} - z^{k}\|^2\nonumber\\
    &\hspace{0.4cm} +  6 \cdot \sum\limits_{k=0}^{K-1} p^k\EE\| e^k\|^2 +  6 \cdot \sum\limits_{k=0}^{K-1} p^k \frac{1}{M} \sum\limits_{m=1}^M \EE\left\| e^{k}_m \right\|^2.
\end{align}
Next we will estimate "error" term:
\begin{align*}
 \EE\| e^{k+1}\|^2 &= 
 \EE\Bigg\|e^k + \frac{1}{M} \sum\limits_{m=1}^M C^{\text{dev}}_m(\gamma  F_m(z^{k+1/2}) - \gamma  F_m(w^k)+ e^k_m ) \\
 &\hspace{0.4cm}- C^{\text{serv}} \left [\frac{1}{M} \sum\limits_{m=1}^M C^{\text{dev}}_m(\gamma  F_m(z^{k+1/2}) - \gamma  F_m(w^k)+ e^k_m ) + e^k\right] \Bigg\|^2
 \\
&\leq  
\left(1-\frac{1}{\delta^{\text{serv}}}\right) \EE\left\|e^k + \frac{1}{M} \sum\limits_{m=1}^M C^{\text{dev}}_m(\gamma  F_m(z^{k+1/2}) - \gamma  F_m(w^k)+ e^k_m )\right\|^2
\\
&\leq  (1+c)\left(1-\frac{1}{\delta^{\text{serv}}}\right) \EE\left\|e^k\right\|^2 \\
&\hspace{0.4cm}+ \left(1 + \frac{1}{c}\right)\left(1-\frac{1}{\delta^{\text{serv}}}\right) \frac{1}{M} \sum\limits_{m=1}^M\EE\left\| C^{\text{dev}}_m(\gamma  F_m(z^{k+1/2}) - \gamma  F_m(w^k)+ e^k_m )\right\|^2.
\end{align*}
Here we use definition of biased compression \eqref{compr}, \eqref{eq:sqs} and inequality $\|a + b\|^2 \leq (1+ c) \| a\|^2 + (1 + 1/c) \| b\|^2$ (for $c > 0$). Is is easy to prove that for baised compressor $C^{\text{dev}}_m$ from \eqref{compr} it holds that $\|C^{\text{dev}}_m(x) \|^2 \leq 4 \|x\|^2$ (see \cite{beznosikov2020biased}). Then
\begin{align*}
\EE\| e^{k+1}\|^2 
&\leq  (1+c)\left(1-\frac{1}{\delta^{\text{serv}}}\right) \EE\left\|e^k\right\|^2 \\
&\hspace{0.4cm}+ \left(1 + \frac{1}{c}\right)\left(1-\frac{1}{\delta^{\text{serv}}}\right) \frac{4}{M} \sum\limits_{m=1}^M\EE\left\| \gamma  F_m(z^{k+1/2}) - \gamma  F_m(w^k)+ e^k_m \right\|^2 \\
&\leq  (1+c)\left(1-\frac{1}{\delta^{\text{serv}}}\right) \EE\left\|e^k\right\|^2 \\
&\hspace{0.4cm}+ \gamma^2\left(1 + \frac{1}{c}\right)\left(1-\frac{1}{\delta^{\text{serv}}}\right) \frac{8}{M} \sum\limits_{m=1}^M\EE\left\|  F_m(z^{k+1/2}) - F_m(w^k)\right\|^2
\\
&\hspace{0.4cm}+ \left(1 + \frac{1}{c}\right)\left(1-\frac{1}{\delta^{\text{serv}}}\right) \frac{8}{M} \sum\limits_{m=1}^M \EE\left\|e^k_m \right\|^2
\\
&\leq  (1+c)\left(1-\frac{1}{\delta^{\text{serv}}}\right) \EE\left\|e^k\right\|^2 + 8\gamma^2 \tilde L^2\left(1 + \frac{1}{c}\right)\left(1-\frac{1}{\delta^{\text{serv}}}\right) \EE\left\|  z^{k+1/2} - w^k\right\|^2
\\
&\hspace{0.4cm}+ \left(1 + \frac{1}{c}\right)\left(1-\frac{1}{\delta^{\text{serv}}}\right) \frac{8}{M} \sum\limits_{m=1}^M\EE\left\|e^k_m \right\|^2.
\end{align*}
In the last we use Assumption \ref{as3} and definition of $\tilde L$ from this Assumption. With $c = \frac{1}{2(\delta - 1)}$ we get 
\begin{align*}
 \EE\| e^{k+1}\|^2
&\leq \left(1-\frac{1}{2\delta^{\text{serv}}}\right) \EE \left\|e^k\right\|^2 + 16\delta^{\text{serv}}\gamma^2 \tilde L^2\cdot \EE\|  z^{k+1/2} - w^k\|^2
+ 16\delta^{\text{serv}} \cdot \frac{1}{M} \sum\limits_{m=1}^M \EE\left\|e^k_m \right\|^2 \\
&\leq  16\delta^{\text{serv}}\gamma^2 \tilde L^2 \sum\limits_{j=0}^k \left(1-\frac{1}{2\delta^{\text{serv}}}\right)^{k-j}  \cdot \left\|  z^{j+1/2} - w^j  \right\|^2 \\ 
&\hspace{0.4cm}+ 16\delta^{\text{serv}}\sum\limits_{j=0}^k \left(1-\frac{1}{2\delta^{\text{serv}}}\right)^{k-j}  \cdot \frac{1}{M} \sum\limits_{m=1}^M\left\|e^j_m \right\|^2.
\end{align*}
We weigh the sequence as follows $\sum\limits_{k=0}^K p^k \EE\left\| e^{k} \right\|^2$. Here we also assume $p$ such that $p^k \leq p^{j} (1 + 1/4\delta^{\text{serv}})^{k-j}$. Then 
\begin{align}
\label{t11}
 \sum\limits_{k=0}^{K-1} p^k \EE\left\| e^{k} \right\|^2
&\leq 16\delta^{\text{serv}}\gamma^2 \tilde L^2 \sum\limits_{k=0}^{K-1} p^k \sum\limits_{j=0}^{k-1} \left(1-\frac{1}{2\delta^{\text{serv}}}\right)^{k-j-1}  \cdot \EE\left\|  z^{j+1/2} -  w^j  \right\|^2 \nonumber\\
&\hspace{0.4cm} +16\delta^{\text{serv}} \sum\limits_{k=0}^{K-1} p^k \sum\limits_{j=0}^{k-1} \left(1-\frac{1}{2\delta^{\text{serv}}}\right)^{k-j-1}  \cdot \frac{1}{M} \sum\limits_{m=1}^M\EE\left\|e^j_m \right\|^2
\nonumber\\
&\leq  \frac{16\delta^{\text{serv}}\gamma^2 \tilde L^2}{(1-1/2\delta^{\text{serv}})} \sum\limits_{k=0}^{K-1} \sum\limits_{j=0}^{k-1} p^{j} \left(1+\frac{1}{4\delta^{\text{serv}}}\right)^{k-j} \left(1-\frac{1}{2\delta^{\text{serv}}}\right)^{k-j}  \cdot \EE \left\|  z^{j+1/2} - w^j  \right\|^2 \nonumber\\
&\hspace{0.4cm} + \frac{16\delta^{\text{serv}}}{(1-1/2\delta^{\text{serv}})} \sum\limits_{k=0}^{K-1} \sum\limits_{j=0}^{k-1} p^{j} \left(1+\frac{1}{4\delta^{\text{serv}}}\right)^{k-j} \left(1-\frac{1}{2\delta^{\text{serv}}}\right)^{k-j}  \cdot \frac{1}{M} \sum\limits_{m=1}^M \EE\left\|e^j_m \right\|^2
\nonumber\\
&\leq  \frac{16\delta^{\text{serv}}\gamma^2 \tilde L^2}{(1-1/2\delta^{\text{serv}})} \sum\limits_{k=0}^{K-1} \sum\limits_{j=0}^{k-1} p^{j} \left(1-\frac{1}{4\delta^{\text{serv}}}\right)^{k-j}  \cdot \EE\left\|  z^{j+1/2} - w^j  \right\|^2 \nonumber\\
&\hspace{0.4cm} + \frac{16\delta^{\text{serv}}}{(1-1/2\delta^{\text{serv}})} \sum\limits_{k=0}^{K-1} \sum\limits_{j=0}^{k-1} p^{j} \left(1-\frac{1}{4\delta^{\text{serv}}}\right)^{k-j}   \cdot \frac{1}{M} \sum\limits_{m=1}^M \EE\left\|e^j_m \right\|^2 \nonumber\\
&\leq  \frac{16\delta^{\text{serv}}\gamma^2 \tilde L^2}{(1-1/2\delta^{\text{serv}})} \sum\limits_{k=0}^{K-1}  p^{k}  \EE\left\|  z^{k+1/2} - w^k  \right\|^2 \cdot \sum\limits_{j=0}^{\infty}  \left(1-\frac{1}{4\delta^{\text{serv}}}\right)^{j} \nonumber\\
&\hspace{0.4cm} + \frac{16\delta^{\text{serv}}}{(1-1/2\delta^{\text{serv}})} \sum\limits_{k=0}^{K-1}  p^{k}   \frac{1}{M} \sum\limits_{m=1}^M \EE\left\|e^k_m \right\|^2 \cdot \sum\limits_{j=0}^{\infty}  \left(1-\frac{1}{4\delta^{\text{serv}}}\right)^{j} \nonumber\\
&\leq  128(\delta^{\text{serv}})^2\gamma^2 \tilde L^2 \sum\limits_{k=0}^{K-1}  p^{k} \EE\left\|  z^{k+1/2} - w^k  \right\|^2  \nonumber\\
&\hspace{0.4cm}+ 128(\delta^{\text{serv}})^2 \sum\limits_{k=0}^{K-1} p^k   \frac{1}{M} \sum\limits_{m=1}^M \EE\left\|e^k_m \right\|^2.
\end{align}
Combining \eqref{t1000} with \eqref{t11}, we obtain
\begin{align*}
 \sum\limits_{k=0}^{K-1} &p^k \left(\EE\|\hat z^{k+1} - z^* \|^2 + \EE\|w^{k+1} - z^* \|^2\right) \nonumber\\
    &\leq \sum\limits_{k=0}^{K-1} p^k \EE\|\hat z^{k} - z^* \|^2 + \sum\limits_{k=0}^{K-1} p^k \EE\|w^{k} - z^* \|^2 - 2 \gamma \mu \sum\limits_{k=0}^{K-1} p^k \EE\| z^{k+1/2} - z^*\|^2 \nonumber\\
    & \hspace{0.4cm}-  (1-\tau - 2\gamma^2 L^2 ) \cdot \sum\limits_{k=0}^{K-1} p^k \EE\| w^k - z^{k+1/2} \|^2 - \left(\tau - \frac{1}{2}\right)\cdot \sum\limits_{k=0}^{K-1} p^k\EE\| z^{k+1/2} - z^{k}\|^2\nonumber\\
    &\hspace{0.4cm} +  768 (\delta^{\text{serv}})^2\gamma^2 \tilde L^2 \cdot \sum\limits_{k=0}^{K-1}  p^{k} \EE\left\|  z^{k+1/2} - w^k  \right\|^2  \nonumber\\
    &\hspace{0.4cm}+ 768(\delta^{\text{serv}})^2 \cdot  \sum\limits_{k=0}^{K-1} p^k   \frac{1}{M} \sum\limits_{m=1}^M \EE\left\|e^k_m \right\|^2 +  6 \cdot \sum\limits_{k=0}^{K-1} p^k \frac{1}{M} \sum\limits_{m=1}^M \EE\left\| e^{k}_m \right\|^2
    \nonumber\\
    &\leq \sum\limits_{k=0}^{K-1} p^k \EE\|\hat z^{k} - z^* \|^2 + \sum\limits_{k=0}^{K-1} p^k \EE\|w^{k} - z^* \|^2 - 2 \gamma \mu \sum\limits_{k=0}^{K-1} p^k \EE\| z^{k+1/2} - z^*\|^2 \nonumber\\
    & \hspace{0.4cm}-  (1-\tau - 2\gamma^2 L^2 - 768 (\delta^{\text{serv}})^2\gamma^2 \tilde L^2) \cdot \sum\limits_{k=0}^{K-1} p^k \EE\| w^k - z^{k+1/2} \|^2 \nonumber\\
    &\hspace{0.4cm}- \left(\tau - \frac{1}{2}\right)\cdot \sum\limits_{k=0}^{K-1} p^k\EE\| z^{k+1/2} - z^{k}\|^2 + 775(\delta^{\text{serv}})^2 \cdot \sum\limits_{k=0}^{K-1} p^k   \frac{1}{M} \sum\limits_{m=1}^M \EE\left\|e^k_m \right\|^2.
\end{align*}
Using $-\| a\|^2 \leq -\frac{1}{2}\| a + b\|^2 + \| b\|^2$, we get
\begin{align*}
 \sum\limits_{k=0}^{K-1} &p^k \left(\EE\|\hat z^{k+1} - z^* \|^2 + \EE\|w^{k+1} - z^* \|^2\right) \nonumber\\
    &\leq \sum\limits_{k=0}^{K-1} p^k \left(1 - \frac{\mu \gamma}{2}\right) \left(\EE\|\hat z^{k} - z^* \|^2 + \EE\|w^{k} - z^* \|^2\right) \nonumber\\
    & \hspace{0.4cm}-  (1-\tau - \gamma \mu - 2\gamma^2 L^2 - 768 (\delta^{\text{serv}})^2\gamma^2 \tilde L^2) \cdot \sum\limits_{k=0}^{K-1} p^k \EE\| w^k - z^{k+1/2} \|^2 \nonumber\\
    &\hspace{0.4cm}- \left(\tau - \frac{1}{2}\right)\cdot \sum\limits_{k=0}^{K-1} p^k\EE\| z^{k+1/2} - z^{k}\|^2 + 775(\delta^{\text{serv}})^2 \cdot \sum\limits_{k=0}^{K-1} p^k   \frac{1}{M} \sum\limits_{m=1}^M \EE\left\|e^k_m \right\|^2 \nonumber\\
    &\hspace{0.4cm}+ \gamma \mu \cdot \sum\limits_{k=0}^{K-1} p^k \EE\| z^{k+1/2} - \hat z^{k}\|^2 \nonumber\\
    &\leq \sum\limits_{k=0}^{K-1} p^k \left(1 - \frac{\mu \gamma}{2}\right) \left(\EE\|\hat z^{k} - z^* \|^2 + \EE\|w^{k} - z^* \|^2\right) \nonumber\\
    & \hspace{0.4cm}-  (1-\tau - \gamma \mu - 2\gamma^2 L^2 - 768 (\delta^{\text{serv}})^2\gamma^2 \tilde L^2) \cdot \sum\limits_{k=0}^{K-1} p^k \EE\| w^k - z^{k+1/2} \|^2 \nonumber\\
    &\hspace{0.4cm}- \left(\tau - 2\gamma \mu - \frac{1}{2}\right)\cdot \sum\limits_{k=0}^{K-1} p^k\EE\| z^{k+1/2} - z^{k}\|^2 \nonumber\\
    &\hspace{0.4cm}
    + (775(\delta^{\text{serv}})^2 + 2\mu\gamma) \cdot \sum\limits_{k=0}^{K-1} p^k   \frac{1}{M} \sum\limits_{m=1}^M \EE\left\|e^k_m \right\|^2.
\end{align*}
With $\gamma \leq \frac{1}{2\mu}$ we get
\begin{align}
\label{t673}
 \sum\limits_{k=0}^{K-1} &p^k \left(\EE\|\hat z^{k+1} - z^* \|^2 + \EE\|w^{k+1} - z^* \|^2\right) \nonumber\\
    &\leq \sum\limits_{k=0}^{K-1} p^k \left(1 - \frac{\mu \gamma}{2}\right) \left(\EE\|\hat z^{k} - z^* \|^2 + \EE\|w^{k} - z^* \|^2\right) \nonumber\\
    & \hspace{0.4cm}-  (1-\tau - \gamma \mu - 2\gamma^2 L^2 - 768 (\delta^{\text{serv}})^2\gamma^2 \tilde L^2) \cdot \sum\limits_{k=0}^{K-1} p^k \EE\| w^k - z^{k+1/2} \|^2 \nonumber\\
    &\hspace{0.4cm}- \left(\tau - 2\gamma \mu - \frac{1}{2}\right)\cdot \sum\limits_{k=0}^{K-1} p^k\EE\| z^{k+1/2} - z^{k}\|^2 \nonumber\\
    &\hspace{0.4cm}
    + 776(\delta^{\text{serv}})^2 \cdot \sum\limits_{k=0}^{K-1} p^k   \frac{1}{M} \sum\limits_{m=1}^M \EE\left\|e^k_m \right\|^2.
\end{align}

Next, we work with the other "error" term. The same way as for \eqref{t11} we get
\begin{align*}
 \frac{1}{M} \sum\limits_{m=1}^M \EE\left\| e_m^{k+1} \right\|^2 &= 
 \frac{1}{M} \sum\limits_{m=1}^M \left\|  e^k_m + \gamma \cdot F_m(z^{k+1/2}) - \gamma \cdot F_m(w^k) - C^{\text{dev}}_m(\gamma \cdot F_m(z^{k+1/2}) - \gamma \cdot F_m(w^k) + e^{k}_m) \right\|^2 \\
&\leq  
\frac{1}{M} \sum\limits_{m=1}^M \left(1-\frac{1}{\delta^{\text{dev}}}\right)\left\|  e^k_m + \gamma \cdot F_m(z^{k+1/2}) - \gamma \cdot F_m(w^k)  \right\|^2 \\
&\leq  \frac{1}{M} \sum\limits_{m=1}^M (1 + c)\left(1-\frac{1}{\delta^{\text{dev}}}\right)\left\|  e^k_m \right\|^2 + \left(1 + \frac{1}{c}\right)\left(1-\frac{1}{\delta^{\text{dev}}}\right)\gamma^2 \cdot \left\|  F_m(z^{k+1/2}) - F_m(w^k)  \right\|^2 .
\end{align*}
With $c = \frac{1}{2(\delta^{\text{dev}} - 1)}$
\begin{align*}
\frac{1}{M} \sum\limits_{m=1}^M \left\| e_m^{k+1} \right\|^2
&\leq  \frac{1}{M} \sum\limits_{m=1}^M \left(1-\frac{1}{2\delta^{\text{dev}}}\right)\left\|  e^k_m \right\|^2 + 2 \delta^{\text{dev}} \gamma^2 \cdot \left\|  F_m(z^{k+1/2}) - F_m(w^k)  \right\|^2 \\
&\leq \left(1-\frac{1}{2\delta^{\text{dev}}}\right) \cdot \frac{1}{M} \sum\limits_{m=1}^M \left\|  e^k_m \right\|^2 + 2  \delta^{\text{dev}} \gamma^2 \tilde L^2 \cdot \left\|  z^{k+1/2} - w^k  \right\|^2 \\
&\leq  2 \delta^{\text{dev}} \gamma^2 \tilde L^2 \sum\limits_{j=0}^k \left(1-\frac{1}{2\delta^{\text{dev}}}\right)^{k-j}  \cdot \left\|  z^{j+1/2} - w^j  \right\|^2.
\end{align*}
We weigh the sequence as follows $\sum\limits_{k=0}^K p^k\frac{1}{M} \sum\limits_{m=1}^M \left\| e_m^{k} \right\|^2$. Here we assume that $p$ such that $p^k \leq p^{j} (1 + 1/4\delta^{\text{dev}})^{k-j}$. Then 
\begin{align}
\label{t111}
 \sum\limits_{k=0}^{K-1} p^k \frac{1}{M} \sum\limits_{m=1}^M \left\| e_m^{k} \right\|^2
&\leq 2  \delta^{\text{dev}} \gamma^2 \tilde L^2 \sum\limits_{k=0}^{K-1} p^k \sum\limits_{j=0}^{k-1}  \left(1-\frac{1}{2\delta^{\text{dev}}}\right)^{k-j-1}  \cdot \left\|  z^{j+1/2} -  w^j  \right\|^2 \nonumber\\
&\leq  \frac{2  \delta^{\text{dev}} \gamma^2 \tilde L^2}{(1-1/2\delta^{\text{dev}})} \sum\limits_{k=0}^{K-1} \sum\limits_{j=0}^{k-1} p^{j} \left(1+\frac{1}{4\delta^{\text{dev}}}\right)^{k-j} \left(1-\frac{1}{2\delta^{\text{dev}}}\right)^{k-j}  \cdot \left\|  z^{j+1/2} - w^j  \right\|^2 \nonumber\\
&\leq  \frac{2  \delta^{\text{dev}} \gamma^2 \tilde L^2}{(1-1/2\delta^{\text{dev}})} \sum\limits_{k=0}^{K-1} \sum\limits_{j=0}^{k-1} p^{j}  \left(1-\frac{1}{4\delta^{\text{dev}}}\right)^{k-j}  \cdot \left\|  z^{j+1/2} - w^j  \right\|^2 \nonumber\\
&\leq  \frac{2  \delta^{\text{dev}} \gamma^2 \tilde L^2}{(1-1/2\delta^{\text{dev}})} \sum\limits_{k=0}^{K-1} p^{k}\left\|  z^{k+1/2} - w^k  \right\|^2   \cdot \sum\limits_{j=0}^{\infty}  \left(1-\frac{1}{4\delta^{\text{dev}}}\right)^{j}  \nonumber\\
&\leq  16(\delta^{\text{dev}})^2 \gamma^2 \tilde L^2 \sum\limits_{k=0}^{K-1} p^{k} \left\|  z^{k+1/2} - w^k  \right\|^2 .
\end{align}
\eqref{t673} together with \eqref{t111} gives
\begin{align}
\label{t507}
 \sum\limits_{k=0}^{K-1} &p^k \left(\EE\|\hat z^{k+1} - z^* \|^2 + \EE\|w^{k+1} - z^* \|^2\right) \nonumber\\
    &\leq \sum\limits_{k=0}^{K-1} p^k \left(1 - \frac{\mu \gamma}{2}\right) \left(\EE\|\hat z^{k} - z^* \|^2 + \EE\|w^{k} - z^* \|^2\right) \nonumber\\
    & \hspace{0.4cm}-  (1-\tau - \gamma \mu - 2\gamma^2 L^2 - 13200(\delta^{\text{serv}})^2 (\delta^{\text{dev}})^2 \gamma^2 \tilde L^2  ) \cdot \sum\limits_{k=0}^{K-1} p^k \EE\| w^k - z^{k+1/2} \|^2 \nonumber\\
    &\hspace{0.4cm}- \left(\tau - 2\gamma\mu - \frac{1}{2}\right)\cdot \sum\limits_{k=0}^{K-1} p^k\EE\| z^{k+1/2} - z^{k}\|^2 .
\end{align}
With $\tau \geq \frac{3}{4}$, $\gamma \leq \min\left[ \frac{1-\tau}{8\mu}; \frac{\sqrt{1-\tau}}{2L + 165 \delta^{\text{serv}}\delta^{\text{dev}} \tilde L}\right]$ we obtain
\begin{align*}
 \sum\limits_{k=0}^{K-1} &p^k \left(\EE\|\hat z^{k+1} - z^* \|^2 + \EE\|w^{k+1} - z^* \|^2\right) \leq \sum\limits_{k=0}^{K-1} p^k \left(1 - \frac{\mu\gamma}{2}\right) \left(\EE\|\hat z^{k+1} - z^* \|^2 + \EE\|w^{k+1} - z^* \|^2\right).
\end{align*}
Then we just need to take $p = 1/(1 - \mu \gamma / 2)$ (easy to check that $p^k \leq p^{j} (1 + 1/8\delta^{\text{serv}})^{k-j}$ and $p^k \leq p^{j} (1 + 1/8\delta^{\text{dev}})^{k-j}$ work with our $\gamma \leq \frac{\sqrt{1-\tau}}{2L + 150 \delta^{\text{serv}}\delta^{\text{dev}} \tilde L}$) and get
\begin{align*}
 \EE\left(\|\hat z^{K} - z^* \|^2 + \|w^{K} - z^* \|^2\right) &\leq \left(1-  \frac{\mu \gamma}{2}\right)^K\left(\|\hat z^{0} - z^* \|^2 + \|w^{0} - z^* \|^2\right).
\end{align*}
This ends the proof for strongly-monotone case.
\EndProof

\subsubsection{Monotone}

Let us comeback and start from \eqref{t22}:
\begin{align*}
    2 \gamma \la F(z^{k+1/2}) ,z^{k+1/2} - z\ra
    &\leq \|\hat z^{k} - z\|^2 - \|\hat z^{k+1} - z\|^2 - (1 - \tau)\|z^{k} - z \|^2 + (1 - \tau)\| w^k  - z\|^2  \nonumber\\
    &\hspace{0.4cm} -  (1-\tau - 2\gamma^2 L^2)\| w^k - z^{k+1/2} \|^2 \nonumber\\
    &\hspace{0.4cm} +  6 \| e^k\|^2 +  \frac{6}{M} \sum\limits_{m=1}^M \left\| e^{k}_m \right\|^2 
    - \left(\tau - \frac{1}{2}\right)\| z^{k+1/2} - z^{k}\|^2.
\end{align*}
Then we use monotonicity (Assumption \ref{as1} (M)) and get
\begin{align*}
    2 \gamma \la F(z) ,z^{k+1/2} - z\ra
    &\leq \|\hat z^{k} - z\|^2 - \|\hat z^{k+1} - z\|^2 - (1 - \tau)\|z^{k} - z \|^2 + (1 - \tau)\| w^k  - z\|^2  \nonumber\\
    &\hspace{0.4cm} -  (1-\tau - 2\gamma^2 L^2)\| w^k - z^{k+1/2} \|^2 \nonumber\\
    &\hspace{0.4cm} +  6 \| e^k\|^2 +  \frac{6}{M} \sum\limits_{m=1}^M \left\| e^{k}_m \right\|^2 
    - \left(\tau - \frac{1}{2}\right)\| z^{k+1/2} - z^{k}\|^2 \\
    &= \|\hat z^{k} - z\|^2 - \|\hat z^{k+1} - z\|^2 + \| w^k  - z\|^2 - \| w^{k+1}  - z\|^2 \nonumber\\
    &\hspace{0.4cm}
    \| w^{k+1}  - z\|^2 - (1 - \tau)\|z^{k} - z \|^2 - \tau\| w^k  - z\|^2  \nonumber\\
    &\hspace{0.4cm} -  (1-\tau - 2\gamma^2 L^2)\| w^k - z^{k+1/2} \|^2 \nonumber\\
    &\hspace{0.4cm} +  6 \| e^k\|^2 +  \frac{6}{M} \sum\limits_{m=1}^M \left\| e^{k}_m \right\|^2 
    - \left(\tau - \frac{1}{2}\right)\| z^{k+1/2} - z^{k}\|^2.
\end{align*}
Next, we sum from $0$ to $K-1$:
\begin{align*}
    2 \gamma &\sum\limits_{k=0}^{K-1} \la F(z) ,z^{k+1/2} - z\ra \\
    &\leq \|\hat z^{0} - z\|^2 + \| w^0 - z \|^2
    + \sum\limits_{k=0}^{K-1} \left(\| w^{k+1}  - z\|^2 - (1 - \tau)\|z^{k} - z \|^2 - \tau\| w^k  - z\|^2\right)  \nonumber\\
    &\hspace{0.4cm} -  (1-\tau - 2\gamma^2 L^2)\cdot \sum\limits_{k=0}^{K-1} \| w^k - z^{k+1/2} \|^2 - \left(\tau - \frac{1}{2}\right) \cdot \sum\limits_{k=0}^{K-1} \| z^{k+1/2} - z^{k}\|^2 \nonumber\\
    &\hspace{0.4cm} +  6 \cdot \sum\limits_{k=0}^{K-1} \| e^k\|^2 +  6 \cdot \sum\limits_{k=0}^K  \frac{1}{M} \sum\limits_{m=1}^M \left\| e^{k}_m \right\|^2.
\end{align*}
Then we take maximum of both sides over $z \in \mathcal{C}$, after take expectation and get
\begin{align*}
    2 \gamma &\EE\left[\max_{z \in \mathcal{C}}\sum\limits_{k=0}^{K-1} \la F(z) ,z^{k+1/2} - z\ra \right] \nonumber\\
    &\leq \EE\left[\max_{z \in \mathcal{C}}\|\hat z^{0} - z\|^2 \right] + \EE\left[\max_{z \in \mathcal{C}}\|w^{0} - z\|^2 \right] 
    \nonumber\\
    &\hspace{0.4cm} + \EE\left[\max_{z \in \mathcal{C}}\sum\limits_{k=0}^{K-1} \left(\| w^{k+1}  - z\|^2 - (1 - \tau)\|z^{k} - z \|^2 - \tau\| w^k  - z\|^2\right)\right]  \nonumber\\
    &\hspace{0.4cm} -  (1-\tau - 2\gamma^2 L^2)\cdot \sum\limits_{k=0}^{K-1} \EE\| w^k - z^{k+1/2} \|^2 - \left(\tau - \frac{1}{2}\right) \cdot \sum\limits_{k=0}^{K-1} \EE\| z^{k+1/2} - z^{k}\|^2 \nonumber\\
    &\hspace{0.4cm} +  6 \cdot \sum\limits_{k=0}^{K-1} \EE\| e^k\|^2 +  6 \cdot \sum\limits_{k=0}^{K-1}  \frac{1}{M} \sum\limits_{m=1}^M \EE\left\| e^{k}_m \right\|^2.
\end{align*}
We star with using \eqref{t11} with $p=1$ and get
\begin{align*}
    2 \gamma &\EE\left[\max_{z \in \mathcal{C}}\sum\limits_{k=0}^{K-1} \la F(z) ,z^{k+1/2} - z\ra \right] \nonumber\\
    &\leq \EE\left[\max_{z \in \mathcal{C}}\|\hat z^{0} - z\|^2 \right] + \EE\left[\max_{z \in \mathcal{C}}\|w^{0} - z\|^2 \right] 
    \nonumber\\
    &\hspace{0.4cm} + \EE\left[\max_{z \in \mathcal{C}}\sum\limits_{k=0}^{K-1} \left(\| w^{k+1}  - z\|^2 - (1 - \tau)\|z^{k} - z \|^2 - \tau\| w^k  - z\|^2\right)\right]  \nonumber\\
    &\hspace{0.4cm} -  (1-\tau - 2\gamma^2 L^2 - 768(\delta^{\text{serv}})^2\gamma^2 \tilde L^2)\cdot \sum\limits_{k=0}^{K-1} \EE\| w^k - z^{k+1/2} \|^2  \nonumber\\
    &\hspace{0.4cm}  - \left(\tau - \frac{1}{2}\right) \cdot \sum\limits_{k=0}^{K-1} \EE\| z^{k+1/2} - z^{k}\|^2 + 775(\delta^{\text{serv}})^2 \cdot \sum\limits_{k=0}^K p^k   \frac{1}{M} \sum\limits_{m=1}^M \EE\left\|e^k_m \right\|^2.
\end{align*}
And then \eqref{t111} (also with $p=1$):
\begin{align*}
    2 \gamma &\EE\left[\max_{z \in \mathcal{C}}\sum\limits_{k=0}^{K-1} \la F(z) ,z^{k+1/2} - z\ra \right] \nonumber\\
    &\leq \EE\left[\max_{z \in \mathcal{C}}\|\hat z^{0} - z\|^2 \right] + \EE\left[\max_{z \in \mathcal{C}}\|w^{0} - z\|^2 \right] 
    \nonumber\\
    &\hspace{0.4cm} + \EE\left[\max_{z \in \mathcal{C}}\sum\limits_{k=0}^{K-1} \left(\| w^{k+1}  - z\|^2 - (1 - \tau)\|z^{k} - z \|^2 - \tau\| w^k  - z\|^2\right)\right]  \nonumber\\
    &\hspace{0.4cm} -  (1-\tau - 2\gamma^2 L^2 - 13200(\delta^{\text{serv}})^2(\delta^{\text{dev}})^2\gamma^2 \tilde L^2)\cdot \sum\limits_{k=0}^{K-1} \EE\| w^k - z^{k+1/2} \|^2  \nonumber\\
    &\hspace{0.4cm}  - \left(\tau - \frac{1}{2}\right) \cdot \sum\limits_{k=0}^{K-1} \EE\| z^{k+1/2} - z^{k}\|^2.
\end{align*}
With $t \geq \frac{3}{4}$ and $\gamma \leq \frac{\sqrt{1-\tau}}{2L + 165 \delta^{\text{serv}}\delta^{\text{dev}} \tilde L}$ we get
\begin{align}
    \label{t506}
    2 \gamma &\EE\left[\max_{z \in \mathcal{C}}\sum\limits_{k=0}^{K-1} \la F(z) ,z^{k+1/2} - z\ra \right] \nonumber\\
    &\leq \EE\left[\max_{z \in \mathcal{C}}\|\hat z^{0} - z\|^2 \right] + \EE\left[\max_{z \in \mathcal{C}}\|w^{0} - z\|^2 \right] 
    \nonumber\\
    &\hspace{0.4cm} + \EE\left[\max_{z \in \mathcal{C}}\sum\limits_{k=0}^{K-1} \left(\| w^{k+1}  - z\|^2 - (1 - \tau)\|z^{k} - z \|^2 - \tau\| w^k  - z\|^2\right)\right].
\end{align}

Let us estimate $\EE\left[\max_{z \in \mathcal{C}}\sum\limits_{k=0}^{K-1} \left(\| w^{k+1}  - z\|^2 - (1 - \tau)\|z^{k} - z \|^2 - \tau\| w^k  - z\|^2\right)\right]$. For this we note that
\begin{align*}
    \EE&\left[\max_{z \in \mathcal{C}} \sum\limits_{k=0}^{K-1} \left[ -\tau \| w^k - z \|^2 - (1 - \tau) \| z^{k} - z\|^2 + \| w^{k+1} - z\|^2\right] \right] \\
    &= \EE\left[\max_{z \in \mathcal{C}} \sum\limits_{k=0}^{K-1} \left[ -2 \langle (1 - \tau) z^{k} + \tau w^k - w^{k+1}, z\rangle - (1 -\tau)\|z^{k} \|^2 - \tau \| w^k\|^2 + \|w^{k+1} \|^2\right] \right] \\
    &= \EE\left[\max_{z \in \mathcal{C}} \sum\limits_{k=0}^{K-1} \left[ -2 \langle (1 - \tau) z^{k} + \tau w^k - w^{k+1}, z\rangle\right]\right] \\
    &\hspace{0.4cm} +\EE\left[\sum\limits_{k=0}^{K-1} -(1 -\tau)\|z^{k} \|^2 - \tau \| w^k\|^2 + \|w^{k+1} \|^2\right] .
\end{align*}
One can note that by definition $w^{k+1}$: $\EE\left[ (1 -\tau)\|z^{k} \|^2 + \tau \| w^k\|^2 - \|w^{k+1} \|^2\right] = 0$, then 
\begin{align*}
    \EE&\left[\max_{z \in \mathcal{C}} \sum\limits_{k=0}^{K-1} \left[ -\tau \| w^k - z \|^2 - (1 - \tau) \| z^{k} - z\|^2 + \| w^{k+1} - z\|^2\right] \right] \\
    &\hspace{4cm}= 2\EE\left[\max_{z \in \mathcal{C}} \sum\limits_{k=0}^{K-1} \langle (1 - \tau) z^{k} + \tau w^k - w^{k+1}, -z\rangle\right] 
    \\
    &\hspace{4cm}= 2\EE\left[\max_{z \in \mathcal{C}} \sum\limits_{k=0}^{K-1} \langle (1 - \tau) z^{k} + \tau w^k - w^{k+1}, z\rangle\right] .
\end{align*}
Let define sequence $v$: $v^0 = z^{0}$, $v^{k+1} = v^k-\delta_k$ with $\delta^k = (1 - \tau) z^{k} + \tau w^k - w^{k+1}$. Then we have
\begin{align}
    \label{t44412}
    \sum\limits_{k=0}^{K-1} \langle \delta^k, z^{k+1/2} - z \rangle = \sum\limits_{k=0}^{K-1} \langle \delta^k, z^{k+1/2} - v^k \rangle + 
    \sum\limits_{k=0}^{K-1} \langle \delta^k,  v^k - z \rangle . 
\end{align}
By the definition of $v^{k+1}$, we have for all $z$
\begin{align*}
    \langle v^{k+1} - v^{k} +\delta^k, z - v^{k+1} \rangle = 0.
\end{align*}
Rewriting this inequality, we get
\begin{align*}
    \langle \delta^k, v^k  - z \rangle &= \langle \delta^k, v^k - v^{k+1} \rangle  + \langle v^{k+1} - v^k, z - v^{k+1} \rangle \nonumber\\
&= \langle \delta^k, v^k - v^{k+1} \rangle + \frac{1}{2}\|v^k - z\|^2 -  \frac{1}{2}\|v^{k+1} - z\|^2 - \frac{1}{2}\| v^k - v^{k+1}\|^2
\nonumber\\
&= \frac{1}{2} \|\delta^k\|^2  + \frac{1}{2}\|v^k - v^{k+1}\|^2 + \frac{1}{2}\|v^k - z\|^2 -  \frac{1}{2}\|v^{k+1} - z\|^2 - \frac{1}{2}\| v^k - v^{k+1}\|^2
\nonumber\\
&= \frac{1}{2} \|\delta^k\|^2  + \frac{1}{2}\|v^k - z\|^2 -  \frac{1}{2}\|v^{k+1} - z\|^2 .
\end{align*}
With \eqref{t44412} it gives
\begin{align*}
    \sum\limits_{k=0}^{K-1} \langle \delta^k, z^{k+1/2} - z \rangle &\leq \sum\limits_{k=0}^{K-1} \langle \delta^k, z^{k+1/2} - v^k \rangle + 
    \sum\limits_{k=0}^{K-1} \left(\frac{1}{2} \|\delta^k\|^2  + \frac{1}{2}\|v^k - z\|^2 -  \frac{1}{2}\|v^{k+1} - z\|^2 \right) \nonumber\\
    &\leq \sum\limits_{k=0}^{K-1} \langle \delta^k, z^{k+1/2} - v^k \rangle + 
    \frac{1}{2}\sum\limits_{k=0}^{K-1} \|\delta^k\|^2 + \frac{1}{2}\|z^0 - z\|^2. 
\end{align*}
We take the maximum on  $z$ and get
\begin{align*}
     \max_{z \in \mathcal{C}} \sum\limits_{k=0}^{K-1} \langle \delta^k, z^{k+1/2} - z \rangle &\leq \sum\limits_{k=0}^{K-1} \langle \delta^k, z^{k+1/2} - v^k \rangle \nonumber\\
     &\hspace{0.4cm}+
    \frac{1}{2}\sum\limits_{k=0}^{K-1} \|(1 - \tau) z^{k} + \tau w^k - w^{k+1}\|^2 + \frac{1}{2} \max_{z \in \mathcal{C}}  \|z^0 - z\|^2.  
\end{align*}
Taking the full expectation, we get
\begin{align*}
     \EE&\left[ \max_{z \in \mathcal{C}} \sum\limits_{k=0}^{K-1} \langle \delta^k, z^{k+1/2} - z \rangle\right] \leq \EE\left[\sum\limits_{k=0}^{K-1} \langle \delta^k, z^{k+1/2} - v^k \rangle\right] \nonumber\\
    &\hspace{0.4cm}+ 
    \frac{1}{2}\sum\limits_{k=0}^{K-1} \EE\left[\|(1 - \tau) z^{k} + \tau w^k - w^{k+1}\|^2 \right] +\frac{1}{2}\EE\left[\max_{z \in \mathcal{C}} \|z^0 - z\|^2\right] \nonumber\\
    &= \EE\left[\sum\limits_{k=0}^{K-1} \langle \EE_{w^{k+1}}\left[(1 - \tau) z^{k} + \tau w^k - w^{k+1} \right], z^{k+1/2} - v^k \rangle\right] \nonumber\\
    &\hspace{0.4cm}+ 
    \frac{1}{2}\sum\limits_{k=0}^{K-1} \EE\left[\|(1 - \tau) z^{k} + \tau w^k - w^{k+1}\|^2 \right]+ \frac{1}{2}\EE\left[\max_{z \in \mathcal{C}} \|z^0 - z\|^2\right] \nonumber\\
    &= \frac{1}{2}\sum\limits_{k=0}^{K-1} \EE\left[\|(1 - \tau) z^{k} + \tau w^k - w^{k+1}\|^2 \right]+ \frac{1}{2}\EE\left[\max_{z \in \mathcal{C}} \|z^0 - z\|^2\right]\notag\\
    &\leq \frac{1}{2}\sum\limits_{k=0}^{K-1} \EE\left[\|(1 - \tau) z^{k} + \tau w^k - w^{k+1}\|^2 \right]+ \frac{1}{2}\EE\left[\max_{z \in \mathcal{C}} \|z^0 - z\|^2\right] \notag\\
    &= \frac{1}{2}\sum\limits_{k=0}^{K-1} \EE\left[\|\EE_{w^{k+1}} [w^{k+1}] - w^{k+1}\|^2 \right]+ \frac{1}{2}\EE\left[\max_{z \in \mathcal{C}} \|z^0 - z\|^2\right] \notag\\
    &= \frac{1}{2}\sum\limits_{k=0}^{K-1} \EE\left[-\|\EE_{w^{k+1}} [w^{k+1}] \|^2  + \EE_{w^{k+1}} \| w^{k+1}\|^2 \right]+ \frac{1}{2}\EE\left[\max_{z \in \mathcal{C}} \|z^0 - z\|^2\right] \notag\\
    &= \frac{1}{2}\sum\limits_{k=0}^{K-1} \EE\left[-\|(1 - \tau) z^{k} + \tau w^k \|^2  + (1 - \tau)\| z^{k}\|^2 + \tau\| w^{k}\|^2 \right]+ \frac{1}{2}\EE\left[\max_{z \in \mathcal{C}} \|z^0 - z\|^2\right] \notag\\
    &= \frac{1}{2}\sum\limits_{k=0}^{K-1} \tau(1 - \tau)\EE\left[\| z^{k} - w^k \|^2 \right]+ \frac{1}{2}\EE\left[\max_{z \in \mathcal{C}} \|z^0 - z\|^2\right].
\end{align*}
Finally, we have
\begin{align}
    \label{t405}
    \EE&\left[\max_{z \in \mathcal{C}} \sum\limits_{k=0}^{K-1} \left[ -\tau \| w^k - z \|^2 - (1 - \tau) \| z^{k} - z\|^2 + \| w^{k+1} - z\|^2\right] \right] \notag\\
    &\leq \sum\limits_{k=0}^{K-1} \tau(1 - \tau)\EE\left[\| z^{k} - w^k \|^2 \right]+ \max_{z \in \mathcal{C}} \|z^0 - z\|^2.
\end{align}
Together with \eqref{t506} we obtain
\begin{align}
\label{t508}
    2 \gamma &\EE\left[\max_{z \in \mathcal{C}}\sum\limits_{k=0}^{K-1} \la F(z) ,z^{k+1/2} - z\ra \right] \nonumber\\
    &\leq \EE\left[\max_{z \in \mathcal{C}}\|\hat z^{0} - z\|^2 \right] + \EE\left[\max_{z \in \mathcal{C}}\|w^{0} - z\|^2 \right] 
    \nonumber\\
    &\hspace{0.4cm} + \EE\left[\max_{z \in \mathcal{C}} \|z^0 - z\|^2\right] + (1 - \tau)\cdot \sum\limits_{k=0}^{K-1} \EE\left[\| z^{k} - w^k \|^2  \right] \nonumber\\
    &\leq \EE\left[\max_{z \in \mathcal{C}}\|\hat z^{0} - z\|^2 \right] + \EE\left[\max_{z \in \mathcal{C}}\|w^{0} - z\|^2 \right]  + \EE\left[\max_{z \in \mathcal{C}} \|z^0 - z\|^2\right] \nonumber\\
    & \hspace{0.4cm} + 2(1 - \tau)\cdot \sum\limits_{k=0}^{K-1} \left(\EE\left[\| z^{k+1/2} - w^k \|^2  \right] + \EE\left[\| z^{k+1/2} - z^k \|^2  \right] \right)
    .
\end{align}
Let us use \eqref{t507} with $p=1$ and $\mu =0$ (monotone case):
\begin{align*}
 \sum\limits_{k=0}^{K-1} & \left(\EE\|\hat z^{k+1} - z^* \|^2 + \EE\|w^{k+1} - z^* \|^2\right) \nonumber\\
    &\leq \sum\limits_{k=0}^{K-1} \left(\EE\|\hat z^{k} - z^* \|^2 + \EE\|w^{k} - z^* \|^2\right) \nonumber\\
    & \hspace{0.4cm}-  (1-\tau - 2\gamma^2 L^2 - 13200(\delta^{\text{serv}})^2 (\delta^{\text{dev}})^2 \gamma^2 \tilde L^2  ) \cdot \sum\limits_{k=0}^{K-1} \EE\| w^k - z^{k+1/2} \|^2 \nonumber\\
    &\hspace{0.4cm}- \left(\tau - \frac{1}{2}\right)\cdot \sum\limits_{k=0}^{K-1} \EE\| z^{k+1/2} - z^{k}\|^2 .
\end{align*}
Taking into account that $\gamma \leq \frac{\sqrt{1-\tau}}{2L + 165 \delta^{\text{serv}}\delta^{\text{dev}} \tilde L}$ we get
\begin{align*}
 \sum\limits_{k=0}^{K-1} \left(\EE\|\hat z^{k+1} - z^* \|^2 + \EE\|w^{k+1} - z^* \|^2\right)  
    &\leq \sum\limits_{k=0}^{K-1} \left(\EE\|\hat z^{k} - z^* \|^2 + \EE\|w^{k} - z^* \|^2\right)
    \nonumber\\
    &\hspace{0.4cm}-  \frac{1-\tau}{2} \cdot \sum\limits_{k=0}^{K-1} \left(\EE\| w^k - z^{k+1/2} \|^2
    +\EE\| z^{k+1/2} - z^{k}\|^2 \right).
\end{align*}
Small rearrangement gives
\begin{align*}
  2(1-\tau) \cdot \sum\limits_{k=0}^{K-1} \left(\EE\| w^k - z^{k+1/2} \|^2
    +\EE\| z^{k+1/2} - z^{k}\|^2 \right)
    &\leq 4\left(\EE\|\hat z^{0} - z^* \|^2 + \EE\|w^{0} - z^* \|^2\right).
\end{align*}
Substituting this expression to \eqref{t508}, we get:
\begin{align*}
    2 \gamma \EE\left[\max_{z \in \mathcal{C}}\sum\limits_{k=0}^{K-1} \la F(z) ,z^{k+1/2} - z\ra \right] 
    &\leq \EE\left[\max_{z \in \mathcal{C}}\|\hat z^{0} - z\|^2 \right] + 2\EE\left[\max_{z \in \mathcal{C}} \|z^0 - z\|^2\right] 
    \nonumber\\
    &\hspace{0.4cm}
    + 4\left(\EE\|\hat z^{0} - z^* \|^2 + \EE\|w^{0} - z^* \|^2\right)
    .
\end{align*}
Then we can obtain
\begin{align*}
    \EE\left[\max_{z \in \mathcal{C}} \la F(z) , \left(\frac{1}{K}\sum\limits_{k=0}^{K-1} z^{k+1/2}\right) - z\ra \right] 
    &\leq \frac{2\max_{z \in \mathcal{C}}\| z^{0} - z\|^2 + 4\| z^{0} - z^*\|^2}{\gamma K},
\end{align*}
and finish the proof.
\EndProof

\subsubsection{Non-monotone}

Again we start from \eqref{t22}:
\begin{align*}
    \EE\|\hat z^{k+1} - z\|^2
    &\leq \EE\|\hat z^{k} - z\|^2 - (1 - \tau)\EE\|z^{k} - z \|^2 + (1 - \tau)\EE\| w^k  - z\|^2  \nonumber\\
    &\hspace{0.4cm}-2 \gamma \EE\left[\la F(z^{k+1/2}) ,z^{k+1/2} - z\ra \right]-  (1-\tau - 2\gamma^2 L^2)\EE\| w^k - z^{k+1/2} \|^2 \nonumber\\
    &\hspace{0.4cm} +  6 \EE\| e^k\|^2 +  \frac{6}{M} \sum\limits_{m=1}^M \EE\left\| e^{k}_m \right\|^2 
    - \left(\tau - \frac{1}{2}\right)\EE\| z^{k+1/2} - z^{k}\|^2.
\end{align*}
Putting $z = z^*$ and using non-monotonicity (Assumption \ref{as1} (NM)), we get
\begin{align}
    \label{t6123}
    \EE\|\hat z^{k+1} - z^*\|^2
    &\leq \EE\|\hat z^{k} - z^*\|^2 - (1 - \tau)\EE\|z^{k} - z^* \|^2 + (1 - \tau)\EE\| w^k  - z^*\|^2  \nonumber\\
    &\hspace{0.4cm}-  (1-\tau - 2\gamma^2 L^2)\EE\| w^k - z^{k+1/2} \|^2 \nonumber\\
    &\hspace{0.4cm} +  6 \EE\| e^k\|^2 +  \frac{6}{M} \sum\limits_{m=1}^M \EE\left\| e^{k}_m \right\|^2 
    - \left(\tau - \frac{1}{2}\right)\EE\| z^{k+1/2} - z^{k}\|^2.
\end{align}
With $\tau \geq \frac{3}{4}$ and rule for $z^{k+1/2}$ (line \ref{alg2:z1/2}) we obtain
\begin{align*}
    \EE\|\hat z^{k+1} - z^*\|^2
    &\leq \EE\|\hat z^{k} - z^*\|^2 - (1 - \tau)\EE\|z^{k} - z^* \|^2 + (1 - \tau)\| w^k  - z^*\|^2  \nonumber\\
    &\hspace{0.4cm}-  (1-\tau - 2\gamma^2 L^2)\EE\| w^k - z^{k+1/2} \|^2 \nonumber\\
    &\hspace{0.4cm} +  6 \EE\| e^k\|^2 +  \frac{6}{M} \sum\limits_{m=1}^M \EE\left\| e^{k}_m \right\|^2 
    - \frac{1}{4}\| z^{k+1/2} - z^{k}\|^2 \\
    &= \EE\|\hat z^{k} - z^*\|^2 - (1 - \tau)\EE\|z^{k} - z^* \|^2 + (1 - \tau)\EE\| w^k  - z^*\|^2  \nonumber\\
    &\hspace{0.4cm}-  (1-\tau - 2\gamma^2 L^2)\EE\| w^k - z^{k+1/2} \|^2 - \frac{1}{8}\EE\| z^{k+1/2} - z^{k}\|^2\nonumber\\
    &\hspace{0.4cm} +  6 \EE\| e^k\|^2 +  \frac{6}{M} \sum\limits_{m=1}^M \EE\left\| e^{k}_m \right\|^2 
    - \frac{1}{8}\EE\| \tau z^k + (1 - \tau) w^k - \gamma  F(w^k) - z^{k}\|^2 \\
    &= \EE\|\hat z^{k} - z^*\|^2 - (1 - \tau)\EE\|z^{k} - z^* \|^2 + (1 - \tau)\EE\| w^k  - z^*\|^2  \nonumber\\
    &\hspace{0.4cm}-  (1-\tau - 2\gamma^2 L^2)\EE\| w^k - z^{k+1/2} \|^2 - \frac{1}{8}\EE\| z^{k+1/2} - z^{k}\|^2\nonumber\\
    &\hspace{0.4cm} +  6 \EE\| e^k\|^2 +  \frac{6}{M} \sum\limits_{m=1}^M \EE\left\| e^{k}_m \right\|^2 
    - \frac{1}{8}\EE\|(1 - \tau) (w^k - z^k) - \gamma  F(w^k)\|^2.
\end{align*}
Using $-\| a\|^2 \leq -\frac{1}{2}\| a + b\|^2 + \| b\|^2$ gives
\begin{align*}
    \EE\|\hat z^{k+1} - z^*\|^2
    &\leq \EE\|\hat z^{k} - z^*\|^2 - (1 - \tau)\EE\|z^{k} - z^* \|^2 + (1 - \tau)\EE\| w^k  - z^*\|^2  \nonumber\\
    &\hspace{0.4cm}-  (1-\tau - 2\gamma^2 L^2)\EE\| w^k - z^{k+1/2} \|^2 - \frac{1}{8}\EE\| z^{k+1/2} - z^{k}\|^2\nonumber\\
    &\hspace{0.4cm} +  6 \EE\| e^k\|^2 +  \frac{6}{M} \sum\limits_{m=1}^M \EE\left\| e^{k}_m \right\|^2 
    - \frac{\gamma^2}{16}\EE\| F(w^k)\|^2 + \frac{(1 - \tau)^2}{8} \EE\|w^k - z^k\|^2.
\end{align*}
And then
\begin{align*}
    \EE\|\hat z^{k+1} - z^*\|^2
    &\leq \EE\|\hat z^{k} - z^*\|^2 - (1 - \tau)\EE\|z^{k} - z^* \|^2 + (1 - \tau)\EE\| w^k  - z^*\|^2  \nonumber\\
    &\hspace{0.4cm}-  (1-\tau - 2\gamma^2 L^2)\EE\| w^k - z^{k+1/2} \|^2 - \frac{1}{8}\EE\| z^{k+1/2} - z^{k}\|^2\nonumber\\
    &\hspace{0.4cm} +  6 \EE\| e^k\|^2 +  \frac{6}{M} \sum\limits_{m=1}^M \EE\left\| e^{k}_m \right\|^2 
    - \frac{\gamma^2}{16}\EE\| F(w^k)\|^2 + \frac{(1 - \tau)^2}{4} \EE\|z^{k+1/2} - z^k\|^2 \\
    &\hspace{0.4cm} + \frac{(1 - \tau)^2}{4} \EE\|z^{k+1/2} - w^k\|^2\\
    &\leq \EE\|\hat z^{k} - z^*\|^2 - (1 - \tau)\EE\|z^{k} - z^* \|^2 + (1 - \tau)\EE\| w^k  - z^*\|^2  \nonumber\\
    &\hspace{0.4cm}-  (1-\tau - 2\gamma^2 L^2)\EE\| w^k - z^{k+1/2} \|^2 - \frac{1}{8}\EE\| z^{k+1/2} - z^{k}\|^2\nonumber\\
    &\hspace{0.4cm} +  6 \EE\| e^k\|^2 +  \frac{6}{M} \sum\limits_{m=1}^M \EE\left\| e^{k}_m \right\|^2 
    - \frac{\gamma^2}{16}\EE\| F(w^k)\|^2 + \frac{1}{64} \EE\|z^{k+1/2} - z^k\|^2 \\
    &\hspace{0.4cm} + \frac{1-\tau}{16} \EE\|z^{k+1/2} - w^k\|^2\\
    &\leq \EE\|\hat z^{k} - z^*\|^2 - (1 - \tau)\EE\|z^{k} - z^* \|^2 + (1 - \tau)\EE\| w^k  - z^*\|^2  \nonumber\\
    &\hspace{0.4cm}-  \left(\frac{1-\tau}{2} - 2\gamma^2 L^2\right)\EE\| w^k - z^{k+1/2} \|^2 \nonumber\\
    &\hspace{0.4cm} +  6 \EE\| e^k\|^2 +  \frac{6}{M} \sum\limits_{m=1}^M \EE\left\| e^{k}_m \right\|^2 
    - \frac{\gamma^2}{16}\EE\| F(w^k)\|^2.
\end{align*}
Here we additionally use that $\tau \geq \frac{3}{4}$. 
Then we add \eqref{twk}
\begin{align*}
    \EE\|\hat z^{k+1} - z^*\|^2 &+ \EE\|w^{k+1} - z^*\|^2 \\
    &\leq \EE\|\hat z^{k} - z^*\|^2 + \EE\| w^k  - z^*\|^2  -  \left(\frac{1-\tau}{2} - 2\gamma^2 L^2\right)\EE\| w^k - z^{k+1/2} \|^2 \nonumber\\
    &\hspace{0.4cm} +  6 \EE\| e^k\|^2 +  \frac{6}{M} \sum\limits_{m=1}^M \EE\left\| e^{k}_m \right\|^2 
    - \frac{\gamma^2}{16}\EE\| F(w^k)\|^2.
\end{align*}
Next, we sum over all $k$ from $0$ to $K-1$ and get
\begin{align*}
    \frac{\gamma^2}{16}\sum\limits_{k=0}^{K-1}\EE\| F(w^k)\|^2
    &\leq \EE\|\hat z^{0} - z^*\|^2 + \EE\|w^{0} - z^*\|^2 - \EE\|\hat z^{K} - z^*\|^2 - \EE\| w^{K}  - z^*\|^2  \nonumber\\
    &\hspace{0.4cm}-  \left(\frac{1-\tau}{2} - 2\gamma^2 L^2\right)\sum\limits_{k=0}^{K-1}\EE\| w^k - z^{k+1/2} \|^2 \\
    &\hspace{0.4cm}+  6 \sum\limits_{k=0}^{K-1}\EE\| e^k\|^2 +  6\sum\limits_{k=0}^{K-1}\frac{1}{M} \sum\limits_{m=1}^M \EE\left\| e^{k}_m \right\|^2.
\end{align*}
It remains to use \eqref{t11} and \eqref{t111} with $p=1$:
\begin{align*}
    \frac{\gamma^2}{16}\sum\limits_{k=0}^{K-1}\EE\| F(w^k)\|^2
    &\leq \EE\|\hat z^{0} - z^*\|^2 + \EE\|w^{0} - z^*\|^2 - \EE\|\hat z^{K} - z^*\|^2 - \EE\| w^{K}  - z^*\|^2  \nonumber\\
    &\hspace{0.4cm}-  \left(\frac{1-\tau}{2} - 2\gamma^2 L^2\right)\sum\limits_{k=0}^{K-1}\EE\| w^k - z^{k+1/2} \|^2 \\
    &\hspace{0.4cm}+  768(\delta^{\text{serv}})^2\gamma^2 \tilde L^2 \sum\limits_{k=0}^{K-1}  \EE\left\|  z^{k+1/2} - w^k  \right\|^2  \\
    &\hspace{0.4cm}+ 768(\delta^{\text{serv}})^2 \sum\limits_{k=0}^{K-1} \frac{1}{M} \sum\limits_{m=1}^M\EE\left\|e^k_m \right\|^2 +  6\sum\limits_{k=0}^{K-1}\frac{1}{M} \sum\limits_{m=1}^M \EE\left\| e^{k}_m \right\|^2 \\
    &\leq \EE\|\hat z^{0} - z^*\|^2 + \EE\|w^{0} - z^*\|^2 \nonumber\\
    &\hspace{0.4cm}-  \left(\frac{1-\tau}{2} - 2 \gamma^2 L^2 - 768(\delta^{\text{serv}})^2\gamma^2 \tilde L^2\right)\sum\limits_{k=0}^{K-1}\EE\| w^k - z^{k+1/2} \|^2 \\
    &\hspace{0.4cm}+ 775(\delta^{\text{serv}})^2 \sum\limits_{k=0}^{K-1} \frac{1}{M} \sum\limits_{m=1}^M\EE\left\|e^k_m \right\|^2\\
    &\leq \EE\|\hat z^{0} - z^*\|^2 + \EE\|w^{0} - z^*\|^2 \nonumber\\
    &\hspace{0.4cm}-  \left(\frac{1-\tau}{2} - 2 \gamma^2 L^2 - 768(\delta^{\text{serv}})^2\gamma^2 \tilde L^2\right)\sum\limits_{k=0}^{K-1}\EE\| w^k - z^{k+1/2} \|^2 \\
    &\hspace{0.4cm}+ 12400(\delta^{\text{serv}})^2 (\delta^{\text{dev}})^2 \gamma^2 \tilde L^2 \sum\limits_{k=0}^{K-1}  \EE\left\|  z^{k+1/2} - w^k  \right\|^2 \\
    &\leq \EE\|\hat z^{0} - z^*\|^2 + \EE\|w^{0} - z^*\|^2 \nonumber\\
    &\hspace{0.4cm}-  \frac{1}{2}\left(1-\tau - 4 \gamma^2 L^2 - 26400(\delta^{\text{serv}})^2(\delta^{\text{dev}})^2\gamma^2 L^2\right)\sum\limits_{k=0}^{K-1}\EE\| w^k - z^{k+1/2} \|^2.
\end{align*}
Then we choose $\gamma \leq \frac{\sqrt{1-\tau}}{2L + 165\delta^{\text{serv}}\delta^{\text{dev}}\tilde L}$ and get
\begin{align*}
    \frac{1}{K}\sum\limits_{k=0}^{K-1}\EE\| F(w^k)\|^2
    &\leq \frac{16(\EE\|\hat z^{0} - z^*\|^2 + \EE\|w^{0} - z^*\|^2)}{\gamma^2 K}.
\end{align*}
\EndProof

\newpage

\section{Stochastic case and variance reduction} \label{sec:stoch}

In this subsection, we assume that the local operators on each node has either a finite-sum form:
\begin{equation}
    \label{MK_fs}
  \textstyle    
  F_m(z) \eqdef   \frac{1}{r}\sum\limits_{i=1}^r F_{m,i}(z).
\end{equation}
This case corresponds to the stochastic setting, when it is expensive to calculate the full operator $F_m$, and it is cheaper to calculate the value $F_{m,i}$ one of the terms (batches). For this setup we additionally assume that
\begin{assumption}\label{as5}
Each operator $F_{m,i}$ is $L_{m,i}$-Lipschitz continuous, i.e. for all $z_1, z_2 \in \R^d$ it holds
\begin{align}
\label{Fmilip}
  \|F_{m,i}(z_1) - F_{m,i}(z_2)\|^2 \leq L_{m,i} \|z_1-z_2\|.
\end{align}
Let $\tilde L_m^2 = \frac{1}{r} \sum\limits_{i=1}^r L_{m,i}^2$ and $\hat L^2 = \frac{1}{M} \sum\limits_{m=1}^M \frac{1}{r} \sum\limits_{i=1}^r L_{m,i}^2$.
\end{assumption}
Next, we  modify \algname{MASHA1} and \algname{MASHA2} for this setup. 
Modifications of the other steps (computing $g^k$, $z^{k+1}$, $e^k_m$, $e^k$ etc.) in \algname{VR-MASHA1} and \algname{VR-MASHA2} occur according to the new $g^k_m$.


\subsection{\algnamenormal{VR-MASHA1}: stochastic and batch version} \label{sec:vr_masha1}

In this section, we provide information about \algname{VR-MASHA1}. This is a modification of \algname{MASHA1} for the stochastic case of a finite sum. Changes compared to  \algname{MASHA1} are highlighted in blue -- see Algorithm \ref{alg_qeg_vr}.  Note that without compression  \algname{VR-MASHA1} is an analogue of methods from \cite{alacaoglu2021stochastic}.

\begin{algorithm}[th]
	\caption{\algname{VR-MASHA1}}
	\label{alg_qeg_vr}
	\begin{algorithmic}[1]
\State
\noindent {\bf Parameters:}  Stepsize $\gamma>0$, parameter $\tau$, number of iterations $K$.\\
\noindent {\bf Initialization:} Choose  $z^0 = w^0 \in \mathcal{Z}$. \\
Server  sends to devices $z^0 = w^0$ and devices compute $F_m(w^0)$ and send to server and get $F(w^0)$
\For {$k=0,1, 2, \ldots, K-1$ }
\For {each device $m$ in parallel}
\State $\bar z^k = \tau z^k + (1 - \tau) w^k$
\State $z^{k+1/2} = \bar z_k - \gamma F(w^k)$,
\State {\color{blue} Generate $\pi^{k}_m$ from $\{1, \ldots, r\}$ independently} 
\State  Compute $F_{m, {\color{blue}\pi^{k}_m}}(z^{k+1/2})$ \&  send $Q^{\text{dev}}_m(F_{m, {\color{blue}\pi^{k}_m}}(z^{k+1/2}) - F_{m, {\color{blue}\pi^{k}_m}}(w^k))$ to server
\EndFor
\For {server}
\State Compute $Q^{\text{serv}} \left [\frac{1}{M} \sum \limits_{m=1}^M Q^{\text{dev}}_m(F_{m, {\color{blue}\pi^{k}_m}}(z^{k+1/2}) - F_{m, {\color{blue}\pi^{k}_m}}(w^k))\right]$ \& send to devices
\State Sends to devices one bit $b_k$: 1 with probability $1 - \tau$, 0 with with probability $\tau$
\EndFor
\For {each device $m$ in parallel}
\State $z^{k+1} = z^{k+1/2} - \gamma  Q^{\text{serv}}\left[\frac{1}{M} \sum \limits_{m=1}^M Q^{\text{dev}}_m(F_{m, {\color{blue}\pi^{k}_m}}(z^{k+1/2}) - F_{m, {\color{blue}\pi^{k}_m}}(w^k))\right] $
\If {$b_k = 1$}
\State $w^{k+1} = z^{k+1}$
\State Compute $F_m(w^{k+1})$ \& send it to server; and get $F(w^{k+1})$ as a response from server
\Else 
\State $w^{k+1} = w^k$
\EndIf
\EndFor
\EndFor
	\end{algorithmic}
\end{algorithm}

The following theorem gives the convergence of \algname{VR-MASHA1}. 

\begin{theorem} \label{th_q2}
Let distributed variational inequality \eqref{VI} + \eqref{MK} + \eqref{MK_fs} is solved by Algorithm \ref{alg_qeg_vr} with unbiased compressor operators \eqref{quant}: on server with $q^{\text{serv}}$ parameter, on devices with $\{q^{\text{dev}}_m\}$. Let Assumption \ref{as3} and one case of Assumption \ref{as1} are satisfied. Then the following estimates holds 

$\bullet$ in strongly-monotone case with $\gamma \leq \min\left[\frac{\sqrt{1 - \tau}}{2 \tilde C_q}; \frac{1 - \tau}{2\mu}\right]$ \\ (where $\tilde C_q = \sqrt{\frac{q^{\text{serv}}}{M^2}  \cdot \sum_{m=1}^M (q_m^{\text{dev}} \tilde L_m^2 + (M-1) \tilde L^2)}$):
\begin{align*}
 \EE\left(\| z^{K} - z^* \|^2 + \|w^{K} - z^* \|^2\right) &\leq \left( 1 -\frac{\mu\gamma}{2} \right)^K \cdot 2\|z^{0} - z^* \|^2;
\end{align*}

$\bullet$ in monotone case with $\gamma \leq \frac{\sqrt{1 - \tau}}{2 \tilde C_q + 4 \tilde L}$:
\begin{align*}
\EE\left[\max_{z \in \mathcal{C}} \left[ \langle F(u),  \left(\frac{1}{K}\sum\limits_{k=0}^{K-1} z^{k+1/2}\right) - u  \rangle \right] \right]
\leq  \frac{2\max_{z \in \mathcal{C}}\left[\| z^0 - z \|^2\right] + 6\|z^{0} - z^* \|^2}{\gamma K};
\end{align*}

$\bullet$ in non-monotone case with $\gamma \leq \frac{\sqrt{1 - \tau}}{2 \tilde C_q}$:
\begin{align*}
    \EE\left(\frac{1}{K}\sum\limits_{k=0}^{K-1}\| F(w^k)\|^2\right)
    &\leq \frac{16\|z^{0} - z^* \|^2}{\gamma^2 K}.
\end{align*}
\end{theorem}

For \algname{VR-MASHA1} we consider the case of only devices compression. For simplicity, we put $Q^{\text{dev}}_m = Q$ with $q^{\text{dev}}_m = q$ and $\beta^{\text{dev}}_m = \beta$, also $\tilde L_m = \tilde L = L$.
Let us discuss the difference (with \algname{MASHA1}) in choosing $\tau$. When $b_k = 1$, we need not only to send uncompressed information to the server, but also to compute the full $F_m$, which in the stochastic case is $r$ times more expensive than computing one batch $F_{m,i}$. Then, at each iteration, we send $\mathcal{O}\left( \frac{1}{\beta} + 1-\tau\right)$ bits of information, and also count $\mathcal{O}\left( 1 + r(1-\tau)\right)$ batches. Therefore, the optimal choice of $\tau$ depends on two factors and $1 - \tau = \frac{1}{\max\{\beta, r \}}$.

\begin{corollary} \label{cor:vr_masha1}
Let distributed variational inequality \eqref{VI} + \eqref{MK} + \eqref{MK_fs} is solved by Algorithm \ref{alg_qeg_vr} without compression on server ($q^{\text{serv}} = 1$) and with unbiased compressor operators \eqref{quant} on devices with $\{q^{\text{dev}}_m = q\}$. Let Assumption \ref{as3} and one case of Assumption \ref{as1} are satisfied. Then the following estimates holds 

$\bullet$ in strongly-monotone case with $\gamma \leq \min\left[\frac{1}{2 L} \cdot \left(\sqrt{\frac{q}{M} + 1 } \right)^{-1} \cdot \left(\sqrt{ \max\{\beta, r \}}\right)^{-1} ; \frac{1}{2\mu} \cdot \left( \max\{\beta, r \}\right)^{-1}\right]$:
\begin{align*}
 \EE\left(\| z^{K} - z^* \|^2 + \|w^{K} - z^* \|^2\right) &\leq \left( 1 -\frac{\mu\gamma}{2} \right)^K \cdot 2\|z^{0} - z^* \|^2;
\end{align*}

$\bullet$ in monotone case with $\gamma \leq \frac{1}{6L} \cdot \left(\sqrt{\frac{q}{M} + 1 } \right)^{-1} \cdot \left(\sqrt{ \max\{\beta, r \}}\right)^{-1}$:
\begin{align*}
\EE\left[\max_{z \in \mathcal{C}} \left[ \langle F(u),  \left(\frac{1}{K}\sum\limits_{k=0}^{K-1} z^{k+1/2}\right) - u  \rangle \right] \right]
\leq  \frac{2\max_{z \in \mathcal{C}}\left[\| z^0 - z \|^2\right] + 6\|z^{0} - z^* \|^2}{\gamma K};
\end{align*}

$\bullet$ in non-monotone case with $\gamma \leq \frac{1}{2 L} \cdot \left(\sqrt{\frac{q}{M} + 1 } \right)^{-1} \cdot \left(\sqrt{ \max\{\beta, r \}}\right)^{-1}$:
\begin{align*}
    \EE\left(\frac{1}{K}\sum\limits_{k=0}^{K-1}\| F(w^k)\|^2\right)
    &\leq \frac{16\|z^{0} - z^* \|^2}{\gamma^2 K}.
\end{align*}
\end{corollary}
In the line 5 of  Table \ref{tab:comparison1} we put complexities to achieve $\varepsilon$-solution.

\subsubsection{Proof of the convergence of \algnamenormal{VR-MASHA1}} \label{prof_th5}

\textbf{Proof of Theorem \ref{th_q2}:}

The proof is very close to the proof of Theorem \ref{th_q}. Only two estimates need to be modified. First is \eqref{temp33}
\begin{align*}
\EE\left[\| z^{k+1} -  z^{k+1/2}\|^2\right] &= \gamma^2 \cdot \EE\left[\left\| Q^{\text{serv}} \left [\frac{1}{M} \sum_{m=1}^M Q^{\text{dev}}_m(F_{m, \pi^{k}_m}(z^{k+1/2}) - F_{m, \pi^{k}_m}(w^k))\right]\right\|^2\right]  \\
&\leq \gamma^2 \cdot\frac{q^{\text{serv}}}{M^2} \EE\left[ \left\| \sum_{m=1}^M Q^{\text{dev}}_m(F_{m, \pi^{k}_m}(z^{k+1/2}) - F_{m, \pi^{k}_m}(w^k))\right\|^2\right]  \\
&= \gamma^2 \cdot \frac{q^{\text{serv}}}{M^2} \sum_{m=1}^M \EE\left[\left \|  Q^{\text{dev}}_m(F_{m, \pi^{k}_m}(z^{k+1/2}) - F_{m, \pi^{k}_m}(w^k))\right\|^2\right] \\
&\hspace{0.4cm}+ \gamma^2 \cdot \frac{q^{\text{serv}}}{M^2} \sum_{m \neq l}\EE \left[\langle Q^{\text{dev}}_m (F_{m, \pi^{k}_m}(z^{k+1/2}) - F_{m, \pi^{k}_m}(w^k)) ; Q^{\text{dev}}_m (F_l(z^{k+1/2}) - F_{l, \pi^{k}_l}(w^k)) \rangle \right]
\end{align*}
Next we apply \eqref{quant} and Assumption \ref{as3} for the first term and independence and unbiasedness of $Q$ and uniformess of $\xi$ for the second term:
\begin{align*}
\EE\left[\| z^{k+1} -  z^{k+1/2}\|^2\right] &\leq \gamma^2 \cdot\frac{q^{\text{serv}}}{M^2} \sum_{m=1}^M q_m^{\text{dev}} \frac{1}{r}\sum\limits_{i=1}^r L_{m,i}^2 \EE\left[\left \|  z^{k+1/2} - w^k\right\|^2\right] \nonumber\\
&\hspace{0.4cm}+ \gamma^2 \cdot\frac{q^{\text{serv}}}{M^2} \sum_{m \neq l}\EE \left[\langle F_m(z^{k+1/2}) - F_m(w^k) ; F_l(z^{k+1/2}) - F_l(w^k) \rangle \right] \nonumber\\
&\leq \gamma^2 \cdot\frac{q^{\text{serv}}}{M^2} \sum_{m=1}^M q_m^{\text{dev}} \tilde L_m^2 \EE\left[\left \|  z^{k+1/2} - w^k\right\|^2\right] \nonumber\\
&\hspace{0.4cm}+ \gamma^2 \cdot\frac{q^{\text{serv}}}{2M^2} \sum_{m \neq l}\EE \left[\| F_m(z^{k+1/2}) - F_m(w^k) \|^2 +  \| F_l(z^{k+1/2}) - F_l(w^k) \|^2 \right] \nonumber\\
&\leq \gamma^2 \cdot\frac{q^{\text{serv}}}{M^2} \sum_{m=1}^M q_m^{\text{dev}} \tilde L_m^2 \EE\left[\left \|  z^{k+1/2} - w^k\right\|^2\right] \nonumber\\
&\hspace{0.4cm}+ \gamma^2 \cdot\frac{q^{\text{serv}}}{2M^2} \sum_{m \neq l}  \EE \left[L^2_m \| z^{k+1/2} - w^k\|^2 +  L_l^2\| z^{k+1/2} - w^k \|^2 \right] \nonumber\\
&= \gamma^2 \cdot\frac{q^{\text{serv}}}{M^2} \sum_{m=1}^M q_m^{\text{dev}} \tilde L_m^2 \EE\left[\left \|  z^{k+1/2} - w^k\right\|^2\right] \nonumber\\
&\hspace{0.4cm}
+ \gamma^2 \cdot\frac{q^{\text{serv}} (M-1)}{M} \tilde L^2 \EE \left[\| z^{k+1/2} - w^k\|^2\right]
\nonumber\\
&= \gamma^2 \cdot\frac{q^{\text{serv}}}{M^2} \EE \left[\| z^{k+1/2} - w^k\|^2\right] \cdot \sum_{m=1}^M q_m^{\text{dev}} \tilde L_m^2 + (M-1) \tilde L^2
\end{align*}
Here we can use new $\tilde C_q = \sqrt{\frac{q^{\text{serv}}}{M^2}  \cdot \sum_{m=1}^M (q_m^{\text{dev}} \tilde L_m^2 + (M-1) \tilde L^2)}$.

The second modified estimate is \eqref{temp35}:
\begin{align*}
\EE&\left[\left\|Q^{\text{serv}} \left [\frac{1}{M} \sum_{m=1}^M Q^{\text{dev}}_m(F_m(z^{k+1/2}) - F_m(w^k))\right] + F(w^k) - F(z^{k+1/2})\right\|^2 \right]
\nonumber\\
&\leq \tilde C_q^2 \EE\left[\left\|z^{k+1/2} - w^k \right\|^2 \right] + 2\tilde L^2 \EE\left[\left\|  z^{k+1/2} - w^k\right\|^2 \right].
\end{align*}
\EndProof

\newpage

\subsection{\algnamenormal{VR-MASHA2}: stochastic and batch version}

In this section, we provide information about \algname{VR-MASHA2}. This is a modification of \algname{MASHA2} for the stochastic case of a finite sum. Changes compared to  \algname{MASHA2} are highlighted in blue -- see Algorithm \ref{alg_qfbvr}.

\begin{algorithm}[th]
	\caption{\algname{VR-MASHA2}}
	\label{alg_qfbvr}
	\begin{algorithmic}[1]
\State
\noindent {\bf Parameters:}  Stepsize $\gamma>0$, parameter $\tau$, number of iterations $K$.\\
\noindent {\bf Initialization:} Choose  $z^0 = w^0 \in \mathcal{Z}$, $e^0_m = 0$, $e^0 = 0$. \\
Server  sends to devices $z^0 = w^0$ and devices compute $F_m(w^0)$ and send to server and get $F(w^0)$
\For {$k=0,1, 2, \ldots, K-1$ }
\For {each device $m$ in parallel}
\State $\bar z^k = \tau z^k + (1 - \tau) w^k$ 
\State $z^{k+1/2} = \bar z_k - \gamma F(w^k)$ 
\State {\color{blue} Generate $\pi^{k}_m$ from $\{1, \ldots, r\}$ independently}
\State Compute $F_{m, {\color{blue}\pi^{k}_m}}(z^{k+1/2})$ and send $C^{\text{dev}}_m(\gamma F_{m, {\color{blue}\pi^{k}_m}}(z^{k+1/2}) - \gamma F_{m, {\color{blue}\pi^{k}_m}}(w^k) + e^k_m)$
\State  \text{\small{$e^{k+1}_m = e^k_m + \gamma F_{m, {\color{blue}\pi^{k}_m}}(z^{k+1/2}) - \gamma F_{m, {\color{blue}\pi^{k}_m}}(w^k) - C^{\text{dev}}_m(\gamma F_{m, {\color{blue}\pi^{k}_m}}(z^{k+1/2}) - \gamma  F_{m, {\color{blue}\pi^{k}_m}}(w^k) + e^k_m)$}}
\EndFor
\For {server}
\State  Compute $g^k = C^{\text{serv}} \left [\frac{1}{M} \sum\limits_{m=1}^M C^{\text{dev}}_m(\gamma  F_{m, {\color{blue}\pi^{k}_m}}(z^{k+1/2}) - \gamma  F_{m, {\color{blue}\pi^{k}_m}}(w^k)+ e^k_m )+ e^k\right] $ \& send to devices
\State \text{\small{$e^{k+1} = e^k + \frac{1}{M} \sum\limits_{m=1}^M C^{\text{dev}}_m(\gamma  F_{m, {\color{blue}\pi^{k}_m}}(z^{k+1/2}) - \gamma  F_{m, {\color{blue}\pi^{k}_m}}(w^k)+ e^k_m ) - g^k$}}
\State Sends to devices one bit $b_k$: 1 with probability $1 - \tau$, 0 with with probability $\tau$ 
\EndFor
\For {each device $m$ in parallel}
\State $z^{k+1} = z^{k+1/2} - C^{\text{serv}} \left [\frac{1}{M} \sum\limits_{m=1}^M C^{\text{dev}}_m(\gamma  F_{m, {\color{blue}\pi^{k}_m}}(z^{k+1/2}) - \gamma  F_{m, {\color{blue}\pi^{k}_m}}(w^k)+ e^k_m )+ e^k\right] $
\If {$b_k = 1$}
\State $w^{k+1} = z^{k}$ 
\State Compute $F_m(w^{k+1})$ and it send to server; and get $F(w^{k+1})$
\Else 
\State $w^{k+1} = w^k$ 
\EndIf
\EndFor
\EndFor
	\end{algorithmic}
\end{algorithm}

The following theorem gives the convergence of \algname{VR-MASHA2}. \label{sec:vr_masha2}

\begin{theorem} \label{th_comp_vr}
Let distributed variational inequality \eqref{VI} + \eqref{MK} + \eqref{MK_fs} is solved by Algorithm \ref{alg_qfbvr} with $\tau \geq \frac{3}{4}$ and biased compressor operators \eqref{compr}: on server with $\delta^{\text{serv}}$ parameter, on devices with $\delta^{\text{dev}}$. Let Assumption \ref{as3} and one case of Assumption \ref{as1} are satisfied. Then the following estimates holds 

$\bullet$ in strongly-monotone case with $\gamma \leq \min\left[ \frac{1-\tau}{8\mu}; \frac{\sqrt{1-\tau}}{ 2L + 165 \delta^{\text{serv}}\delta^{\text{dev}} \hat L}\right]$:
\begin{align*}
 \EE\left(\|\hat z^{K} - z^* \|^2 + \|w^{K} - z^* \|^2\right) &\leq \left(1-  \frac{\mu \gamma}{2}\right)^K \cdot 2\|z^{0} - z^* \|^2 ;
\end{align*}

$\bullet$ in monotone case with $\gamma \leq \frac{\sqrt{1-\tau}}{2L + 165 \delta^{\text{serv}}\delta^{\text{dev}} \hat L}$:
\begin{align*}
    \EE\left[\max_{z \in \mathcal{C}} \la F(z) , \left(\frac{1}{K}\sum\limits_{k=0}^{K-1} z^{k+1/2}\right) - z\ra \right] 
    &\leq \frac{2\max_{z \in \mathcal{C}}\| z^{0} - z\|^2 + 6\| z^{0} - z^*\|^2}{\gamma K};
\end{align*}

$\bullet$ in non-monotone case with $\gamma \leq \frac{\sqrt{1-\tau}}{2L +  165\delta^{\text{serv}}\delta^{\text{dev}} \hat L}$:
\begin{align*}
    \frac{1}{K}\sum\limits_{k=0}^{K-1}\EE\| F(w^k)\|^2
    &\leq \frac{32\EE\|z^{0} - z^*\|^2 }{\gamma^2 K}.
\end{align*}
\end{theorem}

We consider the only devices compression. For simplicity, we put $\tilde L = \hat L = L$. We use  the same reasoning as in Section \ref{sec:vr_masha1}. The optimal choice is $1 - \tau = \frac{1}{\max\{\beta, r \}}$.

\begin{corollary} \label{cor:vr_masha2} 
Let distributed variational inequality \eqref{VI} + \eqref{MK}+\eqref{MK_fs} is solved by Algorithm \ref{alg_qfbvr} without compression on server ($\delta^{\text{serv}} = 1$) and with  biased compressor operators \eqref{compr} on devices with $\delta^{\text{dev}} = \delta$. Let Assumption \ref{as3} and one case of Assumption \ref{as1} are satisfied. Then the following estimates holds 

$\bullet$ in strongly-monotone case with $\gamma \leq \min\left[ \frac{1}{8\mu}  \cdot \left( \max\{\beta, r \}\right)^{-1} ; \frac{1}{167 \delta L}  \cdot \left(\sqrt{ \max\{\beta, r \}}\right)^{-1}\right]$:
\begin{align*}
 \EE\left(\|\hat z^{K} - z^* \|^2 + \|w^{K} - z^* \|^2\right) &\leq \left(1-  \frac{\mu \gamma}{2}\right)^K \cdot 2\|z^{0} - z^* \|^2 ;
\end{align*}

$\bullet$ in monotone case with $\gamma \leq \frac{1}{167 \delta  L}  \cdot \left(\sqrt{ \max\{\beta, r \}}\right)^{-1}$:
\begin{align*}
    \EE\left[\max_{z \in \mathcal{C}} \la F(z) , \left(\frac{1}{K}\sum\limits_{k=0}^{K-1} z^{k+1/2}\right) - z\ra \right] 
    &\leq \frac{2\max_{z \in \mathcal{C}}\| z^{0} - z\|^2 + 4\| z^{0} - z^*\|^2}{\gamma K};
\end{align*}

$\bullet$ in non-monotone case with $\gamma \leq \frac{1}{167\delta  L}  \cdot \left(\sqrt{ \max\{\beta, r \}}\right)^{-1}$:
\begin{align*}
    \EE\left(\frac{1}{K}\sum\limits_{k=0}^{K-1}\| F(w^k)\|^2\right)
    &\leq \frac{32\EE\|z^{0} - z^*\|^2 }{\gamma^2 K}.
\end{align*}
\end{corollary}
In the line 6 of  Table \ref{tab:comparison1} we put complexities to achieve $\varepsilon$-solution.

\subsubsection{Proof of the convergence of \algnamenormal{VR-MASHA2}} \label{prof_th3}

\textbf{Proof of Theorem \ref{th_comp_vr}:}
The proofs of Theorem \ref{th_comp_vr} partially repeat the proofs of Theorem \ref{th_comp}. We note the main changes in comparison with Theorem \ref{th_comp}. 

The first difference is an update of "hat" sequence \eqref{t88}: 
\begin{align*}
    \hat z^{k+1} &= z^{k+1} - e^{k+1} - \frac{1}{M} \sum\limits_{m=1}^M e^{k+1}_m \nonumber\\
    &= z^{k+1/2} - C^{\text{serv}} \left [\frac{1}{M} \sum\limits_{m=1}^M C^{\text{dev}}_m(\gamma  F_{m, \pi^{k}_m}(z^{k+1/2}) - \gamma  F_{m, \pi^{k}_m}(w^k)+ e^k_m )+ e^k\right] \nonumber\\
    &\hspace{0.4cm}-e^k - \frac{1}{M} \sum\limits_{m=1}^M C^{\text{dev}}_m(\gamma  F_{m, \pi^{k}_m}(z^{k+1/2}) - \gamma  F_{m, \pi^{k}_m}(w^k)+ e^k_m )
    \nonumber\\
    &\hspace{0.4cm}+ C^{\text{serv}} \left [\frac{1}{M} \sum\limits_{m=1}^M C^{\text{dev}}_m(\gamma  F_{m, \pi^{k}_m}(z^{k+1/2}) - \gamma  F_{m, \pi^{k}_m}(w^k)+ e^k_m )+ e^k\right] 
    \nonumber\\
    &\hspace{0.4cm}- \frac{1}{M} \sum\limits_{m=1}^M \left[e^k_m + \gamma F_{m, \pi^{k}_m}(z^{k+1/2}) - \gamma F_{m, \pi^{k}_m}(w^k) - C^{\text{dev}}_m(\gamma \cdot F_{m, \pi^{k}_m}(z^{k+1/2}) - \gamma  F_{m, \pi^{k}_m}(w^k) + e^k_m)\right] \nonumber\\
    &= z^{k+1/2} - e^k - \frac{1}{M} \sum_{m=1}^M e^k_m - \gamma \cdot \frac{1}{M} \sum\limits_{m=1}^M  (F_{m, \pi^{k}_m}(z^{k+1/2}) - \gamma  F_{m, \pi^{k}_m}(w^k)) \nonumber \\
    &= \hat z^{k+1/2} - \gamma \cdot \frac{1}{M} \sum\limits_{m=1}^M  (F_{m, \pi^{k}_m}(z^{k+1/2}) - \gamma  F_{m, \pi^{k}_m}(w^k)).
\end{align*}
Hence, we need to modify \eqref{t66} 
\begin{align*}
    \|\hat z^{k+1} - z \|^2
    &\leq
    \|\hat z^{k} - z \|^2 + 2 \la \hat z^{k+1} - \hat z^{k}, z^{k+1/2} - z\ra \nonumber\\
    &\hspace{0.4cm}+ 2  \gamma^2 \cdot \left\| \frac{1}{M} \sum\limits_{m=1}^M  (F_{m, \pi^{k}_m}(z^{k+1/2}) - F_{m, \pi^{k}_m}(w^k))\right\|^2 + 4 \| e^k\|^2 +  \frac{4}{M} \sum\limits_{m=1}^M \left\| e^{k}_m \right\|^2 \nonumber\\
    &\hspace{0.4cm}- \| z^{k+1/2} - \hat z^{k}\|^2;
\end{align*}
and \eqref{t686}:
\begin{align*}
\hat z^{k+1}-\hat z^k &= \hat z^{k+1} - \hat z^{k+1/2} + \hat z^{k+1/2} - \hat z^k \\
&= - \gamma \cdot \left(\frac{1}{M} \sum\limits_{m=1}^M  (F_{m, \pi^{k}_m}(z^{k+1/2}) - F_{m, \pi^{k}_m}(w^k))\right) + z^{k+1/2} - z^k \\
&= - \gamma \cdot \left(\frac{1}{M} \sum\limits_{m=1}^M  (F_{m, \pi^{k}_m}(z^{k+1/2}) - F_{m, \pi^{k}_m}(w^k))\right) - \gamma \cdot F(w^k)  + \bar z^{k} -  z^k,
\end{align*}
Then \eqref{t22} is also modified:
\begin{align}
    \label{t2021}
    \|\hat z^{k+1} - z\|^2
    &\leq \|\hat z^{k} - z\|^2 - (1 - \tau)\|z^{k} - z \|^2 + (1 - \tau)\| w^k  - z\|^2  \nonumber\\
    &\hspace{0.4cm}-2 \gamma \la \left(\frac{1}{M} \sum\limits_{m=1}^M  (F_{m, \pi^{k}_m}(z^{k+1/2}) - F_{m, \pi^{k}_m}(w^k))\right) + F(w^k) ,z^{k+1/2} - z\ra \nonumber\\
    &\hspace{0.4cm} -  (1-\tau)\| w^k - z^{k+1/2} \|^2 +  2  \gamma^2 \cdot \left\| \frac{1}{M} \sum\limits_{m=1}^M  (F_{m, \pi^{k}_m}(z^{k+1/2}) - F_{m, \pi^{k}_m}(w^k))\right\|^2\nonumber\\
    &\hspace{0.4cm} +  6 \| e^k\|^2 +  \frac{6}{M} \sum\limits_{m=1}^M \left\| e^{k}_m \right\|^2 
    - \left(\tau - \frac{1}{2}\right)\| z^{k+1/2} - z^{k}\|^2.
\end{align}
Next, we move to different cases of monotonicity.

\textbf{Strongly-monotone}

The same way as in Theorem \ref{th_comp} we put $z=z^*$, use property of the solution and then take full expectation:
\begin{align}
\label{t609}
    \EE\|\hat z^{k+1} - z^*\|^2
    &\leq \EE\|\hat z^{k} - z^*\|^2 - (1 - \tau)\EE\|z^{k} - z^* \|^2 + (1 - \tau)\EE\| w^k  - z^*\|^2  \nonumber\\
    &\hspace{0.4cm}-2 \gamma \EE \left[\la \left(\frac{1}{M} \sum\limits_{m=1}^M  (F_{m, \pi^{k}_m}(z^{k+1/2}) - F_{m, \pi^{k}_m}(w^k))\right) + F(w^k) - F(z^*),z^{k+1/2} - z^*\ra \right]\nonumber\\
    &\hspace{0.4cm} -  (1-\tau)\EE\| w^k - z^{k+1/2} \|^2 +  2  \gamma^2 \cdot \EE\left\| \frac{1}{M} \sum\limits_{m=1}^M  (F_{m, \pi^{k}_m}(z^{k+1/2}) - F_{m, \pi^{k}_m}(w^k))\right\|^2\nonumber\\
    &\hspace{0.4cm} +  6 \EE\| e^k\|^2 +  \frac{6}{M} \sum\limits_{m=1}^M \EE\left\| e^{k}_m \right\|^2 
    - \left(\tau - \frac{1}{2}\right)\EE\| z^{k+1/2} - z^{k}\|^2 \nonumber\\
    &\leq \EE\|\hat z^{k} - z^*\|^2 - (1 - \tau)\EE\|z^{k} - z^* \|^2 + (1 - \tau)\EE\| w^k  - z^*\|^2  \nonumber\\
    &\hspace{0.4cm}-2 \gamma \EE \left[\la \EE_{\pi^{k}}\left[\frac{1}{M} \sum\limits_{m=1}^M  (F_{m, \pi^{k}_m}(z^{k+1/2}) - F_{m, \pi^{k}_m}(w^k)) + F(w^k) - F(z^*)\right],z^{k+1/2} - z^*\ra \right]\nonumber\\
    &\hspace{0.4cm} -  (1-\tau)\EE\| w^k - z^{k+1/2} \|^2 +  2  \gamma^2 \cdot \frac{1}{M} \sum\limits_{m=1}^M \EE \left[\EE_{\pi^{k}}\left\|   F_{m, \pi^{k}_m}(z^{k+1/2}) - F_{m, \pi^{k}_m}(w^k)\right\|^2\right]\nonumber\\
    &\hspace{0.4cm} +  6 \EE\| e^k\|^2 +  \frac{6}{M} \sum\limits_{m=1}^M \EE\left\| e^{k}_m \right\|^2 
    - \left(\tau - \frac{1}{2}\right)\EE\| z^{k+1/2} - z^{k}\|^2 \nonumber\\
    &=\EE\|\hat z^{k} - z^*\|^2 - (1 - \tau)\EE\|z^{k} - z^* \|^2 + (1 - \tau)\EE\| w^k  - z^*\|^2  \nonumber\\
    &\hspace{0.4cm}-2 \gamma \EE \left[\la  F(z^{k+1/2} - F(z^*),z^{k+1/2} - z^*\ra \right]\nonumber\\
    &\hspace{0.4cm} -  (1-\tau)\EE\| w^k - z^{k+1/2} \|^2 +  2  \gamma^2 \cdot \frac{1}{M} \sum\limits_{m=1}^M \EE \left[\frac{1}{r} \sum\limits_{i=1}^r\left\|   F_{m, i}(z^{k+1/2}) - F_{m, i}(w^k)\right\|^2\right]\nonumber\\
    &\hspace{0.4cm} +  6 \EE\| e^k\|^2 +  \frac{6}{M} \sum\limits_{m=1}^M \EE\left\| e^{k}_m \right\|^2
    - \left(\tau - \frac{1}{2}\right)\EE\| z^{k+1/2} - z^{k}\|^2 \nonumber\\
    &=\EE\|\hat z^{k} - z^*\|^2 - (1 - \tau)\EE\|z^{k} - z^* \|^2 + (1 - \tau)\EE\| w^k  - z^*\|^2  \nonumber\\
    &\hspace{0.4cm}-2 \gamma \EE \left[\la  F(z^{k+1/2} - F(z^*),z^{k+1/2} - z^*\ra \right]\nonumber\\
    &\hspace{0.4cm} -  (1-\tau)\EE\| w^k - z^{k+1/2} \|^2 +  2  \gamma^2 \hat L^2 \EE\| w^k - z^{k+1/2} \|^2\nonumber\\
    &\hspace{0.4cm} +  6 \EE\| e^k\|^2 +  \frac{6}{M} \sum\limits_{m=1}^M \EE\left\| e^{k}_m \right\|^2
    - \left(\tau - \frac{1}{2}\right)\EE\| z^{k+1/2} - z^{k}\|^2.
\end{align}
In the last we use Assumption \ref{as3} and definition of $\hat L$ from this Assumption. The new inequality \eqref{t609} is absolutely similar to inequality \eqref{t610}  (only $L$ is changed to $\hat L$). Therefore, we can safely reach the analogue of expression \eqref{t1000}:
\begin{align}
\label{t2020}
 \sum\limits_{k=0}^{K-1} &p^k \EE\|\hat z^{k+1} - z^* \|^2 + \sum\limits_{k=0}^{K-1} p^k \EE\|w^{k+1} - z^* \|^2 \nonumber\\
 &\leq \sum\limits_{k=0}^{K-1} p^k \EE\|\hat z^{k} - z^* \|^2 + \sum\limits_{k=0}^{K-1} p^k \EE\|w^{k} - z^* \|^2 - 2 \gamma \mu \sum\limits_{k=0}^{K-1} p^k \EE\| z^{k+1/2} - z^*\|^2 \nonumber\\
    & \hspace{0.4cm}-  (1-\tau - 2\gamma^2 \hat L^2 ) \cdot \sum\limits_{k=0}^{K-1} p^k \EE\| w^k - z^{k+1/2} \|^2 - \left(\tau - \frac{1}{2}\right)\cdot \sum\limits_{k=0}^{K-1} p^k\EE\| z^{k+1/2} - z^{k}\|^2\nonumber\\
    &\hspace{0.4cm} +  6 \cdot \sum\limits_{k=0}^{K-1} p^k\EE\| e^k\|^2 +  6 \cdot \sum\limits_{k=0}^{K-1} p^k \frac{1}{M} \sum\limits_{m=1}^M \EE\left\| e^{k}_m \right\|^2.
\end{align}
The only difference in the estimates on "errors" $e^k$ and $e^k_m$ is in the constant $\tilde L$. It needs to be changed to $\hat L$. And we have analogue of \eqref{t507}:
\begin{align}
\label{t233}
 \sum\limits_{k=0}^{K-1} &p^k \left(\EE\|\hat z^{k+1} - z^* \|^2 + \EE\|w^{k+1} - z^* \|^2\right) \nonumber\\
    &\leq \sum\limits_{k=0}^{K-1} p^k \left(1 - \frac{\mu \gamma}{2}\right) \left(\EE\|\hat z^{k} - z^* \|^2 + \EE\|w^{k} - z^* \|^2\right) \nonumber\\
    & \hspace{0.4cm}-  (1-\tau - \gamma \mu - 13200(\delta^{\text{serv}})^2 (\delta^{\text{dev}})^2 \gamma^2 \hat L^2  ) \cdot \sum\limits_{k=0}^{K-1} p^k \EE\| w^k - z^{k+1/2} \|^2 \nonumber\\
    &\hspace{0.4cm}- \left(\tau - 2\gamma\mu - \frac{1}{2}\right)\cdot \sum\limits_{k=0}^{K-1} p^k\EE\| z^{k+1/2} - z^{k}\|^2 .
\end{align}
Choice $\tau \geq \frac{3}{4}$, $\gamma \leq \min\left[ \frac{1-\tau}{8\mu}; \frac{\sqrt{1-\tau}}{165 \delta^{\text{serv}}\delta^{\text{dev}} \hat L}\right]$ finishes the proof.
\EndProof

\textbf{Monotone case}

We start from \eqref{t2021} with small rearrangements:
\begin{align*}
    2 \gamma &\la F(z^{k+1/2}) ,z^{k+1/2} - z\ra \nonumber\\
    &\leq \|\hat z^{k} - z\|^2 - \|\hat z^{k+1} - z\|^2 +  \| w^k  - z\|^2 - \| w^{k+1}  - z\|^2 \nonumber\\
    &\hspace{0.4cm}+\| w^{k+1}  - z\|^2- (1 - \tau)\|z^{k} - z \|^2 - \tau\| w^k  - z\|^2  \nonumber\\
    &\hspace{0.4cm}-2 \gamma \la \left(\frac{1}{M} \sum\limits_{m=1}^M  (F_{m, \pi^{k}_m}(z^{k+1/2}) - F_{m, \pi^{k}_m}(w^k))\right) + F(w^k) - F(z^{k+1/2}),z^{k+1/2} - z\ra \\
    &\hspace{0.4cm} -  (1-\tau)\| w^k - z^{k+1/2} \|^2 +  2  \gamma^2 \cdot \left\| \frac{1}{M} \sum\limits_{m=1}^M  (F_{m, \pi^{k}_m}(z^{k+1/2}) - F_{m, \pi^{k}_m}(w^k))\right\|^2\nonumber\\
    &\hspace{0.4cm} +  6 \| e^k\|^2 +  \frac{6}{M} \sum\limits_{m=1}^M \left\| e^{k}_m \right\|^2 
    - \left(\tau - \frac{1}{2}\right)\| z^{k+1/2} - z^{k}\|^2.
\end{align*}
The same way as in Theorem \ref{th_comp} we use monotonicity (Assumption \ref{as1} (M)) and then sum from $0$ to $K-1$:
\begin{align*}
    2 \gamma &\sum\limits_{k=0}^{K-1} \la F(z) ,z^{k+1/2} - z\ra \\
    &\leq \|\hat z^{0} - z\|^2 + \| w^{0} - z\|^2 \\
    &\hspace{0.4cm}+ \sum\limits_{k=0}^{K-1} \left(\| w^{k+1}  - z\|^2 - (1 - \tau)\|z^{k} - z \|^2 - \tau\| w^k  - z\|^2\right)  \nonumber\\
    &\hspace{0.4cm}-2 \gamma \cdot \sum\limits_{k=0}^{K-1}\la \left(\frac{1}{M} \sum\limits_{m=1}^M  (F_{m, \pi^{k}_m}(z^{k+1/2}) - F_{m, \pi^{k}_m}(w^k))\right) + F(w^k) - F(z^{k+1/2}),z^{k+1/2} - z\ra \\
    &\hspace{0.4cm} -  (1-\tau)\cdot \sum\limits_{k=0}^{K-1}\| w^k - z^{k+1/2} \|^2 +  2  \gamma^2 \cdot \sum\limits_{k=0}^{K-1} \left\| \frac{1}{M} \sum\limits_{m=1}^M  (F_{m, \pi^{k}_m}(z^{k+1/2}) - F_{m, \pi^{k}_m}(w^k))\right\|^2\nonumber\\
    &\hspace{0.4cm} +  6 \cdot \sum\limits_{k=0}^{K-1} \| e^k\|^2 +  6 \cdot \sum\limits_{k=0}^K  \frac{1}{M} \sum\limits_{m=1}^M \left\| e^{k}_m \right\|^2 - \left(\tau - \frac{1}{2}\right) \cdot \sum\limits_{k=0}^{K-1} \| z^{k+1/2} - z^{k}\|^2.
\end{align*}
Then we take maximum of both sides over $z \in \mathcal{C}$, after take expectation and get
\begin{align*}
    2 \gamma &\EE\left[\max_{z \in \mathcal{C}}\sum\limits_{k=0}^{K-1} \la F(z) ,z^{k+1/2} - z\ra \right] \nonumber\\
    &\leq \EE\left[\max_{z \in \mathcal{C}}\|\hat z^{0} - z\|^2 \right] + \EE\left[\max_{z \in \mathcal{C}}\|w^{0} - z\|^2 \right]  \\
    &\hspace{0.4cm}+ \EE\left[\max_{z \in \mathcal{C}}\sum\limits_{k=0}^{K-1} \left(\| w^{k+1}  - z\|^2 - (1 - \tau)\|z^{k} - z \|^2 - \tau\| w^k  - z\|^2\right)\right]   \nonumber\\
    &\hspace{0.4cm}+2 \gamma \EE\left[\max_{z \in \mathcal{C}} \sum\limits_{k=0}^{K-1}-\la \left(\frac{1}{M} \sum\limits_{m=1}^M  (F_{m, \pi^{k}_m}(z^{k+1/2}) - F_{m, \pi^{k}_m}(w^k))\right) + F(w^k) - F(z^{k+1/2}),z^{k+1/2} - z\ra \right]\\
    &\hspace{0.4cm} -  (1-\tau)\cdot \sum\limits_{k=0}^{K-1}\EE\| w^k - z^{k+1/2} \|^2 +  2  \gamma^2 \cdot \sum\limits_{k=0}^{K-1} \EE\left\| \frac{1}{M} \sum\limits_{m=1}^M  (F_{m, \pi^{k}_m}(z^{k+1/2}) - F_{m, \pi^{k}_m}(w^k))\right\|^2\nonumber\\
    &\hspace{0.4cm} +  6 \cdot \sum\limits_{k=0}^{K-1} \EE\| e^k\|^2 +  6 \cdot \sum\limits_{k=0}^K  \frac{1}{M} \sum\limits_{m=1}^M \EE\left\| e^{k}_m \right\|^2 - \left(\tau - \frac{1}{2}\right) \cdot \sum\limits_{k=0}^{K-1} \EE\| z^{k+1/2} - z^{k}\|^2.
\end{align*}
The same way as in strongly-monotone case of this Theorem (Theorem \ref{th_comp_vr}) we estimate $2  \gamma^2\sum\limits_{k=0}^{K-1} \EE\left\| \frac{1}{M} \sum\limits_{m=1}^M  (F_{m, \pi^{k}_m}(z^{k+1/2}) - F_{m, \pi^{k}_m}(w^k))\right\|^2+  6 \sum\limits_{k=0}^{K-1} \EE\| e^k\|^2 +  6 \sum\limits_{k=0}^K  \frac{1}{M} \sum\limits_{m=1}^M \EE\left\| e^{k}_m \right\|^2$:
\begin{align*}
    2 \gamma &\EE\left[\max_{z \in \mathcal{C}}\sum\limits_{k=0}^{K-1} \la F(z) ,z^{k+1/2} - z\ra \right] \nonumber\\
    &\leq \EE\left[\max_{z \in \mathcal{C}}\|\hat z^{0} - z\|^2 \right] + \EE\left[\max_{z \in \mathcal{C}}\|w^{0} - z\|^2 \right]  \\
    &\hspace{0.4cm}+ \EE\left[\max_{z \in \mathcal{C}}\sum\limits_{k=0}^{K-1} \left(\| w^{k+1}  - z\|^2 - (1 - \tau)\|z^{k} - z \|^2 - \tau\| w^k  - z\|^2\right)\right]   \nonumber\\
    &\hspace{0.4cm}+2 \gamma \EE\left[\max_{z \in \mathcal{C}} \sum\limits_{k=0}^{K-1}-\la \left(\frac{1}{M} \sum\limits_{m=1}^M  (F_{m, \pi^{k}_m}(z^{k+1/2}) - F_{m, \pi^{k}_m}(w^k))\right) + F(w^k) - F(z^{k+1/2}),z^{k+1/2} - z\ra \right]\\
    &\hspace{0.4cm} -  (1-\tau - 13200(\delta^{\text{serv}})^2 (\delta^{\text{dev}})^2 \gamma^2 \hat L^2  ) \cdot \sum\limits_{k=0}^{K-1} \EE\| w^k - z^{k+1/2} \|^2 \nonumber\\
    &\hspace{0.4cm}- \left(\tau - \frac{1}{2}\right)\cdot \sum\limits_{k=0}^{K-1}\EE\| z^{k+1/2} - z^{k}\|^2 .
\end{align*}
With $t \geq \frac{3}{4}$ and $\gamma \leq \frac{\sqrt{1-\tau}}{165 \delta^{\text{serv}}\delta^{\text{dev}} \hat L}$ we get
\begin{align*}
    2 \gamma &\EE\left[\max_{z \in \mathcal{C}}\sum\limits_{k=0}^{K-1} \la F(z) ,z^{k+1/2} - z\ra \right] \nonumber\\
    &\leq \EE\left[\max_{z \in \mathcal{C}}\|\hat z^{0} - z\|^2 \right] + \EE\left[\max_{z \in \mathcal{C}}\|w^{0} - z\|^2 \right]  \\
    &\hspace{0.4cm}+ \EE\left[\max_{z \in \mathcal{C}}\sum\limits_{k=0}^{K-1} \left(\| w^{k+1}  - z\|^2 - (1 - \tau)\|z^{k} - z \|^2 - \tau\| w^k  - z\|^2\right)\right]   \nonumber\\
    &\hspace{0.4cm}+2 \gamma \EE\left[\max_{z \in \mathcal{C}} \sum\limits_{k=0}^{K-1}-\la \left(\frac{1}{M} \sum\limits_{m=1}^M  (F_{m, \pi^{k}_m}(z^{k+1/2}) - F_{m, \pi^{k}_m}(w^k))\right) + F(w^k) - F(z^{k+1/2}),z^{k+1/2} - z\ra \right] .
\end{align*}
Using \eqref{t405}, we obtain
\begin{align}
\label{t449}
    2 \gamma &\EE\left[\max_{z \in \mathcal{C}}\sum\limits_{k=0}^{K-1} \la F(z) ,z^{k+1/2} - z\ra \right] \nonumber\\
    &\leq \EE\left[\max_{z \in \mathcal{C}}\|\hat z^{0} - z\|^2 \right] + \EE\left[\max_{z \in \mathcal{C}}\|w^{0} - z\|^2 \right] + \EE\left[\max_{z \in \mathcal{C}} \|z^0 - z\|^2\right] \nonumber\\
    &\hspace{0.4cm}+  (1 - \tau)\cdot\sum\limits_{k=0}^{K-1} \EE\left[\| z^{k} - w^k \|^2 \right] \\
    &\hspace{0.4cm}+2 \gamma \EE\left[\max_{z \in \mathcal{C}} \sum\limits_{k=0}^{K-1}-\la \left(\frac{1}{M} \sum\limits_{m=1}^M  (F_{m, \pi^{k}_m}(z^{k+1/2}) - F_{m, \pi^{k}_m}(w^k))\right) + F(w^k) - F(z^{k+1/2}),z^{k+1/2} - z\ra \right] \nonumber.
\end{align}
Let us work with the last line. For this define sequence $v$: $v^0 = z^{0}$, $v^{k+1} = 
v^k-\gamma \delta_k$ with $\delta^k = F(z^{k+1/2}) - \left(\frac{1}{M} \sum\limits_{m=1}^M  (F_{m, \pi^{k}_m}(z^{k+1/2}) - F_{m, \pi^{k}_m}(w^k))\right) + F(w^k)$. Then we have
\begin{align}
    \label{t44413}
    \sum\limits_{k=0}^{K-1} \langle \delta^k, z^{k+1/2} - z \rangle = \sum\limits_{k=0}^{K-1} \langle \delta^k, z^{k+1/2} - v^k \rangle + 
    \sum\limits_{k=0}^{K-1} \langle \delta^k,  v^k - z \rangle . 
\end{align}
By the definition of $v^{k+1}$  we get
\begin{align*}
    \langle \gamma\delta^k, v^k  - z \rangle &= \langle \gamma\delta^k, v^k - v^{k+1} \rangle  + \langle v^{k+1} - v^k, z - v^{k+1} \rangle \nonumber\\
&= \langle \gamma\delta^k, v^k - v^{k+1} \rangle + \frac{1}{2}\|v^k - z\|^2 -  \frac{1}{2}\|v^{k+1} - z\|^2 - \frac{1}{2}\| v^k - v^{k+1}\|^2
\nonumber\\
&= \frac{\gamma^2}{2} \|\delta^k\|^2  + \frac{1}{2}\|v^k - v^{k+1}\|^2 + \frac{1}{2}\|v^k - z\|^2 -  \frac{1}{2}\|v^{k+1} - z\|^2 - \frac{1}{2}\| v^k - v^{k+1}\|^2
\nonumber\\
&= \frac{\gamma^2}{2} \|\delta^k\|^2  + \frac{1}{2}\|v^k - z\|^2 -  \frac{1}{2}\|v^{k+1} - z\|^2 .
\end{align*}
With \eqref{t44413} it gives
\begin{align*}
    \sum\limits_{k=0}^{K-1} \langle \delta^k, z^{k+1/2} - z \rangle &\leq \sum\limits_{k=0}^{K-1} \langle \delta^k, z^{k+1/2} - v^k \rangle + 
    \frac{1}{\gamma}\sum\limits_{k=0}^{K-1} \left(\frac{\gamma^2}{2} \|\delta^k\|^2  + \frac{1}{2}\|v^k - z\|^2 -  \frac{1}{2}\|v^{k+1} - z\|^2 \right) \nonumber\\
    &\leq \sum\limits_{k=0}^{K-1} \langle \delta^k, z^{k+1/2} - v^k \rangle + 
    \frac{\gamma}{2}\sum\limits_{k=0}^{K-1} \|\delta^k\|^2 + \frac{1}{2\gamma}\|v^0 - z\|^2. 
\end{align*}

We take the maximum on  $z$ and get
\begin{align*}
     \max_{z \in \mathcal{C}} \sum\limits_{k=0}^{K-1} \langle \delta^k, z^{k+1/2} - z \rangle &\leq \sum\limits_{k=0}^{K-1} \langle \delta^k, z^{k+1/2} - v^k \rangle  + \frac{1}{2\gamma} \max_{z \in \mathcal{C}}  \|z^0 - z\|^2 \\
     &\hspace{0.4cm}+ 
    \frac{\gamma}{2}\sum\limits_{k=0}^{K-1} \|F(z^{k+1/2}) - \left(\frac{1}{M} \sum\limits_{m=1}^M  (F_{m, \pi^{k}_m}(z^{k+1/2}) - F_{m, \pi^{k}_m}(w^k))\right) - F(w^k)\|^2.  
\end{align*}
Taking the full expectation, we get
\begin{align}
\label{temp4043}
     \EE&\left[ \max_{z \in \mathcal{C}} \sum\limits_{k=0}^{K-1} \langle \delta^k, z^{k+1/2} - z \rangle\right] \leq \EE\left[\sum\limits_{k=0}^{K-1} \langle \delta^k, z^{k+1/2} - v^k \rangle\right] \nonumber\\
    &\hspace{0.4cm}+ 
    \frac{\gamma}{2}\sum\limits_{k=0}^{K-1} \EE\left[\|F(z^{k+1/2}) - \left(\frac{1}{M} \sum\limits_{m=1}^M  (F_{m, \pi^{k}_m}(z^{k+1/2}) - F_{m, \pi^{k}_m}(w^k))\right) - F(w^k)\|^2 \right]\nonumber\\  
    &\hspace{0.4cm}+\frac{1}{2\gamma}\max_{z \in \mathcal{C}} \|v^0 - z\|^2 \nonumber\\
    &= \EE\left[\sum\limits_{k=0}^{K-1} \langle \EE_{\pi^k}\left[F(z^{k+1/2}) - \left(\frac{1}{M} \sum\limits_{m=1}^M  (F_{m, \pi^{k}_m}(z^{k+1/2}) - F_{m, \pi^{k}_m}(w^k))\right) - F(w^k) \right], z^{k+1/2} - v^k \rangle\right] \nonumber\\
    &\hspace{0.4cm}+ 
    \frac{\gamma}{2}\sum\limits_{k=0}^{K-1} \EE\left[\|F(z^{k+1/2}) - \left(\frac{1}{M} \sum\limits_{m=1}^M  (F_{m, \pi^{k}_m}(z^{k+1/2}) - F_{m, \pi^{k}_m}(w^k))\right) - F(w^k)\|^2 \right]
    \nonumber\\  
    &\hspace{0.4cm}+ \frac{1}{2\gamma}\max_{z \in \mathcal{C}} \|z^0 - z\|^2 \nonumber\\
    &= \frac{\gamma}{2}\sum\limits_{k=0}^{K-1} \EE\left[\|F(z^{k+1/2}) - \left(\frac{1}{M} \sum\limits_{m=1}^M  (F_{m, \pi^{k}_m}(z^{k+1/2}) - F_{m, \pi^{k}_m}(w^k))\right) - F(w^k)\|^2 \right]\nonumber\\  
    &\hspace{0.4cm}+ \frac{1}{2\gamma}\max_{z \in \mathcal{C}} \|z^0 - z\|^2 \nonumber\\
    &\leq \gamma\sum\limits_{k=0}^{K-1} \EE\left[\frac{1}{M} \sum\limits_{m=1}^M \|  F_{m, \pi^{k}_m}(z^{k+1/2}) - F_{m, \pi^{k}_m}(w^k)\|^2 \right] + \gamma\sum\limits_{k=0}^{K-1} \EE\left[\|F(z^{k+1/2}) - F(w^k)\|^2 \right]\nonumber\\  
    &\hspace{0.4cm}+ \frac{1}{2\gamma}\max_{z \in \mathcal{C}} \|z^0 - z\|^2 
    \nonumber\\
    &\leq \gamma (\hat L^2 + L^2)\sum\limits_{k=0}^{K-1} \EE\|z^{k+1/2}-w^k\|^2 + \frac{1}{2\gamma}\max_{z \in \mathcal{C}} \|z^0 - z\|^2 .
\end{align}
Substituting \eqref{temp4043} to \eqref{t449}, we get
\begin{align*}
    2 \gamma &\EE\left[\max_{z \in \mathcal{C}}\sum\limits_{k=0}^{K-1} \la F(z) ,z^{k+1/2} - z\ra \right] \nonumber\\
    &\leq \EE\left[\max_{z \in \mathcal{C}}\|\hat z^{0} - z\|^2 \right] + \EE\left[\max_{z \in \mathcal{C}}\|w^{0} - z\|^2 \right] + \EE\left[\max_{z \in \mathcal{C}} \|z^0 - z\|^2\right] \nonumber\\
    &\hspace{0.4cm}+  (1 - \tau)\cdot\sum\limits_{k=0}^{K-1} \EE\left[\| z^{k} - w^k \|^2 \right] \nonumber\\
    &\hspace{0.4cm}+2 \gamma^2(\hat L^2 + L^2)\sum\limits_{k=0}^{K-1} \EE\|z^{k+1/2}-w^k\|^2 + \max_{z \in \mathcal{C}} \|z^0 - z\|^2.
\end{align*}   
With $\gamma \leq \frac{\sqrt{1-\tau}}{2L + 2\hat L}$ we have
\begin{align}
    \label{t0010}
    2 \gamma &\EE\left[\max_{z \in \mathcal{C}}\sum\limits_{k=0}^{K-1} \la F(z) ,z^{k+1/2} - z\ra \right] \nonumber\\
    &\leq \EE\left[\max_{z \in \mathcal{C}}\|\hat z^{0} - z\|^2 \right] + \EE\left[\max_{z \in \mathcal{C}}\|w^{0} - z\|^2 \right] + 2\EE\left[\max_{z \in \mathcal{C}} \|z^0 - z\|^2\right] \nonumber\\
    &\hspace{0.4cm}+  3(1 - \tau)\cdot\sum\limits_{k=0}^{K-1} \EE\left[\| z^{k} - z^{k+1/2} \|^2 \right] + \EE\left[\| z^{k+1/2} - w^k \|^2 \right].
\end{align}
Taking into account \eqref{t233} with $p=1$ and $\mu=0$ (monotone case), we get
\begin{align*}
 \sum\limits_{k=0}^{K-1} & \left(\EE\|\hat z^{k+1} - z^* \|^2 + \EE\|w^{k+1} - z^* \|^2\right) \nonumber\\
    &\leq \sum\limits_{k=0}^{K-1} \left(\EE\|\hat z^{k} - z^* \|^2 + \EE\|w^{k} - z^* \|^2\right) \nonumber\\
    & \hspace{0.4cm}-  (1-\tau  - 13200(\delta^{\text{serv}})^2 (\delta^{\text{dev}})^2 \gamma^2 \hat L^2  ) \cdot \sum\limits_{k=0}^{K-1} \EE\| w^k - z^{k+1/2} \|^2 \nonumber\\
    &\hspace{0.4cm}- \left(\tau  - \frac{1}{2}\right)\cdot \sum\limits_{k=0}^{K-1}\EE\| z^{k+1/2} - z^{k}\|^2 .
\end{align*}
With $\gamma \leq \frac{\sqrt{1-\tau}}{2L + 165 \delta^{\text{serv}}\delta^{\text{dev}} \tilde L}$ we get
\begin{align*}
  \nonumber\\
    \frac{1-\tau}{2} \cdot \sum\limits_{k=0}^{K-1} \left(\EE\| w^k - z^{k+1/2} \|^2 + \EE\| z^{k+1/2} - z^{k}\|^2\right) &\leq \left(\EE\|\hat z^{0} - z^* \|^2 + \EE\|w^{0} - z^* \|^2\right).
\end{align*}
Combing this expression with \eqref{t0010}, we obtain
\begin{align*}
    2 \gamma &\EE\left[\max_{z \in \mathcal{C}}\sum\limits_{k=0}^{K-1} \la F(z) ,z^{k+1/2} - z\ra \right] \nonumber\\
    &\leq \EE\left[\max_{z \in \mathcal{C}}\|\hat z^{0} - z\|^2 \right] + \EE\left[\max_{z \in \mathcal{C}}\|w^{0} - z\|^2 \right] + 2\EE\left[\max_{z \in \mathcal{C}} \|z^0 - z\|^2\right] \nonumber\\
    &\hspace{0.4cm}+  6\left(\EE\|\hat z^{0} - z^* \|^2 + \EE\|w^{0} - z^* \|^2\right),
\end{align*}
and finish the proof in the monotone case.
\EndProof

\textbf{Non-monotone case}

We start from \eqref{t2021}, put $z = z^*$, use non-monotonicity assumption and then take a full mathematical expectation:
\begin{align*}
    \EE\|\hat z^{k+1} - z^*\|^2
    &\leq \EE\|\hat z^{k} - z^*\|^2 - (1 - \tau)\EE\|z^{k} - z^* \|^2 + (1 - \tau)\EE\| w^k  - z^*\|^2  \nonumber\\
    &\hspace{0.4cm}-2 \gamma \EE \left[\la \left(\frac{1}{M} \sum\limits_{m=1}^M  (F_{m, \pi^{k}_m}(z^{k+1/2}) - F_{m, \pi^{k}_m}(w^k))\right) + F(w^k),z^{k+1/2} - z^*\ra \right]\nonumber\\
    &\hspace{0.4cm} -  (1-\tau)\EE\| w^k - z^{k+1/2} \|^2 +  2  \gamma^2 \cdot \EE\left\| \frac{1}{M} \sum\limits_{m=1}^M  (F_{m, \pi^{k}_m}(z^{k+1/2}) - F_{m, \pi^{k}_m}(w^k))\right\|^2\nonumber\\
    &\hspace{0.4cm} +  6 \EE\| e^k\|^2 +  \frac{6}{M} \sum\limits_{m=1}^M \EE\left\| e^{k}_m \right\|^2 
    - \left(\tau - \frac{1}{2}\right)\EE\| z^{k+1/2} - z^{k}\|^2 \nonumber\\
    &\leq \EE\|\hat z^{k} - z^*\|^2 - (1 - \tau)\EE\|z^{k} - z^* \|^2 + (1 - \tau)\EE\| w^k  - z^*\|^2  \nonumber\\
    &\hspace{0.4cm}-2 \gamma \EE \left[\la \EE_{\pi^{k}}\left[\frac{1}{M} \sum\limits_{m=1}^M  (F_{m, \pi^{k}_m}(z^{k+1/2}) - F_{m, \pi^{k}_m}(w^k)) + F(w^k) \right],z^{k+1/2} - z^*\ra \right]\nonumber\\
    &\hspace{0.4cm} -  (1-\tau)\EE\| w^k - z^{k+1/2} \|^2 +  2  \gamma^2 \cdot \frac{1}{M} \sum\limits_{m=1}^M \EE \left[\EE_{\pi^{k}}\left\|   F_{m, \pi^{k}_m}(z^{k+1/2}) - F_{m, \pi^{k}_m}(w^k)\right\|^2\right]\nonumber\\
    &\hspace{0.4cm} +  6 \EE\| e^k\|^2 +  \frac{6}{M} \sum\limits_{m=1}^M \EE\left\| e^{k}_m \right\|^2 
    - \left(\tau - \frac{1}{2}\right)\EE\| z^{k+1/2} - z^{k}\|^2 \nonumber\\
    &=\EE\|\hat z^{k} - z^*\|^2 - (1 - \tau)\EE\|z^{k} - z^* \|^2 + (1 - \tau)\EE\| w^k  - z^*\|^2  \nonumber\\
    &\hspace{0.4cm}-2 \gamma \EE \left[\la  F(z^{k+1/2},z^{k+1/2} - z^*\ra \right]\nonumber\\
    &\hspace{0.4cm} -  (1-\tau)\EE\| w^k - z^{k+1/2} \|^2 +  2  \gamma^2 \cdot \frac{1}{M} \sum\limits_{m=1}^M \EE \left[\frac{1}{r} \sum\limits_{i=1}^r\left\|   F_{m, i}(z^{k+1/2}) - F_{m, i}(w^k)\right\|^2\right]\nonumber\\
    &\hspace{0.4cm} +  6 \EE\| e^k\|^2 +  \frac{6}{M} \sum\limits_{m=1}^M \EE\left\| e^{k}_m \right\|^2
    - \left(\tau - \frac{1}{2}\right)\EE\| z^{k+1/2} - z^{k}\|^2 \nonumber\\
    &=\EE\|\hat z^{k} - z^*\|^2 - (1 - \tau)\EE\|z^{k} - z^* \|^2 + (1 - \tau)\EE\| w^k  - z^*\|^2  \nonumber\\
    &\hspace{0.4cm} -  (1-\tau)\EE\| w^k - z^{k+1/2} \|^2 +  2  \gamma^2 \hat L^2 \EE\| w^k - z^{k+1/2} \|^2\nonumber\\
    &\hspace{0.4cm} +  6 \EE\| e^k\|^2 +  \frac{6}{M} \sum\limits_{m=1}^M \EE\left\| e^{k}_m \right\|^2
    - \left(\tau - \frac{1}{2}\right)\EE\| z^{k+1/2} - z^{k}\|^2.
\end{align*}
This expression is the same with \eqref{t6123}. Then we repeat all steps from Theorem \ref{th_comp}. And then with $\gamma \leq \frac{\sqrt{1-\tau}}{2L + 165\delta^{\text{serv}}\delta^{\text{dev}} \hat L}$ we get
\begin{align*}
    \frac{1}{K}\sum\limits_{k=0}^{K-1}\EE\| F(w^k)\|^2
    &\leq \frac{16(\EE\|\hat z^{0} - z^*\|^2 + \EE\|w^{0} - z^*\|^2)}{\gamma^2 K}.
\end{align*}

\newpage

\section{Federated learning and partial participation} \label{sec:fed}

Here we consider a popular federated learning feature - partial participation. We model it as follows. At each iteration, only $b$ random devices send information to the server. The rest do not compute and do not communicate. More formally, at each iteration we
\begin{equation}
    \label{pp_setup}
  \text{generate subset } \{\xi^{k}_i\}_{i=1}^b \text{ of } \{1, \ldots, M\}
\end{equation}
devices, which takes part in the current iteration.
Next, we show how to modify \algname{MASHA1} and \algname{MASHA2} for partial participation. 



\subsection{\algnamenormal{PP-MASHA1}: federated learning version} \label{sec:pp_masha1}

In this section, we provide information about \algname{PP-MASHA1}. This is a modification of \algname{MASHA1} for the federated learning case. Changes compared to  \algname{MASHA1} are highlighted in blue -- see Algorithm \ref{alg_qeg_pp}.

\begin{algorithm}[th]
	\caption{\algname{PP-MASHA1}}
	\label{alg_qeg_pp}
	\begin{algorithmic}[1]
\State
\noindent {\bf Parameters:}  Stepsize $\gamma>0$, parameters $\tau$ and $b$, number of iterations $K$.\\
\noindent {\bf Initialization:} Choose  $z^0 = w^0 \in \mathcal{Z}$. \\
Server  sends to devices $z^0 = w^0$ and devices compute $F_m(w^0)$ and send to server and get $F(w^0)$
\For {$k=0,1, 2, \ldots, K-1$ }
\State {\color{blue} Generate subset $\{\xi^{k}_i\}_{i=1}^b$ of $\{1, \ldots, M\}$ independently}
\For {{\color{blue} each device $m$ from $\{\xi^{k}_i\}_{i=1}^b$ in parallel}}
\State $\bar z^k = \tau z^k + (1 - \tau) w^k$
\State $z^{k+1/2} = \bar z_k - \gamma F(w^k)$,
\State  Compute $F_m(z^{k+1/2})$ \& send  $Q^{\text{dev}}_m(F_m(z^{k+1/2}) - F_m(w^k))$ to server
\EndFor
\For {server}
\State Compute $Q^{\text{serv}} \left [\frac{1}{{\color{blue}b}} {\color{blue}\sum\limits_{i=1}^b} Q^{\text{dev}}_{{\color{blue}\xi^{k}_i}}(  F_{{\color{blue}\xi^{k}_i}}(z^{k+1/2}) -  F_{{\color{blue}\xi^{k}_i}}(w^k))\right] $  \& send to devices 
\State Sends to devices one bit $b_k$: 1 with probability $1 - \tau$, 0 with with probability $\tau$
\EndFor
\For {each device $m$ in parallel}
\State $z^{k+1} = z^{k+1/2} - \gamma  Q^{\text{serv}} \left [\frac{1}{{\color{blue}b}} {\color{blue}\sum\limits_{i=1}^b} Q^{\text{dev}}_{{\color{blue}\xi^{k}_i}}(  F_{{\color{blue}\xi^{k}_i}}(z^{k+1/2}) -  F_{{\color{blue}\xi^{k}_i}}(w^k))\right] $
\If {$b_k = 1$}
\State $w^{k+1} = z^{k+1}$
\State Compute $F_m(w^{k+1})$ \& send it to server; and get $F(w^{k+1})$ as a response from server
\Else 
\State $w^{k+1} = w^k$
\EndIf
\EndFor
\EndFor
	\end{algorithmic}
\end{algorithm}

The following theorem gives the convergence of \algname{PP-MASHA1}. 

\begin{theorem} \label{th_q3}
Let distributed variational inequality \eqref{VI} + \eqref{MK} + \eqref{pp_setup} is solved by Algorithm \ref{alg_qeg_pp} with unbiased compressor operators \eqref{quant}: on server with $q^{\text{serv}}$ parameter, on devices with $\{q^{\text{dev}}_m\}$. Let Assumption \ref{as3} and one case of Assumption \ref{as1} are satisfied. Then the following estimates holds 

$\bullet$ in strongly-monotone case with $\gamma \leq \min\left[\frac{\sqrt{1 - \tau}}{2  C^b_q}; \frac{1 - \tau}{2\mu}\right]$ (where $C^b_q = \sqrt{\frac{q^{\text{serv}}}{bM}  \cdot \sum_{m=1}^M (q_m^{\text{dev}} \tilde L_m^2 + (b-1) \tilde L^2)}$):
\begin{align*}
 \EE\left(\| z^{K} - z^* \|^2 + \|w^{K} - z^* \|^2\right) &\leq \left( 1 -\frac{\mu\gamma}{2} \right)^K \cdot 2\|z^{0} - z^* \|^2;
\end{align*}

$\bullet$ in monotone case with $\gamma \leq \frac{\sqrt{1 - \tau}}{2  C^b_q + 4 \tilde L}$:
\begin{align*}
\EE\left[\max_{z \in \mathcal{C}} \left[ \langle F(u),  \left(\frac{1}{K}\sum\limits_{k=0}^{K-1} z^{k+1/2}\right) - u  \rangle \right] \right]
\leq  \frac{2\max_{z \in \mathcal{C}}\left[\| z^0 - z \|^2\right] + 6\|z^{0} - z^* \|^2}{\gamma K};
\end{align*}

$\bullet$ in non-monotone case with $\gamma \leq \frac{\sqrt{1 - \tau}}{2  C^b_q}$:
\begin{align*}
    \EE\left(\frac{1}{K}\sum\limits_{k=0}^{K-1}\| F(w^k)\|^2\right)
    &\leq \frac{16\|z^{0} - z^* \|^2}{\gamma^2 K}.
\end{align*}
\end{theorem}

For \algname{PP-MASHA1} we consider the case of only devices compression. For simplicity, we put $Q^{\text{dev}}_m = Q$ with $q^{\text{dev}}_m = q$ and $\beta^{\text{dev}}_m = \beta$, also $\tilde L_m = \tilde L = L$.
Let us discuss the difference (with \algname{MASHA1}) in choosing $\tau$. When $b_k = 1$, all devices send uncompressed information to the server, but only $b$ devices send compressed information (line 9 \algname{PP-MASHA1}). Then, at each iteration, we send $\mathcal{O}\left( \frac{b}{\beta} + M( 1-\tau)\right)$ bits of information. Therefore, the optimal choice of $\tau$ is $1 - \tau = \frac{b}{\beta M}$.

\begin{corollary} \label{cor:pp_masha1}
Let distributed variational inequality \eqref{VI} + \eqref{MK} + \eqref{pp_setup} is solved by Algorithm \ref{alg_qeg_pp} without compression on server ($q^{\text{serv}} = 1$) and with unbiased compressor operators \eqref{quant} on devices with $\{q^{\text{dev}}_m = q\}$. Let Assumption \ref{as3} and one case of Assumption \ref{as1} are satisfied. Then the following estimates holds 

$\bullet$ in strongly-monotone case with $\gamma \leq \min\left[\frac{1}{2 L} \cdot \left(\sqrt{\frac{q\beta }{b} + \frac{\beta M}{b} } \right)^{-1} ; \frac{b}{2\mu \beta M}\right]$:
\begin{align*}
 \EE\left(\| z^{K} - z^* \|^2 + \|w^{K} - z^* \|^2\right) &\leq \left( 1 -\frac{\mu\gamma}{2} \right)^K \cdot 2\|z^{0} - z^* \|^2;
\end{align*}

$\bullet$ in monotone case with $\gamma \leq \frac{1}{6L} \cdot \left(\sqrt{\frac{q\beta }{b} + \frac{\beta M}{b} } \right)^{-1}$:
\begin{align*}
\EE\left[\max_{z \in \mathcal{C}} \left[ \langle F(u),  \left(\frac{1}{K}\sum\limits_{k=0}^{K-1} z^{k+1/2}\right) - u  \rangle \right] \right]
\leq  \frac{2\max_{z \in \mathcal{C}}\left[\| z^0 - z \|^2\right] + 6\|z^{0} - z^* \|^2}{\gamma K};
\end{align*}

$\bullet$ in non-monotone case with $\gamma \leq \frac{1}{2 L} \cdot \left(\sqrt{\frac{q\beta }{b} + \frac{\beta M}{b} } \right)^{-1}$:
\begin{align*}
    \EE\left(\frac{1}{K}\sum\limits_{k=0}^{K-1}\| F(w^k)\|^2\right)
    &\leq \frac{16\|z^{0} - z^* \|^2}{\gamma^2 K}.
\end{align*}
\end{corollary}
In the line 7 of  Table \ref{tab:comparison1} we put complexities to achieve $\varepsilon$-solution.

\subsubsection{Proof of the convergence of \algnamenormal{PP-MASHA1}} \label{prof_th6}

\textbf{Proof of Theorem \ref{th_q3}:}

The proof is very close to the proof of Theorem \ref{th_q}. Only two estimates need to be modified. First is \eqref{temp33}
\begin{align*}
\EE\left[\| z^{k+1} -  z^{k+1/2}\|^2\right] &= \gamma^2 \cdot \EE\left[\left\| Q^{\text{serv}} \left [\frac{1}{b} \sum_{i=1}^b Q^{\text{dev}}_{\xi^{k}_i}(F_{\xi^{k}_i}(z^{k+1/2}) - F_{\xi^{k}_i}(w^k))\right]\right\|^2\right]  \\
&\leq \gamma^2 \cdot\frac{q^{\text{serv}}}{b^2} \EE\left[ \left\| \sum_{m=1}^M Q^{\text{dev}}_{\xi^{k}_i}(F_{\xi^{k}_i}(z^{k+1/2}) - F_{\xi^{k}_i}(w^k))\right\|^2\right]  \\
&= \gamma^2 \cdot \frac{q^{\text{serv}}}{b^2} \sum_{i=1}^b \EE\left[\left \|  Q^{\text{dev}}_{\xi^{k}_i}(F_{\xi^{k}_i}(z^{k+1/2}) - F_{\xi^{k}_i}(w^k))\right\|^2\right] \\
&\hspace{0.4cm}+ \gamma^2 \cdot \frac{q^{\text{serv}}}{b^2} \sum_{i \neq j}\EE \left[\langle Q^{\text{dev}}_{\xi^{k}_i} (F_{\xi^{k}_i}(z^{k+1/2}) - F_{\xi^{k}_i}(w^k)) ; Q^{\text{dev}}_{\xi^{k}_j} (F_{\xi^{k}_j}(z^{k+1/2}) - F_{\xi^{k}_j}(w^k)) \rangle \right]
\end{align*}
Next we apply \eqref{quant} and Assumption \ref{as3} for the first term and independence and unbiasedness of $Q$ and uniformess of $\xi$ for the second term:
\begin{align*}
\EE\left[\| z^{k+1} -  z^{k+1/2}\|^2\right] &\leq \gamma^2 \cdot\frac{q^{\text{serv}}}{b^2} \sum_{i=1}^b \EE\left[ \EE_{\xi^k}\left[q_{\xi^{k}_i}^{\text{dev}} L_{\xi^{k}_i}^2\right] \left \|  z^{k+1/2} - w^k\right\|^2\right] \nonumber\\
&\hspace{0.4cm}+ \gamma^2 \cdot\frac{q^{\text{serv}}}{b^2} \sum_{i \neq j}\EE \left[\langle F_{\xi^{k}_i}(z^{k+1/2}) - F_{\xi^{k}_i}(w^k) ; F_{\xi^{k}_j}(z^{k+1/2}) - F_{\xi^{k}_j}(w^k) \rangle \right] \nonumber\\
&\leq \gamma^2 \cdot\frac{q^{\text{serv}}}{bM} \sum_{m=1}^M q_m^{\text{dev}} \tilde L_m^2 \EE\left[\left \|  z^{k+1/2} - w^k\right\|^2\right] \nonumber\\
&\hspace{0.4cm}+ \gamma^2 \cdot\frac{q^{\text{serv}}}{2b^2} \sum_{i \neq j}\EE \left[\| F_{\xi^{k}_i}(z^{k+1/2}) - F_{\xi^{k}_i}(w^k) \|^2 +  \| F_{\xi^{k}_j}(z^{k+1/2}) - F_{\xi^{k}_j}(w^k) \|^2 \right] \nonumber\\
&\leq \gamma^2 \cdot\frac{q^{\text{serv}}}{bM} \sum_{m=1}^M q_m^{\text{dev}} \tilde L_m^2 \EE\left[\left \|  z^{k+1/2} - w^k\right\|^2\right] \nonumber\\
&\hspace{0.4cm}+ \gamma^2 \cdot\frac{q^{\text{serv}}}{2b^2} \sum_{i \neq j}  \EE \left[L^2_{\xi^{k}_i} \| z^{k+1/2} - w^k\|^2 +  L_{\xi^{k}_j}^2\| z^{k+1/2} - w^k \|^2 \right] \nonumber\\
&= \gamma^2 \cdot\frac{q^{\text{serv}}}{bM} \sum_{m=1}^M q_m^{\text{dev}} \tilde L_m^2 \EE\left[\left \|  z^{k+1/2} - w^k\right\|^2\right] \nonumber\\
&\hspace{0.4cm}
+ \gamma^2 \cdot\frac{q^{\text{serv}} (b-1)}{b} \tilde L^2 \EE \left[\| z^{k+1/2} - w^k\|^2\right]
\nonumber\\
&= \gamma^2 \cdot\frac{q^{\text{serv}}}{M^2} \EE \left[\| z^{k+1/2} - w^k\|^2\right] \cdot \sum_{m=1}^M q_m^{\text{dev}} \tilde L_m^2 + (M-1) \tilde L^2
\end{align*}
Here we can use new $ C^b_q = \sqrt{\frac{q^{\text{serv}}}{bM}  \cdot \sum_{m=1}^M (q_m^{\text{dev}} \tilde L_m^2 + (b-1) \tilde L^2)}$.

The second modified estimate is \eqref{temp35}:
\begin{align*}
\EE&\left[\left\|Q^{\text{serv}} \left [\frac{1}{M} \sum_{m=1}^M Q^{\text{dev}}_m(F_m(z^{k+1/2}) - F_m(w^k))\right] + F(w^k) - F(z^{k+1/2})\right\|^2 \right]
\nonumber\\
&\leq (C^b_q)^2 \EE\left[\left\|z^{k+1/2} - w^k \right\|^2 \right] + 2\tilde L^2 \EE\left[\left\|  z^{k+1/2} - w^k\right\|^2 \right].
\end{align*}
\EndProof

\newpage

\subsection{\algnamenormal{PP-MASHA2}: federated learning version} \label{sec:pp_masha2}

In this section, we provide information about \algname{PP-MASHA2} from Section \ref{sec:fed}. This is a modification of \algname{MASHA2} for the federated learning case. Changes compared to  \algname{MASHA2} are highlighted in blue -- see Algorithm \ref{alg_qfb_pp}.

\begin{algorithm}[th]
	\caption{\algname{PP-MASHA2} }
	\label{alg_qfb_pp}
	\begin{algorithmic}[1]
\State
\noindent {\bf Parameters:}  Stepsize $\gamma>0$, parameters $\tau$ and {\color{blue}$b$}, number of iterations $K$.\\
\noindent {\bf Initialization:} Choose  $z^0 = w^0 \in \mathcal{Z}$, $e^0_m = 0$, $e^0 = 0$. \\
Server  sends to devices $z^0 = w^0$ and devices compute $F_m(w^0)$ and send to server and get $F(w^0)$
\For {$k=0,1, 2, \ldots, K-1$ }
\State {\color{blue} Generate subset $\{\xi^{k}_i\}_{i=1}^b$ of $\{1, \ldots, M\}$ independently}
\For {{\color{blue} each device $m$ from $\{\xi^{k}_i\}_{i=1}^b$ in parallel}}
\State $\bar z^k = \tau z^k + (1 - \tau) w^k$ 
\State $z^{k+1/2} = \bar z_k - \gamma F(w^k)$ 
\State  Compute $F_m(z^{k+1/2})$ and send to server $C^{\text{dev}}_m(\gamma F_m(z^{k+1/2}) - \gamma F_m(w^k) + e^k_m)$ 
\State $e^{k+1}_m = e^k_m + \gamma F_m(z^{k+1/2}) - \gamma F_m(w^k) - C^{\text{dev}}_m(\gamma F_m(z^{k+1/2}) - \gamma F_m(w^k) + e^k_m)$
\EndFor
\For {{\color{blue}devices not from $\{\xi^{k}_i\}_{i=1}^b$ in parallel}}
\State {\color{blue}$e^{k+1}_m = e^k_m$}
\EndFor
\For {server}
\State Compute $g^k = C^{\text{serv}} \left [\frac{1}{{\color{blue}b}} {\color{blue}\sum\limits_{i=1}^b} C^{\text{dev}}_{{\color{blue}\xi^{k}_i}}(\gamma  F_{{\color{blue}\xi^{k}_i}}(z^{k+1/2}) - \gamma  F_{{\color{blue}\xi^{k}_i}}(w^k)+ e^k_{{\color{blue}\xi^{k}_i}} ) + e^k\right] $  \& send to devices 
\State $e^{k+1} = e^k + \frac{1}{{\color{blue}b}} {\color{blue}\sum\limits_{i=1}^b} C^{\text{dev}}_{{\color{blue}\xi^{k}_i}}(\gamma  F_{{\color{blue}\xi^{k}_i}}(z^{k+1/2}) - \gamma  F_{{\color{blue}\xi^{k}_i}}(w^k)+ e^k_{{\color{blue}\xi^{k}_i}} ) - g^k$.
\State Sends to devices one bit $b_k$: 1 with probability $1 - \tau$, 0 with with probability $\tau$ 
\EndFor
\For {each device $m$ in parallel}
\State $z^{k+1} = z^{k+1/2} - C^{\text{serv}} \left [\frac{1}{{\color{blue}b}} {\color{blue}\sum\limits_{i=1}^b} C^{\text{dev}}_{{\color{blue}\xi^{k}_i}}(\gamma  F_{{\color{blue}\xi^{k}_i}}(z^{k+1/2}) - \gamma  F_{{\color{blue}\xi^{k}_i}}(w^k)+ e^k_{{\color{blue}\xi^{k}_i}} ) + e^k\right] $
\If {$b_k = 1$}
\State $w^{k+1} = z^{k}$ 
\State Compute $F_m(w^{k+1})$ and it send to server; and get $F(w^{k+1})$
\Else 
\State $w^{k+1} = w^k$ 
\EndIf
\EndFor
\EndFor
	\end{algorithmic}
\end{algorithm}

The following theorem gives the convergence of \algname{PP-MASHA2}. 

\begin{theorem} \label{th_comp_pp}
Let distributed variational inequality \eqref{VI} + \eqref{MK} is solved by Algorithm \ref{alg_qfb_pp} with $\tau \geq \frac{3}{4}$ and biased compressor operators \eqref{compr}: on server with $\delta^{\text{serv}}$ parameter, on devices with $\delta^{\text{dev}}$. Let Assumption \ref{as3} and one case of Assumption \ref{as1} are satisfied. Then the following estimates holds 

$\bullet$ in strongly-monotone case with $\gamma \leq \min\left[ \frac{1-\tau}{8\mu}; \frac{\sqrt{1-\tau}}{(30\delta^{\text{serv}} + 10 \delta^{\text{dev}}\frac{M}{b} + 165 \delta^{\text{dev}}\delta^{\text{serv}} \sqrt{\frac{M}{b}})\tilde L}\right]$:
\begin{align*}
 \EE\left(\|\hat z^{K} - z^* \|^2 + \|w^{K} - z^* \|^2\right) &\leq \left(1-  \frac{\mu \gamma}{2}\right)^K \cdot 2\|z^{0} - z^* \|^2 ;
\end{align*}

$\bullet$ in monotone case with $\gamma \leq \frac{\sqrt{1-\tau}}{(30\delta^{\text{serv}} + 10 \delta^{\text{dev}}\frac{M}{b} + 165 \delta^{\text{dev}}\delta^{\text{serv}} \sqrt{\frac{M}{b}})\tilde L}$:
\begin{align*}
    \EE\left[\max_{z \in \mathcal{C}} \la F(z) , \left(\frac{1}{K}\sum\limits_{k=0}^{K-1} z^{k+1/2}\right) - z\ra \right] 
    &\leq \frac{2\max_{z \in \mathcal{C}}\| z^{0} - z\|^2 + 6\| z^{0} - z^*\|^2}{\gamma K};
\end{align*}

$\bullet$ in non-monotone case with $\gamma \leq \frac{\sqrt{1-\tau}}{(30\delta^{\text{serv}} + 10 \delta^{\text{dev}}\frac{M}{b} + 165 \delta^{\text{dev}}\delta^{\text{serv}} \sqrt{\frac{M}{b}})\tilde L}$:
\begin{align*}
    \frac{1}{K}\sum\limits_{k=0}^{K-1}\EE\| F(w^k)\|^2
    &\leq \frac{32\EE\|z^{0} - z^*\|^2 }{\gamma^2 K}.
\end{align*}
\end{theorem}

We consider the only devices compression. For simplicity, we put $\tilde L = \hat L = L$. We use  the same reasoning as in Section \ref{sec:pp_masha1}. The optimal choice is $1 - \tau = \frac{b}{\beta M}$.

\begin{corollary} \label{cor:pp_masha2} 
Let distributed variational inequality \eqref{VI} + \eqref{MK} + \eqref{pp_setup} is solved by Algorithm \ref{alg_qfb_pp} without compression on server ($\delta^{\text{serv}} = 1$) and with  biased compressor operators \eqref{compr} on devices with $\delta^{\text{dev}} = \delta$. Let Assumption \ref{as3} and one case of Assumption \ref{as1} are satisfied. Then the following estimates holds 

$\bullet$ in strongly-monotone case with $\gamma \leq \min\left[ \frac{1}{8\mu \beta} ; \frac{\sqrt{b^3}}{205 \delta  \sqrt{\beta M^3} L}\right]$:
\begin{align*}
 \EE\left(\|\hat z^{K} - z^* \|^2 + \|w^{K} - z^* \|^2\right) &\leq \left(1-  \frac{\mu \gamma}{2}\right)^K \cdot 2\|z^{0} - z^* \|^2 ;
\end{align*}

$\bullet$ in monotone case with $\gamma \leq \frac{\sqrt{b^3}}{205 \delta  \sqrt{\beta M^3} L} $:
\begin{align*}
    \EE\left[\max_{z \in \mathcal{C}} \la F(z) , \left(\frac{1}{K}\sum\limits_{k=0}^{K-1} z^{k+1/2}\right) - z\ra \right] 
    &\leq \frac{2\max_{z \in \mathcal{C}}\| z^{0} - z\|^2 + 4\| z^{0} - z^*\|^2}{\gamma K};
\end{align*}

$\bullet$ in non-monotone case with $\gamma \leq \frac{\sqrt{b^3}}{205 \delta  \sqrt{\beta M^3} L} $:
\begin{align*}
    \EE\left(\frac{1}{K}\sum\limits_{k=0}^{K-1}\| F(w^k)\|^2\right)
    &\leq \frac{32\EE\|z^{0} - z^*\|^2 }{\gamma^2 K}.
\end{align*}
\end{corollary}
In the line 8 of  Table \ref{tab:comparison1} we put complexities to achieve $\varepsilon$-solution.

\subsubsection{Proof of the convergence of \algnamenormal{PP-MASHA2}} \label{prof_th4}

\textbf{Proof of Theorem \ref{th_comp_pp}:}
The proofs of Theorem \ref{th_comp_pp} partially repeat the proofs of Theorem \ref{th_comp}. We note the main changes in comparison with Theorem \ref{th_comp}. 

The first difference is definition of "hat" sequences:
$$
\hat z^{k} = z^k - e^k - \frac{1}{b} \sum\limits_{m=1}^M e^{k}_m, \quad \hat z^{k+1/2} = z^{k+1/2} - e^k - \frac{1}{b} \sum\limits_{m=1}^M e^{k}_m, \quad \hat w^{k} = w^k - e^k - \frac{1}{b} \sum\limits_{m=1}^M e^{k}_m.
$$
Then we modify an update of "hat" sequence \eqref{t88}: 
\begin{align*}
    \hat z^{k+1} &= z^{k+1} - e^{k+1} - \frac{1}{b} \sum\limits_{m=1}^M e^{k+1}_m \nonumber\\
    &= z^{k+1/2} - C^{\text{serv}} \left [\frac{1}{b} \sum\limits_{i=1}^b C^{\text{dev}}_{\xi^{k}_i}(\gamma  F_{\xi^{k}_i}(z^{k+1/2}) - \gamma  F_{\xi^{k}_i}(w^k)+ e^k_{\xi^{k}_i} ) + e^k\right]  \nonumber\\
    &\hspace{0.4cm}-e^k - \frac{1}{b} \sum\limits_{i=1}^b C^{\text{dev}}_{\xi^{k}_i}(\gamma  F_{\xi^{k}_i}(z^{k+1/2}) - \gamma  F_{\xi^{k}_i}(w^k)+ e^k_{\xi^{k}_i} )
    \nonumber\\
    &\hspace{0.4cm}+ C^{\text{serv}} \left [\frac{1}{b} \sum\limits_{i=1}^b C^{\text{dev}}_{\xi^{k}_i}(\gamma  F_{\xi^{k}_i}(z^{k+1/2}) - \gamma  F_{\xi^{k}_i}(w^k)+ e^k_{\xi^{k}_i} ) + e^k\right]  
    \nonumber\\
    &\hspace{0.4cm}- \frac{1}{b} \sum\limits_{i=1}^b \left[e^k_{\xi^{k}_i} + \gamma F_{\xi^{k}_i}(z^{k+1/2}) - \gamma  F_{\xi^{k}_i}(w^k) - C^{\text{dev}}_{\xi^{k}_i}(\gamma  F_{\xi^{k}_i}(z^{k+1/2}) - \gamma  F_{\xi^{k}_i}(w^k)+ e^k_{\xi^{k}_i} )\right] \nonumber\\
    &\hspace{0.4cm} - \frac{1}{b} \sum\limits_{j \notin \{\xi^{k}_i\}_{i=1}^b} e_j^k \nonumber \\
    &= z^{k+1/2} - e^k - \frac{1}{b} \sum_{m=1}^M e^k_m - \gamma \cdot \frac{1}{b} \sum\limits_{i=1}^b  (F_{\xi^{k}_i}(z^{k+1/2}) -  F_{\xi^{k}_i}(w^k)) \nonumber \\
    &= \hat z^{k+1/2} - \gamma \cdot \frac{1}{b} \sum\limits_{i=1}^b  (F_{\xi^{k}_i}(z^{k+1/2}) -   F_{\xi^{k}_i}(w^k)).
\end{align*}
Hence, we need to modify \eqref{t66} 
\begin{align*}
    \|\hat z^{k+1} - z \|^2
    &\leq
    \|\hat z^{k} - z \|^2 + 2 \la \hat z^{k+1} - \hat z^{k}, z^{k+1/2} - z\ra \nonumber\\
    &\hspace{0.4cm}+ 2  \gamma^2 \cdot \left\| \frac{1}{b} \sum\limits_{i=1}^b  (F_{\xi^{k}_i}(z^{k+1/2}) -   F_{\xi^{k}_i}(w^k))\right\|^2 + 4 \| e^k\|^2 +  \frac{4M}{b^2} \sum\limits_{m=1}^M \left\| e^{k}_m \right\|^2 \nonumber\\
    &\hspace{0.4cm}- \| z^{k+1/2} - \hat z^{k}\|^2;
\end{align*}
and \eqref{t686}:
\begin{align*}
\hat z^{k+1}-\hat z^k &= \hat z^{k+1} - \hat z^{k+1/2} + \hat z^{k+1/2} - \hat z^k \\
&= - \gamma \cdot \left(\frac{1}{b} \sum\limits_{i=1}^b  (F_{\xi^{k}_i}(z^{k+1/2}) -   F_{\xi^{k}_i}(w^k))\right) + z^{k+1/2} - z^k \\
&= - \gamma \cdot \left(\frac{1}{b} \sum\limits_{i=1}^b  (F_{\xi^{k}_i}(z^{k+1/2}) -  F_{\xi^{k}_i}(w^k))\right) - \gamma \cdot F(w^k)  + \bar z^{k} -  z^k,
\end{align*}
Then \eqref{t22} is also modified:
\begin{align}
    \label{t20214}
    \|\hat z^{k+1} - z\|^2
    &\leq \|\hat z^{k} - z\|^2 - (1 - \tau)\|z^{k} - z \|^2 + (1 - \tau)\| w^k  - z\|^2  \nonumber\\
    &\hspace{0.4cm}-2 \gamma \la \left(\frac{1}{b} \sum\limits_{i=1}^b  (F_{\xi^{k}_i}(z^{k+1/2}) - F_{\xi^{k}_i}(w^k))\right) + F(w^k) ,z^{k+1/2} - z\ra \nonumber\\
    &\hspace{0.4cm} -  (1-\tau)\| w^k - z^{k+1/2} \|^2 +  2  \gamma^2 \cdot \left\| \frac{1}{b} \sum\limits_{i=1}^b  (F_{\xi^{k}_i}(z^{k+1/2}) - F_{\xi^{k}_i}(w^k))\right\|^2\nonumber\\
    &\hspace{0.4cm} +  6 \| e^k\|^2 +  \frac{6M}{b^2} \sum\limits_{m=1}^M \left\| e^{k}_m \right\|^2 
    - \left(\tau - \frac{1}{2}\right)\| z^{k+1/2} - z^{k}\|^2.
\end{align}
Next, we move to different cases of monotonicity.

\textbf{Strongly-monotone}

The same way as in Theorem \ref{th_comp} we put $z=z^*$, use property of the solution and then take full expectation:
\begin{align}
\label{t6094}
    \EE\|\hat z^{k+1} - z^*\|^2
    &\leq \EE\|\hat z^{k} - z^*\|^2 - (1 - \tau)\EE\|z^{k} - z^* \|^2 + (1 - \tau)\EE\| w^k  - z^*\|^2  \nonumber\\
    &\hspace{0.4cm}-2 \gamma \EE \left[\la \left(\frac{1}{b} \sum\limits_{i=1}^b  (F_{\xi^{k}_i}(z^{k+1/2}) - F_{\xi^{k}_i}(w^k))\right) + F(w^k) - F(z^*),z^{k+1/2} - z^*\ra \right]\nonumber\\
    &\hspace{0.4cm} -  (1-\tau)\EE\| w^k - z^{k+1/2} \|^2 +  2  \gamma^2 \cdot \EE\left\| \frac{1}{b} \sum\limits_{i=1}^b  (F_{\xi^{k}_i}(z^{k+1/2}) - F_{\xi^{k}_i}(w^k))\right\|^2\nonumber\\
    &\hspace{0.4cm} +  6 \EE\| e^k\|^2 +  \frac{6M}{b^2} \sum\limits_{m=1}^M \EE\left\| e^{k}_m \right\|^2 
    - \left(\tau - \frac{1}{2}\right)\EE\| z^{k+1/2} - z^{k}\|^2 \nonumber\\
    &\leq \EE\|\hat z^{k} - z^*\|^2 - (1 - \tau)\EE\|z^{k} - z^* \|^2 + (1 - \tau)\EE\| w^k  - z^*\|^2  \nonumber\\
    &\hspace{0.4cm}-2 \gamma \EE \left[\la \EE_{\xi^{k}}\left[\frac{1}{b} \sum\limits_{i=1}^b  (F_{\xi^{k}_i}(z^{k+1/2}) - F_{\xi^{k}_i}(w^k)) + F(w^k) - F(z^*)\right],z^{k+1/2} - z^*\ra \right]\nonumber\\
    &\hspace{0.4cm} -  (1-\tau)\EE\| w^k - z^{k+1/2} \|^2 +  2  \gamma^2 \cdot \frac{1}{b} \sum\limits_{i=1}^b \EE \left[\EE_{\xi^{k}}L^2_{\xi^{k}_i}\left\|   z^{k+1/2} -  w^k\right\|^2\right]\nonumber\\
    &\hspace{0.4cm} +  6 \EE\| e^k\|^2 +  \frac{6M}{b^2} \sum\limits_{m=1}^M \EE\left\| e^{k}_m \right\|^2 
    - \left(\tau - \frac{1}{2}\right)\EE\| z^{k+1/2} - z^{k}\|^2 \nonumber\\
    &=\EE\|\hat z^{k} - z^*\|^2 - (1 - \tau)\EE\|z^{k} - z^* \|^2 + (1 - \tau)\EE\| w^k  - z^*\|^2  \nonumber\\
    &\hspace{0.4cm}-2 \gamma \EE \left[\la  F(z^{k+1/2} - F(z^*),z^{k+1/2} - z^*\ra \right]\nonumber\\
    &\hspace{0.4cm} -  (1-\tau)\EE\| w^k - z^{k+1/2} \|^2 +  2  \gamma^2 \cdot \frac{1}{M} \sum\limits_{m=1}^M L_m^2 \EE \left\|   z^{k+1/2} -  w^k\right\|^2\nonumber\\
    &\hspace{0.4cm} +  6 \EE\| e^k\|^2 +  \frac{6M}{b^2} \sum\limits_{m=1}^M \EE\left\| e^{k}_m \right\|^2 
    - \left(\tau - \frac{1}{2}\right)\EE\| z^{k+1/2} - z^{k}\|^2 \nonumber\\
    &=\EE\|\hat z^{k} - z^*\|^2 - (1 - \tau)\EE\|z^{k} - z^* \|^2 + (1 - \tau)\EE\| w^k  - z^*\|^2  \nonumber\\
    &\hspace{0.4cm}-2 \gamma \EE \left[\la  F(z^{k+1/2} - F(z^*),z^{k+1/2} - z^*\ra \right]\nonumber\\
    &\hspace{0.4cm} -  (1-\tau)\EE\| w^k - z^{k+1/2} \|^2 +  2  \gamma^2 \tilde L^2 \EE\| w^k - z^{k+1/2} \|^2\nonumber\\
    &\hspace{0.4cm} +  6 \EE\| e^k\|^2 +  \frac{6M}{b^2} \sum\limits_{m=1}^M \EE\left\| e^{k}_m \right\|^2
    - \left(\tau - \frac{1}{2}\right)\EE\| z^{k+1/2} - z^{k}\|^2.
\end{align}
In the last we use Assumption \ref{as3} and definition of $\tilde L$ from this Assumption. The new inequality \eqref{t6094} is absolutely similar to inequality \eqref{t610}  (only $L$ is changed to $\tilde L$ and a coefficient near $\sum\limits_{m=1}^M \EE\left\| e^{k}_m \right\|^2$). Therefore, we can safely reach the analogue of expression \eqref{t1000}:
\begin{align}
\label{t20204}
 \sum\limits_{k=0}^{K-1} &p^k \EE\|\hat z^{k+1} - z^* \|^2 + \sum\limits_{k=0}^{K-1} p^k \EE\|w^{k+1} - z^* \|^2 \nonumber\\
 &\leq \sum\limits_{k=0}^{K-1} p^k \EE\|\hat z^{k} - z^* \|^2 + \sum\limits_{k=0}^{K-1} p^k \EE\|w^{k} - z^* \|^2 - 2 \gamma \mu \sum\limits_{k=0}^{K-1} p^k \EE\| z^{k+1/2} - z^*\|^2 \nonumber\\
    & \hspace{0.4cm}-  (1-\tau - 2\gamma^2 \tilde L^2 ) \cdot \sum\limits_{k=0}^{K-1} p^k \EE\| w^k - z^{k+1/2} \|^2 - \left(\tau - \frac{1}{2}\right)\cdot \sum\limits_{k=0}^{K-1} p^k\EE\| z^{k+1/2} - z^{k}\|^2\nonumber\\
    &\hspace{0.4cm} +  6 \cdot \sum\limits_{k=0}^{K-1} p^k\EE\| e^k\|^2 +  6 \cdot \sum\limits_{k=0}^{K-1} p^k \frac{M}{b^2} \sum\limits_{m=1}^M \EE\left\| e^{k}_m \right\|^2.
\end{align}
Next, we need modify estimates on "error" terms:
\begin{align*}
 \EE\| e^{k+1}\|^2 &= 
 \EE\Bigg\|e^k + \frac{1}{b} \sum\limits_{i=1}^b C^{\text{dev}}_{\xi^{k}_i}(\gamma  F_{\xi^{k}_i}(z^{k+1/2}) - \gamma  F_{\xi^{k}_i}(w^k)+ e^k_{\xi^{k}_i} ) \\
 &\hspace{0.4cm}- C^{\text{serv}} \left [\frac{1}{b} \sum\limits_{i=1}^b C^{\text{dev}}_{\xi^{k}_i}(\gamma  F_{\xi^{k}_i}(z^{k+1/2}) - \gamma  F_{\xi^{k}_i}(w^k)+ e^k_{\xi^{k}_i} ) + e^k\right] \Bigg\|^2
 \\
&\leq  
\left(1-\frac{1}{\delta^{\text{serv}}}\right) \EE\left\|e^k + \frac{1}{b} \sum\limits_{i=1}^b C^{\text{dev}}_{\xi^{k}_i}(\gamma  F_{\xi^{k}_i}(z^{k+1/2}) - \gamma  F_{\xi^{k}_i}(w^k)+ e^k_{\xi^{k}_i} )\right\|^2
\\
&\leq  (1+c)\left(1-\frac{1}{\delta^{\text{serv}}}\right) \EE\left\|e^k\right\|^2 \\
&\hspace{0.4cm}+ \left(1 + \frac{1}{c}\right)\left(1-\frac{1}{\delta^{\text{serv}}}\right) \frac{1}{b} \sum\limits_{i=1}^b\EE\left\| C^{\text{dev}}_{\xi^{k}_i}(\gamma  F_{\xi^{k}_i}(z^{k+1/2}) - \gamma  F_{\xi^{k}_i}(w^k)+ e^k_{\xi^{k}_i} )\right\|^2.
\end{align*}
Here we use definition of biased compression \eqref{compr}, \eqref{eq:sqs} and inequality $\|a + b\|^2 \leq (1+ c) \| a\|^2 + (1 + 1/c) \| b\|^2$ (for $c > 0$). Is is easy to prove that for baised compressor $C^{\text{dev}}_m$ from \eqref{compr} it holds that $\|C^{\text{dev}}_m(x) \|^2 \leq 4 \|x\|^2$ (see \cite{beznosikov2020biased}). Then
\begin{align*}
\EE\| e^{k+1}\|^2 
&\leq  (1+c)\left(1-\frac{1}{\delta^{\text{serv}}}\right) \EE\left\|e^k\right\|^2 \\
&\hspace{0.4cm}+ \left(1 + \frac{1}{c}\right)\left(1-\frac{1}{\delta^{\text{serv}}}\right) \frac{4}{b} \sum\limits_{i=1}^b\EE\left\| \gamma  F_{\xi^{k}_i}(z^{k+1/2}) - \gamma  F_{\xi^{k}_i}(w^k)+ e^k_{\xi^{k}_i} \right\|^2 \\
&\leq  (1+c)\left(1-\frac{1}{\delta^{\text{serv}}}\right) \EE\left\|e^k\right\|^2 \\
&\hspace{0.4cm}+ \gamma^2\left(1 + \frac{1}{c}\right)\left(1-\frac{1}{\delta^{\text{serv}}}\right) \frac{8}{b} \sum\limits_{i=1}^b\EE\left\|  F_{\xi^{k}_i}(z^{k+1/2}) -  F_{\xi^{k}_i}(w^k)\right\|^2
\\
&\hspace{0.4cm}+ \left(1 + \frac{1}{c}\right)\left(1-\frac{1}{\delta^{\text{serv}}}\right) \frac{8}{b} \sum\limits_{i=1}^b \EE\left\|e^k_{\xi^{k}_i} \right\|^2
\\
&\leq  (1+c)\left(1-\frac{1}{\delta^{\text{serv}}}\right) \EE\left\|e^k\right\|^2 + 8\gamma^2 \tilde L^2\left(1 + \frac{1}{c}\right)\left(1-\frac{1}{\delta^{\text{serv}}}\right) \EE\left\|  z^{k+1/2} - w^k\right\|^2
\\
&\hspace{0.4cm}+ \left(1 + \frac{1}{c}\right)\left(1-\frac{1}{\delta^{\text{serv}}}\right) \frac{8}{b} \sum\limits_{m=1}^M\EE\left\|e^k_m \right\|^2.
\end{align*}
In the last we use Assumption \ref{as3} and definition of $\tilde L$ from this Assumption. With $c = \frac{1}{2(\delta - 1)}$ we get 
\begin{align*}
 \EE\| e^{k+1}\|^2
&\leq \left(1-\frac{1}{2\delta^{\text{serv}}}\right) \EE \left\|e^k\right\|^2 + 16\delta^{\text{serv}}\gamma^2 \tilde L^2\cdot \EE\|  z^{k+1/2} - w^k\|^2
+ 16\delta^{\text{serv}} \cdot \frac{1}{b} \sum\limits_{m=1}^M \EE\left\|e^k_m \right\|^2 \\
&\leq  16\delta^{\text{serv}}\gamma^2 \tilde L^2 \sum\limits_{j=0}^k \left(1-\frac{1}{2\delta^{\text{serv}}}\right)^{k-j}  \cdot \left\|  z^{j+1/2} - w^j  \right\|^2 \\ 
&\hspace{0.4cm}+ 16\delta^{\text{serv}}\sum\limits_{j=0}^k \left(1-\frac{1}{2\delta^{\text{serv}}}\right)^{k-j}  \cdot \frac{1}{b} \sum\limits_{m=1}^M\left\|e^j_m \right\|^2.
\end{align*}
The same way we can get analogue of \eqref{t11}:
\begin{align}
\label{t114}
 \sum\limits_{k=0}^{K-1} p^k \EE\left\| e^{k} \right\|^2
&\leq  128(\delta^{\text{serv}})^2\gamma^2 \tilde L^2 \sum\limits_{k=0}^{K-1}  p^{k} \EE\left\|  z^{k+1/2} - w^k  \right\|^2  \nonumber\\
&\hspace{0.4cm}+ 128(\delta^{\text{serv}})^2 \sum\limits_{k=0}^{K-1} p^k   \frac{1}{b} \sum\limits_{m=1}^M \EE\left\|e^k_m \right\|^2.
\end{align}
Combining \eqref{t20204} with \eqref{t114}, we obtain
\begin{align}
\label{t697}
 \sum\limits_{k=0}^{K-1} &p^k \EE\|\hat z^{k+1} - z^* \|^2 + \sum\limits_{k=0}^{K-1} p^k \EE\|w^{k+1} - z^* \|^2 \nonumber\\
 &\leq \sum\limits_{k=0}^{K-1} p^k \EE\|\hat z^{k} - z^* \|^2 + \sum\limits_{k=0}^{K-1} p^k \EE\|w^{k} - z^* \|^2 - 2 \gamma \mu \sum\limits_{k=0}^{K-1} p^k \EE\| z^{k+1/2} - z^*\|^2 \nonumber\\
    & \hspace{0.4cm}-  (1-\tau - 2\gamma^2 \tilde L^2 ) \cdot \sum\limits_{k=0}^{K-1} p^k \EE\| w^k - z^{k+1/2} \|^2 - \left(\tau - \frac{1}{2}\right)\cdot \sum\limits_{k=0}^{K-1} p^k\EE\| z^{k+1/2} - z^{k}\|^2\nonumber\\
    &\hspace{0.4cm} +  768(\delta^{\text{serv}})^2\gamma^2 \tilde L^2 \sum\limits_{k=0}^{K-1}  p^{k} \EE\left\|  z^{k+1/2} - w^k  \right\|^2  \nonumber\\
&\hspace{0.4cm}+ \left(768(\delta^{\text{serv}})^2 + \frac{6M}{b} \right) \sum\limits_{k=0}^{K-1} p^k   \frac{1}{b} \sum\limits_{m=1}^M \EE\left\|e^k_m \right\|^2 .
\end{align}
For the other "error" term. Let us note that $\|e^{k+1}_m \|^2 = \| e^{k}_m\|$ with probability $1 - \frac{b}{M}$ and $\|e^{k+1}_m \|^2 = \| e^k_{m} + \gamma F_{m}(z^{k+1/2}) - \gamma  F_{m}(w^k) - C^{\text{dev}}_{m}(\gamma  F_{m}(z^{k+1/2}) - \gamma  F_{m}(w^k)+ e^k_{m} )\|$ with probability $\frac{b}{M}$, then
\begin{align*}
 \frac{1}{b} \sum\limits_{m=1}^M \EE\left\| e_m^{k+1} \right\|^2 &= 
 \frac{1}{b} \sum\limits_{m=1}^M \frac{b}{M} \EE\left\|  e^k_{m} + \gamma F_{m}(z^{k+1/2}) - \gamma  F_{m}(w^k) - C^{\text{dev}}_{m}(\gamma  F_{m}(z^{k+1/2}) - \gamma  F_{m}(w^k)+ e^k_{m} )\right\|^2 \\
 &\hspace{0.4cm}+\frac{1}{b} \sum\limits_{m=1}^M\left(1- \frac{b}{M}\right) \EE\left\|  e^k_{m}\right\|^2 \\
&\leq  
\frac{1}{b} \sum\limits_{m=1}^M \frac{b}{M} \left(1-\frac{1}{\delta^{\text{dev}}}\right)\left\|  e^k_m + \gamma \cdot F_m(z^{k+1/2}) - \gamma \cdot F_m(w^k)  \right\|^2 +\left(1- \frac{b}{M}\right) \EE\left\|  e^k_{m}\right\|^2\\
&\leq  \frac{1}{b} \sum\limits_{m=1}^M \frac{b}{M}(1 + c)\left(1-\frac{1}{\delta^{\text{dev}}}\right)\left\|  e^k_m \right\|^2 + \frac{b}{M}\left(1 + \frac{1}{c}\right)\left(1-\frac{1}{\delta^{\text{dev}}}\right)\gamma^2 \cdot \left\|  F_m(z^{k+1/2}) - F_m(w^k)  \right\|^2 \\
 &\hspace{0.4cm}+\frac{1}{b} \sum\limits_{m=1}^M\left(1- \frac{b}{M}\right) \EE\left\|  e^k_{m}\right\|^2.
\end{align*}
With $c = \frac{1}{2(\delta^{\text{dev}} - 1)}$
\begin{align*}
\frac{1}{b} \sum\limits_{m=1}^M \left\| e_m^{k+1} \right\|^2
&\leq  \frac{1}{b} \sum\limits_{m=1}^M \frac{b}{M} \left(1-\frac{1}{2\delta^{\text{dev}}}\right)\left\|  e^k_m \right\|^2 + \frac{2 b\delta^{\text{dev}} \gamma^2}{M} \cdot \left\|  F_m(z^{k+1/2}) - F_m(w^k)  \right\|^2 \\
 &\hspace{0.4cm}+\frac{1}{b} \sum\limits_{m=1}^M\left(1- \frac{b}{M}\right) \EE\left\|  e^k_{m}\right\|^2 \\
&\leq \left(1-\frac{b}{2\delta^{\text{dev}}M}\right) \cdot \frac{1}{b} \sum\limits_{m=1}^M \left\|  e^k_m \right\|^2 + 2  \delta^{\text{dev}} \gamma^2 \tilde L^2 \cdot \left\|  z^{k+1/2} - w^k  \right\|^2 \\
&\leq  2 \delta^{\text{dev}} \gamma^2 \tilde L^2 \sum\limits_{j=0}^k \left(1-\frac{b}{2\delta^{\text{dev}}M}\right)^{k-j}  \cdot \left\|  z^{j+1/2} - w^j  \right\|^2.
\end{align*}
And then 
\begin{align}
\label{t1114}
 \sum\limits_{k=0}^{K-1} p^k \frac{1}{b} \sum\limits_{m=1}^M \left\| e_m^{k} \right\|^2
&\leq 16(\delta^{\text{dev}})^2 \gamma^2 \tilde L^2 \frac{M}{b} \sum\limits_{k=0}^{K-1} p^{k} \left\|  z^{k+1/2} - w^k  \right\|^2.
\end{align}
Hence, \eqref{t1114} together with \eqref{t697} gives
\begin{align*}
 \sum\limits_{k=0}^{K-1} &p^k \EE\|\hat z^{k+1} - z^* \|^2 + \sum\limits_{k=0}^{K-1} p^k \EE\|w^{k+1} - z^* \|^2 \nonumber\\
 &\leq \sum\limits_{k=0}^{K-1} p^k \EE\|\hat z^{k} - z^* \|^2 + \sum\limits_{k=0}^{K-1} p^k \EE\|w^{k} - z^* \|^2 - 2 \gamma \mu \sum\limits_{k=0}^{K-1} p^k \EE\| z^{k+1/2} - z^*\|^2 \nonumber\\
    & \hspace{0.4cm}-  (1-\tau - 2\gamma^2 \tilde L^2 ) \cdot \sum\limits_{k=0}^{K-1} p^k \EE\| w^k - z^{k+1/2} \|^2 - \left(\tau - \frac{1}{2}\right)\cdot \sum\limits_{k=0}^{K-1} p^k\EE\| z^{k+1/2} - z^{k}\|^2\nonumber\\
    &\hspace{0.4cm} +  768(\delta^{\text{serv}})^2\gamma^2 \tilde L^2 \sum\limits_{k=0}^{K-1}  p^{k} \EE\left\|  z^{k+1/2} - w^k  \right\|^2  \nonumber\\
&\hspace{0.4cm}+ \left(768(\delta^{\text{serv}})^2 + \frac{6M}{b} \right) \cdot 16(\delta^{\text{dev}})^2 \gamma^2 \tilde L^2 \frac{M}{b} \sum\limits_{k=0}^{K-1} p^{k} \left\|  z^{k+1/2} - w^k  \right\|^2 \\
&\leq \sum\limits_{k=0}^{K-1} p^k \EE\|\hat z^{k} - z^* \|^2 + \sum\limits_{k=0}^{K-1} p^k \EE\|w^{k} - z^* \|^2 - 2 \gamma \mu \sum\limits_{k=0}^{K-1} p^k \EE\| z^{k+1/2} - z^*\|^2 \nonumber\\
    & \hspace{0.4cm}-  \left(1-\tau - 768(\delta^{\text{serv}})^2\gamma^2 \tilde L^2 - 12400\gamma^2 (\delta^{\text{serv}})^2 (\delta^{\text{dev}})^2 \frac{M}{b}\tilde L^2 - 96 (\delta^{\text{dev}})^2 \gamma^2 \frac{M^2}{b^2} \tilde L^2  \right) \cdot \sum\limits_{k=0}^{K-1} p^k \EE\| w^k - z^{k+1/2} \|^2 \nonumber\\
&\hspace{0.4cm}
    - \left(\tau - \frac{1}{2}\right)\cdot \sum\limits_{k=0}^{K-1} p^k\EE\| z^{k+1/2} - z^{k}\|^2.
\end{align*}
The same as in Theorem \ref{th_comp} with $\tau \geq \frac{3}{4}$, $\gamma \leq \min\left[ \frac{1-\tau}{8\mu}; \frac{\sqrt{1-\tau}}{(30\delta^{\text{serv}} + 10 \delta^{\text{dev}}\frac{M}{b} + 165 \delta^{\text{dev}}\delta^{\text{serv}} \sqrt{\frac{M}{b}})\tilde L}\right]$

\textbf{Monotone and Non-monotone cases}

The proof of the monotone and non-monotone cases repeat the techniques from Theorem \ref{th_comp_vr} (modifications of Theorem \ref{th_comp}) + techniques and estimates obtained in the proof of the strongly-monotone case in Theorem \ref{th_comp_pp}.

\newpage

\section{\algnamenormal{CEG}: additional method} \label{sec:ceg}

In this section, we present \algname{CEG} with unbiased compression on devices. This is a very simple method. We prove its convergence only in the strongly monotone case and need this result to support the propositions in Sections \ref{sec:masha1_main} and \ref{sec:opt}.

\begin{algorithm}[th]
	\caption{\algname{CEG} : Compressed Extra Gradient}
	\label{alg_ceq}
	\begin{algorithmic}[1]
\State
\noindent {\bf Parameters:}  Stepsize $\gamma>0$, number of iterations $K$.\\
\noindent {\bf Initialization:} Choose  $z^0 \in \mathcal{Z}$. \\
Server  sends to devices $z^0$ 
\For {$k=0,1, 2, \ldots, K-1$ }
\For {each device $m$ in parallel}
\State  Compute $F_m(z^{k})$ \& send  $Q^{\text{dev}}_m(F_m(z^{k}))$ to server
\EndFor
\For {server}
\State Compute $\frac{1}{M} \sum \limits_{m=1}^M Q^{\text{dev}}_m(F_m(z^{k}))$ \& send to devices
\EndFor
\For {each device $m$ in parallel}
\State  $z^{k+1/2} = z^{k} - \gamma  \cdot\frac{1}{M} \sum \limits_{m=1}^M Q^{\text{dev}}_m(F_m(z^{k}))$ 
\State  Compute $F_m(z^{k+1/2})$ \& send  $Q^{\text{dev}}_m(F_m(z^{k+1/2}))$ to server
\EndFor
\For {server}
\State Compute $\frac{1}{M} \sum \limits_{m=1}^M Q^{\text{dev}}_m(F_m(z^{k+1/2}))$ \& send to devices
\EndFor
\For {each device $m$ in parallel}
\State  $z^{k+1} = z^{k} - \gamma  \cdot \frac{1}{M} \sum \limits_{m=1}^M Q^{\text{dev}}_m(F_m(z^{k+1/2}))$ 
\EndFor
\EndFor
	\end{algorithmic}
\end{algorithm}

\begin{theorem} \label{th_ceg}
Let distributed variational inequality \eqref{VI} + \eqref{MK} is solved by Algorithm \ref{alg_ceq} with unbiased compressor operators \eqref{quant} on devices with $\{q^{\text{dev}}_m = q\}$. Let Assumptions \ref{as3} and  \ref{as1} (SM) are satisfied. Then the following estimates holds with $\gamma \leq \min\left[\frac{\mu}{48L^2 } \cdot \left( 1 + \nicefrac{q}{M}\right)^{-1}; \frac{1}{4\mu}\right]$:
\begin{align*}
 \EE\| z^{K} - z^* \|^2 &\leq \left( 1 -\frac{\mu\gamma}{2} \right)^K \cdot \|z^{0} - z^* \|^2 + \frac{16 q \gamma^2}{M^2} \sum\limits_{m=1}^M \EE\left[\left\|F_m(z^*)\right\|^2\right];
\end{align*}
\end{theorem}

With $F_m(z^*) = 0$ we get the following estimates on iteration and bits complexities:
$$
\mathcal{O}\left( \left[ \left( 1 + \frac{q}{M}\right) \frac{L^2}{\mu^2}\right] \log \frac{1}{\varepsilon} \right), \quad \mathcal{O}\left( \left[ \left( \frac{1}{\beta} + \frac{q}{M \beta}\right) \frac{L^2}{\mu^2}\right] \log \frac{1}{\varepsilon} \right).
$$

\textbf{Proof of Theorem \ref{th_ceg}:}
By a classical analysis of Extra Gradient in the strongly monotone (see \cite{gidel2018a} or \cite{beznosikov2021distributed1}) case we get
\begin{align*}
 \E\left[\| z^{k+1} - z^* \|^2\right]  &\leq \E\left[\| z^k - z^* \|^2\right] - \E\left[\| z^{k+1/2} -  z^k \|^2\right] \notag\\
 &\quad- 2 \gamma \E\left[\langle  g^{k+1/2},  z^{k+1/2} - z^* \rangle\right] + \gamma^2 \E\left[\| g^{k+1/2} -  g^{k}\|^2\right],  
 \end{align*}
where $g^{k+1/2} = \frac{1}{M} \sum\limits_{m=1}^M Q\left( F_m(x^{k+1/2}) \right)$ and $g^{k} = \frac{1}{M} \sum\limits_{m=1}^M Q\left( F_m(x^{k}) \right)$. With unbiasedness of compression we get
\begin{align*}
 \E\left[\| z^{k+1} - z^* \|^2\right]  &\leq \E\left[\| z^k - z^* \|^2\right] - \E\left[\| z^{k+1/2} -  z^k \|^2\right] \notag\\
 &\quad- 2 \gamma \E\left[\langle  F(z^{k+1/2}),  z^{k+1/2} - z^* \rangle\right] + 2\gamma^2 \E\left[\| g^{k+1/2} -  F(z^*)\|^2\right]  + 2\gamma^2 \E\left[\| F(z^*) -  g^{k}\|^2\right].  
 \end{align*}
Then 
\begin{align*}
        \EE\left[\left\|\frac{1}{M} \sum\limits_{m=1}^M Q\left( F(z^{k}) \right) - F(z^{*})\right\|^2\right]
        &= \EE\left[\left\|\frac{1}{M} \sum\limits_{m=1}^M \left[ Q\left( F_m(z^{k}) \right) -  F_m(z^{k}) +  F_m(z^{k})\right] - F(z^{*})\right\|^2\right] \\
        &\leq 2\EE\left[\left\|\frac{1}{M} \sum\limits_{m=1}^M Q\left( F_m(z^{k}) \right) -  F_m(z^{k}) \right\|^2\right] \\
        &\quad+ 2\EE\left\| F(z^k) - F(z^{*})\right\|^2
        \\
        &= \frac{2}{M^2} \sum\limits_{m=1}^M \EE\left[\left\|Q\left(F_m(z^{k})\right) - F_m(z^{k}) \right\|^2\right] \\
        &\quad+ \frac{2}{M^2} \sum\limits_{i\neq j}^M \EE\left[\langle Q\left(F_i(z^{k})\right) - F_i(z^{k}), Q\left(F_j(z^{k})\right) - F_j(z^{k}) \rangle\right] \\
        &\quad+ 2\EE\left[\left\| F(z^k) - F(z^{*})\right\|^2\right].
\end{align*}
Using definition of \eqref{quant}, we get
\begin{align*}
        \EE\left[\left\|\frac{1}{M} \sum\limits_{m=1}^M Q\left( F(z^{k}) \right) - F(z^{*})\right\|^2\right]
        &\leq \frac{2 q}{M^2} \sum\limits_{m=1}^M \EE\left[\left\|F_m(z^{k}) \right\|^2\right] \\
        &\quad+ 2\EE\left[\left\| F(z^k) - F(z^{*})\right\|^2\right]\\
        &\leq \frac{4 q}{M^2} \sum\limits_{m=1}^M \EE\left[\left\|F_m(z^{k}) - F_m(z^*) \right\|^2\right] + \frac{4 q}{M^2} \sum\limits_{m=1}^M \EE\left[\left\|F_m(z^*)\right\|^2\right] \\
        &\quad+ 2\left[\left\| F(z^k) - F(z^{*})\right\|^2\right].
\end{align*}
With Assumptions \ref{as3} we get
\begin{align*}
        \EE\left[\left\|\frac{1}{M} \sum\limits_{m=1}^M Q\left( F(z^{k}) \right) - F(z^{*})\right\|^2\right]
        &\leq \frac{4 q L^2}{M} \EE\left[\left\|z^{k} - z^*\right\|^2\right] + \frac{4 q}{M^2} \sum\limits_{m=1}^M \EE\left[\left\|F_m(z^*)\right\|^2\right] \\
        &\quad+ 2 L^2\EE\left[\left\| z^k - z^{*}\right\|^2\right] \\
        &\leq 4 L^2 \left( 1 + \frac{q}{M}\right) \EE\left[\left\|z^{k} - z^*\right\|^2\right] + \frac{4 q}{M^2} \sum\limits_{m=1}^M \EE\left[\left\|F_m(z^*)\right\|^2\right].
\end{align*}
The same way we can get
\begin{align*}
        \EE\left[\left\|\frac{1}{M} \sum\limits_{m=1}^M Q\left( F(z^{k+1/2}) \right) - F(z^{*})\right\|^2\right]
        &\leq \frac{4 q L^2}{M} \EE\left[\left\|z^{k} - z^*\right\|^2\right] + \frac{4 q}{M^2} \sum\limits_{m=1}^M \EE\left[\left\|F_m(z^*)\right\|^2\right] \\
        &\quad+ 2 L^2\EE\left[\left\| z^k - z^{*}\right\|^2\right] \\
        &\leq 4 L^2 \left( 1 + \frac{q}{M}\right) \EE\left[\left\|z^{k+1/2} - z^*\right\|^2\right] + \frac{4 q}{M^2} \sum\limits_{m=1}^M \EE\left[\left\|F_m(z^*)\right\|^2\right].
\end{align*}
Finally, we obtain
\begin{align*}
 \E\left[\| z^{k+1} - z^* \|^2\right]  &\leq \E\left[\| z^k - z^* \|^2\right] - \E\left[\| z^{k+1/2} -  z^k \|^2\right] \notag\\
 &\quad- 2 \gamma \E\left[\langle  F(z^{k+1/2}),  z^{k+1/2} - z^* \rangle\right] + 8\gamma^2 L^2 \left( 1 + \frac{q}{M}\right) \EE\left[\left\|z^{k} - z^*\right\|^2\right] \\
  &\quad + 8\gamma^2 L^2 \left( 1 + \frac{q}{M}\right) \EE\left[\left\|z^{k+1/2} - z^*\right\|^2\right] + \frac{16 q \gamma^2}{M^2} \sum\limits_{m=1}^M \EE\left[\left\|F_m(z^*)\right\|^2\right].  
 \end{align*}
With Assumption \ref{as1}(SM) we obtain
\begin{align*}
 \E\left[\| z^{k+1} - z^* \|^2\right]  &\leq \E\left[\| z^k - z^* \|^2\right] - \E\left[\| z^{k+1/2} -  z^k \|^2\right] \notag\\
 &\quad- 2 \gamma\mu \EE\left[\left\|z^{k+1/2} - z^*\right\|^2\right] + 8\gamma^2 L^2 \left( 1 + \frac{q}{M}\right) \EE\left[\left\|z^{k} - z^*\right\|^2\right] \\
  &\quad + 8\gamma^2 L^2 \left( 1 + \frac{q}{M}\right) \EE\left[\left\|z^{k+1/2} - z^*\right\|^2\right] + \frac{16 q \gamma^2}{M^2} \sum\limits_{m=1}^M \EE\left[\left\|F_m(z^*)\right\|^2\right] \\
   &\leq \E\left[\| z^k - z^* \|^2\right] - \E\left[\| z^{k+1/2} -  z^k \|^2\right] \notag\\
 &\quad- \gamma\mu \EE\left[\left\|z^{k} - z^*\right\|^2\right] + 2\gamma\mu \EE\left[\left\|z^{k+1/2} - z^k\right\|^2\right] + 8\gamma^2 L^2 \left( 1 + \frac{q}{M}\right) \EE\left[\left\|z^{k} - z^*\right\|^2\right] \\
  &\quad + 16\gamma^2 L^2 \left( 1 + \frac{q}{M}\right) \EE\left[\left\|z^{k+1/2} - z^k\right\|^2\right] + 16\gamma^2 L^2 \left( 1 + \frac{q}{M}\right) \EE\left[\left\|z^{k} - z^*\right\|^2\right] \\
  &\quad + \frac{16 q \gamma^2}{M^2} \sum\limits_{m=1}^M \EE\left[\left\|F_m(z^*)\right\|^2\right].  
 \end{align*}
 With $\gamma \leq \frac{1}{4\mu}; \frac{\mu}{48L^2 \left( 1 + \nicefrac{q}{M}\right)}$ we have
 \begin{align*}
 \E\left[\| z^{k+1} - z^* \|^2\right] 
   &\leq \left( 1 - \gamma\mu + 24\gamma^2 L^2 \left( 1 + \frac{q}{M}\right) \right)\E\left[\| z^k - z^* \|^2\right] + \frac{16 q \gamma^2}{M^2} \sum\limits_{m=1}^M \EE\left[\left\|F_m(z^*)\right\|^2\right].  
 \end{align*}
 Choice $\gamma \leq \frac{\mu}{48 L^2  \left( 1 + \frac{q}{M}\right) }$ finishes the proof.
 \EndProof
\newpage

\end{document}